\documentclass[runningheads]{llncs}
\usepackage{graphicx}

\usepackage{tikz}
\usepackage{comment}
\usepackage{amsmath,amssymb} 
\usepackage{color}
\usepackage{float}
\usepackage[accsupp]{axessibility}  

\usepackage{multirow}
\usepackage{array}
\usepackage{makecell}
\usepackage{subfigure}

\begin{document}
\pagestyle{headings}
\mainmatter
\def\ECCVSubNumber{5051}  

\title{Don't Forget Me: Accurate Background Recovery for Text Removal via Modeling Local-Global Context} 

\titlerunning{Contextual-guided Text Removal Network}
\author{Chongyu Liu\inst{1} \and
Lianwen Jin$^{*}$\inst{1,4,5} \and
Yuliang Liu\inst{2}  \and Canjie Luo \inst{1} \and Bangdong Chen \inst{1} \and Fengjun Guo \inst{3} \and Kai Ding \inst{3}}

\authorrunning{Liu et al.}
%
\institute{South China University of Technology, Guangzhou, Guangdong, China\\
	 \email{\{liuchongyu1996, lianwen.jin, canjie.luo\}@gmail.com} \\
	\and 
Huazhong University of Science and Technology, Wuhan, Hubei, China\\ 
	 \email{ylliu@hust.edu.cn} \\
	\and
IntSig Information Co. Ltd, Shanghai, China \\
\email{\{fengjun\_guo, danny\_ding\}@intsig.net}
	\and Pazhou Laboratory (Huangpu), Guangzhou, Guangdong, China \and Peng Cheng Laboratory, Shenzhen, Guangdong, China
}
\maketitle

\begin{abstract}
	Text removal has attracted increasingly attention due to its various applications on privacy protection,
	document restoration, and text editing. It has shown significant progress with deep neural network.
	However, most of the existing methods 
	often generate inconsistent results for complex background. 
	To address this issue, we propose a Contextual-guided Text Removal Network, termed as CTRNet. CTRNet explores both low-level structure and high-level discriminative context feature as prior knowledge to guide the process of text erasure and background restoration. We further propose a Local-global Content Modeling (LGCM) block with CNNs and Transformer-Encoder to capture local features and establish the long-term relationship among pixels globally.
	Finally, we incorporate LGCM with context guidance for feature modeling and decoding.
	Experiments on benchmark datasets, SCUT-EnsText and SCUT-Syn show that CTRNet significantly outperforms the existing state-of-the-art methods. Furthermore, a qualitative experiment on examination papers also demonstrates the generalizability of our method. The code of CTRNet is available at https://github.com/lcy0604/CTRNet. 

\end{abstract}

\section{Introduction}

In recent years, text removal has attracted increasing research interests in the computer vision community. 
It aims to remove the text and fill the regions with plausible content.  
Text removal can help avoid privacy leaks by hiding some private messages such as ID numbers and license plate numbers. Besides, it can be widely used for document restoration in the field of intelligent education. 
It is also a crucial prerequisite step for text editing~\cite{wu2019editing,yang2020swaptext,yu2021mask,krishnan2021textstylebrush,shimoda2021rendering} and has wide applications in areas such as augmented reality translation.

\begin{figure}[t]
	\subfigbottomskip=2pt
	\subfigcapskip=2pt
	\setlength{\abovecaptionskip}{-0.0cm}
	\setlength{\belowcaptionskip}{-0.6cm} 
	\centering
	\subfigure[Input]{
		\begin{minipage}[t]{0.21\linewidth}
			\centering
			\includegraphics[width=2.3cm,height=2cm]{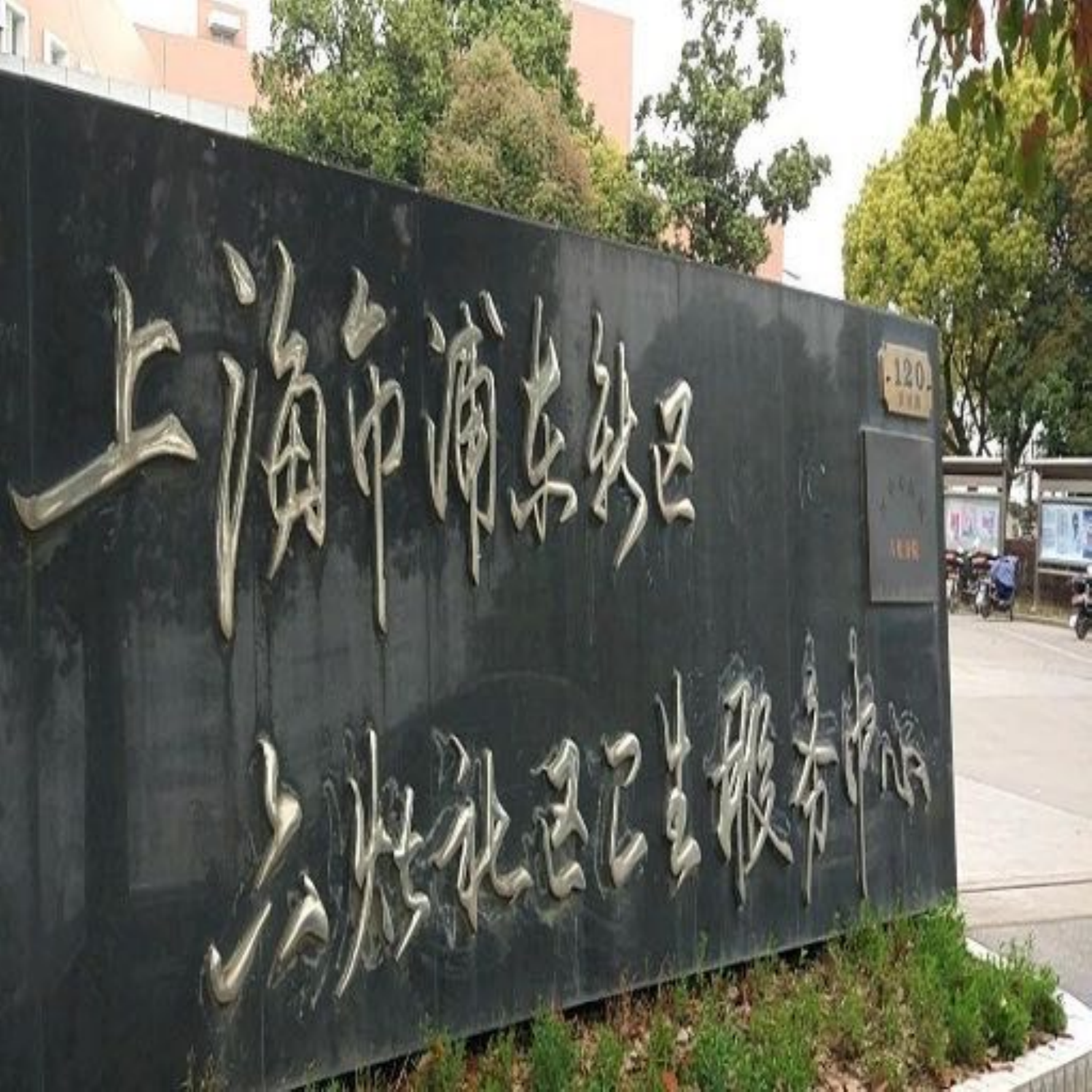}
			\includegraphics[width=2.3cm,height=2cm]{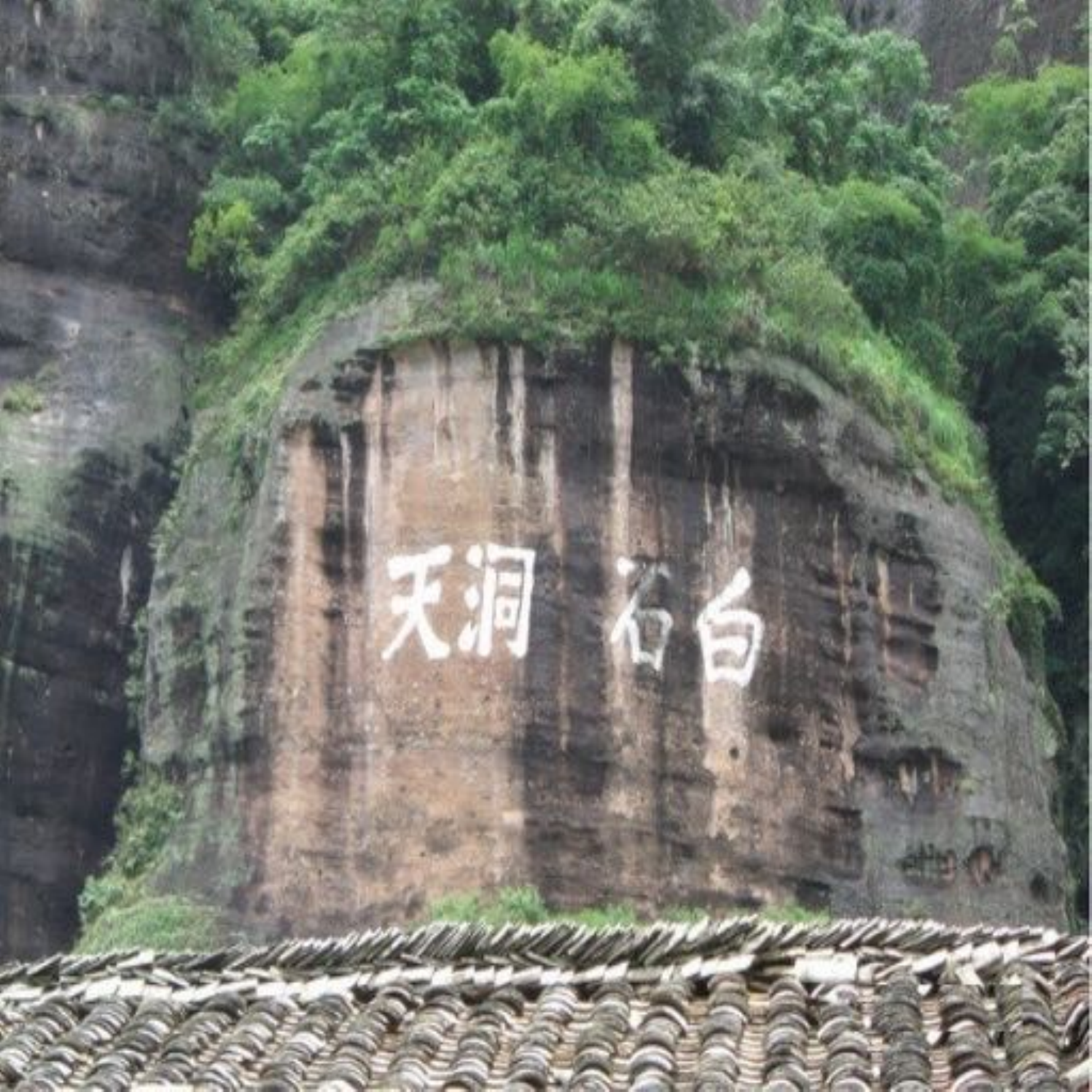}
		\end{minipage}%
	}
	\subfigure[w/o Context]{
	\begin{minipage}[t]{0.21\linewidth}
		\centering
		\includegraphics[width=2.3cm,height=2cm]{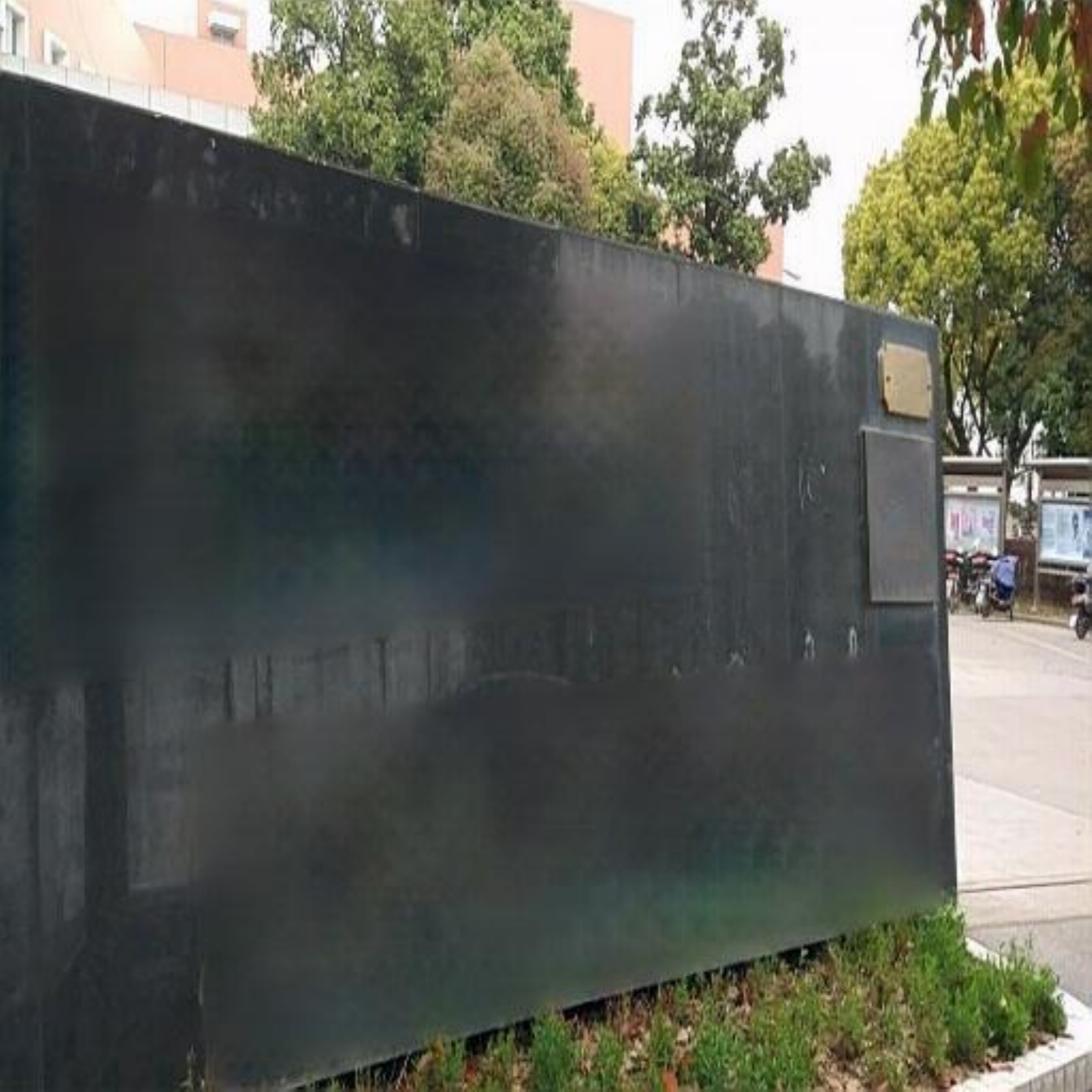}
		\includegraphics[width=2.3cm,height=2cm]{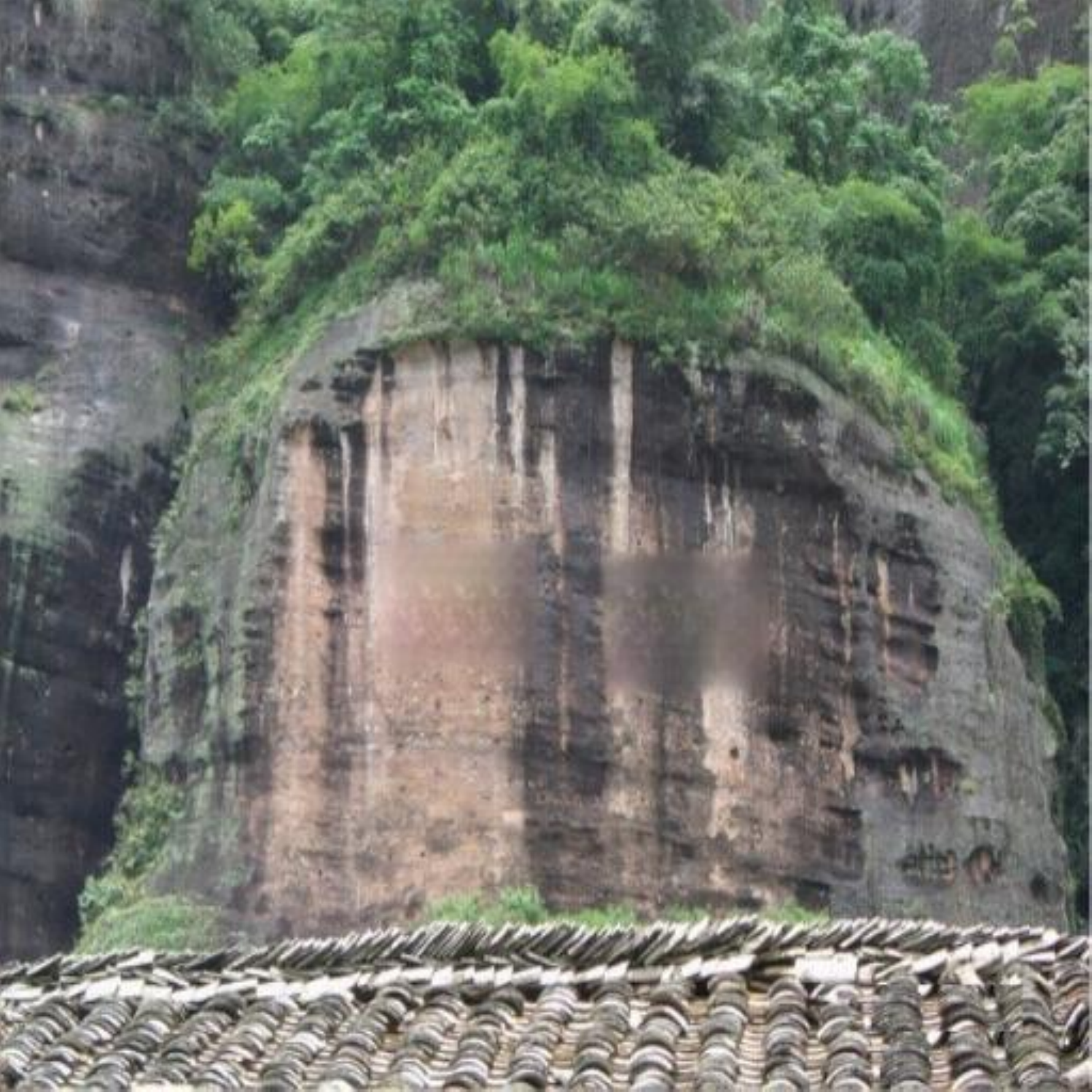}
	\end{minipage}%
}
	\subfigure[w Context]{
	\begin{minipage}[t]{0.21\linewidth}
		\centering
		\includegraphics[width=2.3cm,height=2cm]{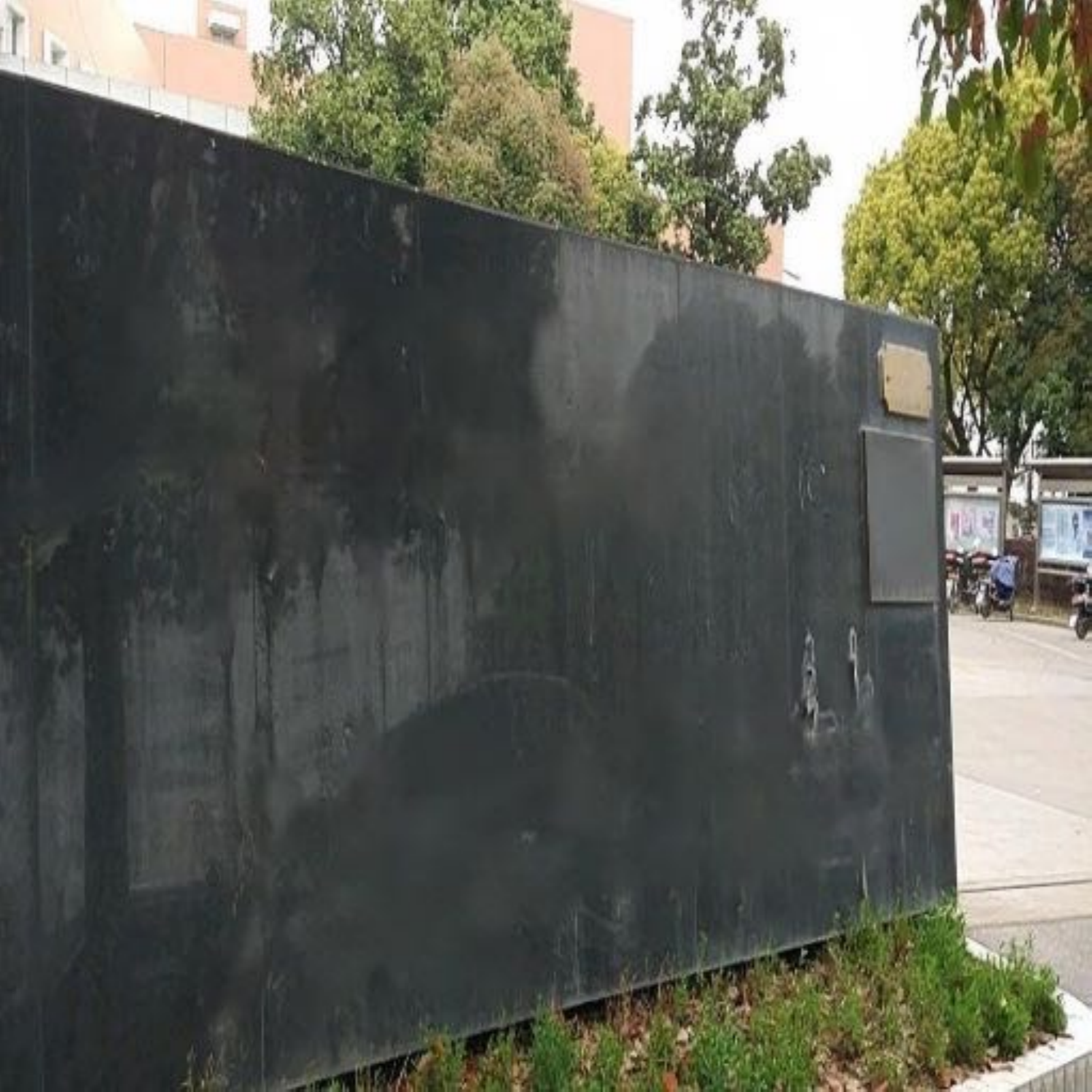}
		\includegraphics[width=2.3cm,height=2cm]{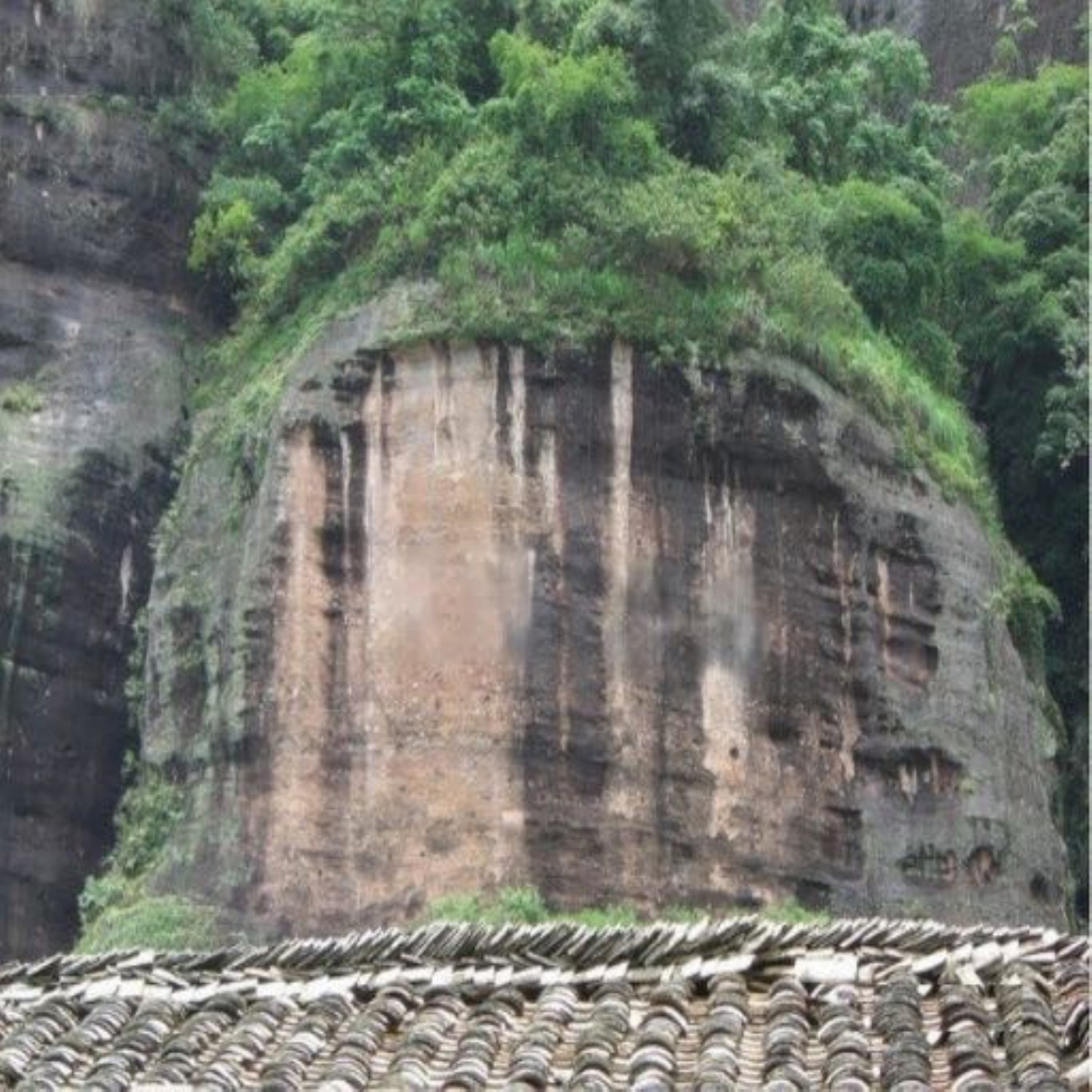}
	\end{minipage}%
}
	\subfigure[GT]{
	\begin{minipage}[t]{0.21\linewidth}
		\centering
	\includegraphics[width=2.3cm,height=2cm]{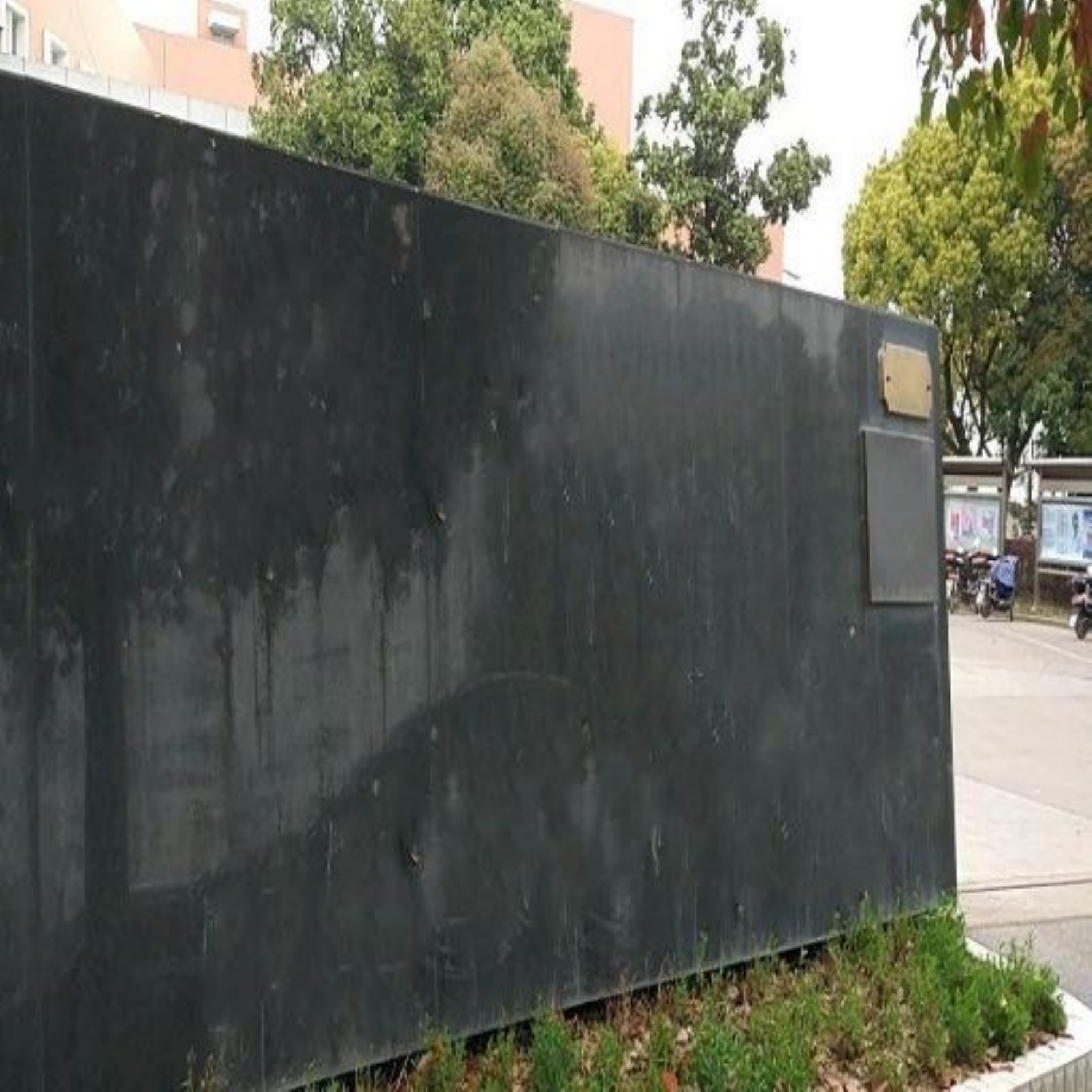}
	\includegraphics[width=2.3cm,height=2cm]{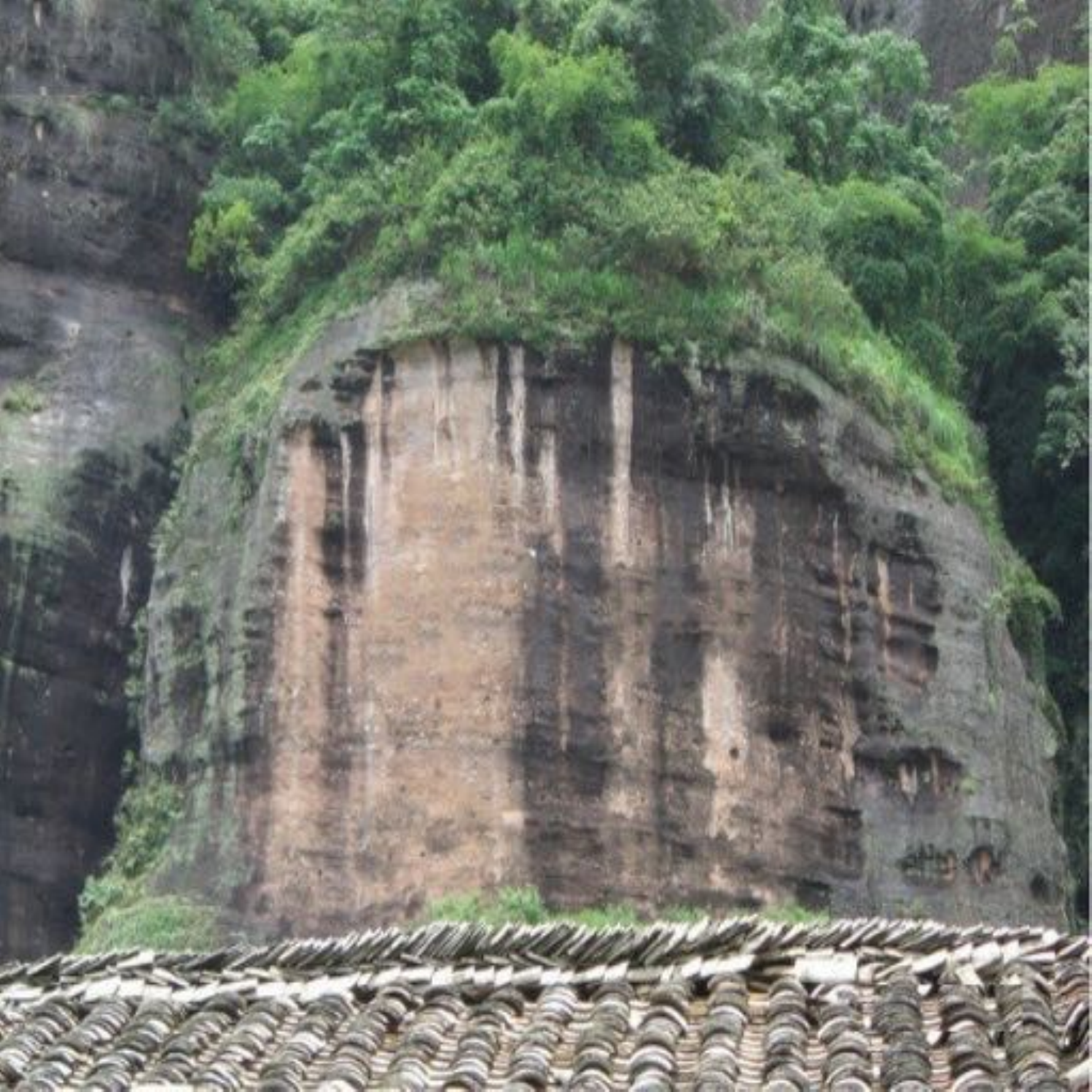}
	\end{minipage}%
}
	\centering
	\caption{ Examples of scene text removal, which also show the comparison of the results with and without context guidance and feature modeling. Zoom-in for best view.} \label{fig:text}
\end{figure}

Recent text removal methods~\cite{zhang2019ensnet,liu2020erasenet,wang2021simple,tang2021stroke,tursun2019mtrnet} have achieved significant improvements with the development of GAN~\cite{gan,cgan,miyato2018spectral}.
Though the state-of-the-art methods~\cite{liu2020erasenet,wang2021simple,tang2021stroke,tursun2020mtrnet++} have reported promising performance, the restoration for complex backgrounds still remains a main challenge. To solve this problem, some researchers propose to directly predict the text stroke~\cite{tang2021stroke,bian2022scene} and focus text region inpainting only on these stroke regions. 
However, text stroke prediction is an another challenging problem to be addressed, especially on image-level (with more than one text)~\cite{xu2021rethinking,wang2021semi}. 
Inspired by previous image inpainting methods~\cite{liu2020rethinking,ren2019structureflow,nazeri2019edgeconnect}, 
we consider that directly transforming the raw image to a final text-erased image in a unified framework is one of the major causes of inconsistent results for text removal. 
This is due to the imbalance between text erasure and the subsequent background restoration. The corruption of text region while erasing may mislead the reconstruction of the high-frequency textures.  
The results with blur and artifacts are shown in Fig. \ref{fig:text} (b).
To address this issue, we propose to mine more efficient context guidance from the existing data in a step-by-step manner to reduce the artifacts of text-erased regions and produce plausible content.

\begin{figure*}[t]
 	\centering
 	\setlength{\abovecaptionskip}{-0.0cm}
 	\setlength{\belowcaptionskip}{-0.4cm} 
 	\includegraphics[width = 10.5cm, height = 6cm]{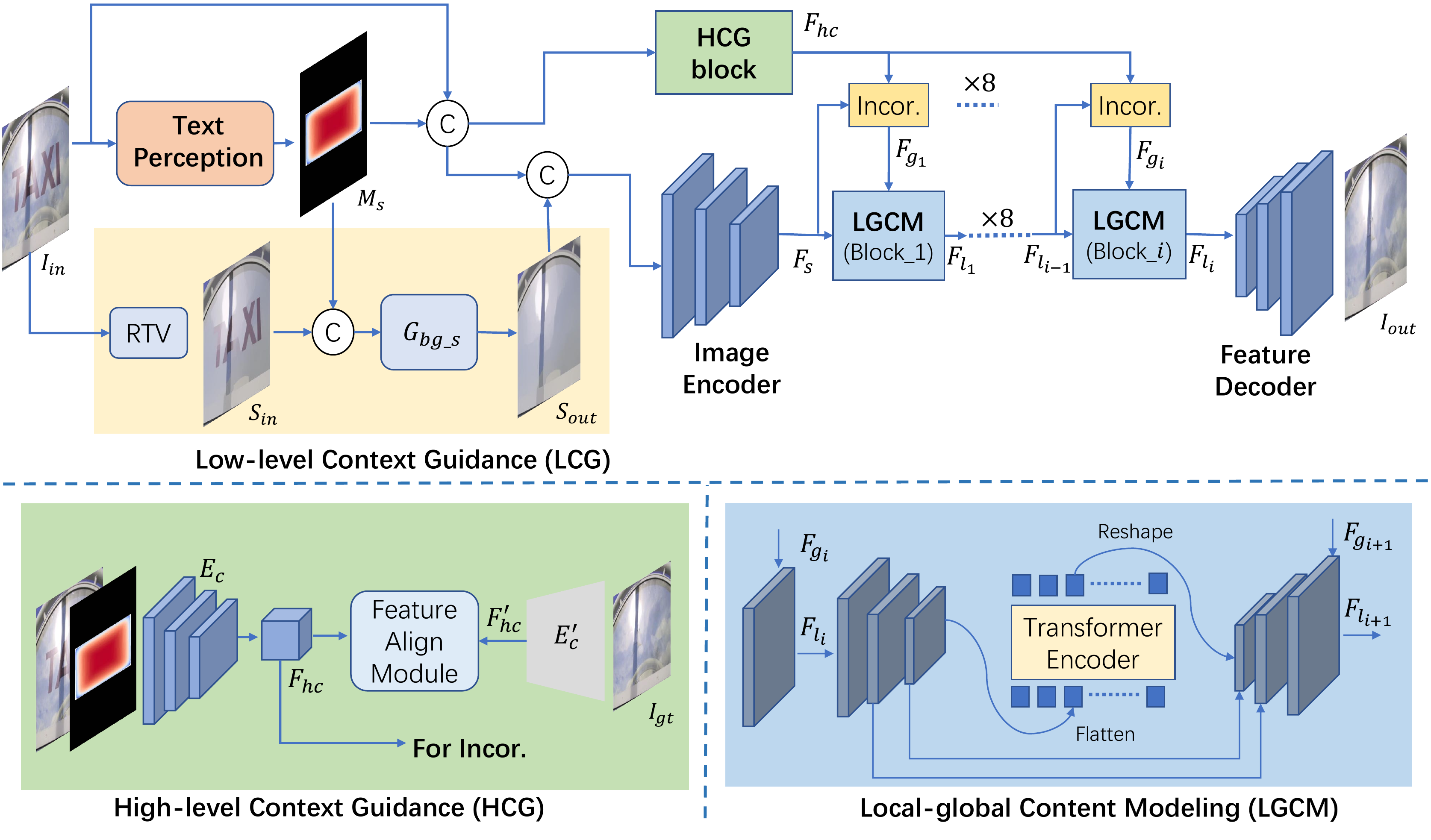}
 	\centering
 	\caption{The overview of the proposed CTRNet. 
 	} \label{fig:model}
 \end{figure*}

Specifically, we propose a novel text removal model, termed as CTRNet. CTRNet decouples the text removal task into four main components: Text Perception Head, Low-level/High-level Contextual Guidance blocks (LCG, HCG), and a Local-global Content Modeling (LGCM) block, as shown in Fig. \ref{fig:model}. Text Perception Head is firstly introduced to detect the text regions and generate text masks. Subsequently, 
the LCG predicts the structure of text-erased images to provide low-level contextual priors,
which is represented by the edge-preserved smoothing method RTV~\cite{xu2012structure}.
Besides, we incorporate an HCG block to learn the high-level discriminative context in latent feature space as another guidance. Structure is served as a local guide for the image encoder, while high-level context provides global knowledge.
As the filling of text regions not only focuses on the information of their own and surroundings, but also uses the global affinity as reference, CTRNet introduces LGCM by the cooperation of CNNs and Transformer-Encoder \cite{vaswani2017attention} to extract local features and establish the long-term global relationship among the pixels, meanwhile incorporates context guidance for both feature modeling and decoding phase. Through such designs, CTRNet can capture sufficient contextual information to remove the text more thoroughly and restore backgrounds with more visually pleasing textures, as shown in Fig.~\ref{fig:text}~(c).

Extensive experiments on the benchmark datasets, SCUT-EnsText \cite{liu2020erasenet} and SCUT-Syn~\cite{zhang2019ensnet} are conducted to verify the effectiveness of CTRNet. Additionally, qualitative experiment is conducted on an in-house examination paper dataset to verify the generalizability of our model.

Text removal takes complete text image as input and aims to preserve the original background of text regions, whereas image inpainting will directly mask the regions for restoration based only on the surrounding texture. Simply applying image inpainting methods to text removal will cause inaccurate background generation. 
We conduct experiments to compare our method with the state-of-the-art image inpainting models in Sec.4.5/4.6, which practically illustrates the difference between these two tasks.

We summarize the contributions of this work as follows:
\begin{itemize}
	\item We propose to learn both Low-level and High-level Contextual Guidance (LCG, HCG), which we find are important and useful as prior knowledge for text erasure and subsequent background texture synthesis.
	\item We propose Local-global Content Modeling blocks (LGCM) to extract local features and capture long-range dependency among the pixels globally. 
	\item The context guidance is incorporated into LGCM for the feature modeling and decoding phase, which further promotes the performance of CTRNet.
	\item Extensive experiments on the benchmark datasets demonstrate the effectiveness of CTRNet not only in removing the text but recovering the background textures as well, significantly outperforming existing SOTA methods.
\end{itemize}

\section{Related work}

\noindent \textbf{Deep learning-based text removal} can be categorized into one-stage methods and two-stage methods. One-stage methods are implemented in an end-to-end manner, requiring models to automatically detect the text regions and remove them in a unified framework. Nakamura et al.~\cite{ste} proposed a patch-based auto-encoder~\cite{bengio2013representation} with skip connections, termed as SceneTextEraser. It was also the first DNN-based text removal method.
Text removal can be also regarded as image-to-image translation. 
Following the idea of Pix2pix~\cite{isola2017image}, EnsNet \cite{zhang2019ensnet} adopted four refined losses and employed a local-aware discriminator to maintain the consistency of text-erased regions. Liu et al. \cite{liu2020erasenet} proposed EraseNet by introducing a coarse-to-refinement architecture and an additional segmentation head to help locate the text. MTRNet++ \cite{tursun2020mtrnet++} shared the same spirit with EraseNet, but separately encoded the image content and text mask in two branches. Cho et al. \cite{cho2021detecting} proposed to jointly predict the text stroke and inpaint the background, allowing the model to focus only on the restoration of text stroke regions. Wang et al. \cite{wang2021simple} presented PERT, which contained a novel progressive structure with shared parameters to remove text more thoroughly, and a region-based modification strategy to effectively guide the erasure process only on text regions.
 
 Two-stage methods follow the procedure of detecting the text, removing it, and then filling the background with plausible content. 
 We further divide them into word-level and image-level. 
 Word-level methods first crop the text regions according to the detected results, then operate the text removal process with single text~\cite{qin2018automatic,tang2021stroke}. Qin et al. \cite{qin2018automatic} utilized cGAN~\cite{gan,cgan} with one encoder and two decoders for both text stroke prediction and background inpaint. Tang et al. \cite{tang2021stroke} proposed to predict the text strokes on word images, then both strokes and images were fed into an image inpainting network with Partial Convolution \cite{pc} to generate the text-erased results. For image-level methods, after obtaining the text mask through detection, they directly predict the results on the entire images. MTRNet \cite{tursun2019mtrnet}, based on Pix2pix, implemented a text mask as an extra input. The method proposed by Keserwani et al. \cite{keserwani2021text} was similar to MTRNet, but employed an additional local discriminator for better prediction. 
Zdenek et al. \cite{Zdenek_2020_WACV} considered the lack of pixel-wise training data and proposed a weak supervision method by introducing a pretrained PSENet \cite{wang2019shape} to detect the text, and then inpainted the text regions through another pretrained image inpainting method \cite{zheng2019pluralistic}. Conrad et al. \cite{conrad2021two} borrowed the concept developed by Zdenek et al. \cite{Zdenek_2020_WACV}, but they proposed to further predict the text stroke before the application of a pretrained EdgeConnect \cite{nazeri2019edgeconnect} for background inpainting. Bian et al. \cite{bian2022scene} proposed a cascaded generative model, which decoupled text removal into text stroke detection and stroke removal.

\section{Proposed Method}
Fig. \ref{fig:model} shows the pipeline of the proposed CTRNet. First, we introduce text perception head to detect the text regions and generate the text masks. To better restore the backgrounds of text regions, we propose to learn more contextual priors from the existing data, including low-level background structure with LCG and high-level context features with HCG. Structure information is served as a local guide and directly fed into the image encoder, while the high-level context feature is embedded into the high-dimensional feature space as a global guide with the Incor operation. Finally, we propose LGCM blocks to capture both local features and long-term correlation among all the pixels, so that CTRNet can make full use of different levels of information for feature decoding.

\subsection{Text Perception Head}

For scene text removal on image-level, 
purely feeding a text image into a model without any positional indication results in failed, mistaken, and incomplete erasures of text regions~\cite{zhang2019ensnet,liu2020erasenet,wang2021simple}.
 Therefore, we introduce a text perception head to help localize the text regions. With the detected results, we generate the corresponding masks and send them together with original images into the subsequent network. We propose to replace the original 0-1 mask (hard mask) with soft mask to help eliminate the defects and discontinuities between text regions and non-text regions. 
 The procedure for soft mask generation is as follows: (1) The vanilla bounding boxes $B$ are shrunk using the Vatti clipping algorithm \cite{vatti1992generic} with the ratio of 0.9 to obtain $B_{s}$, meanwhile dilated with the same offset to $B_{d}$; (2) The soft border of text regions is defined as the minimum distance between the pixel in $B_{s}$ and $B_{d}$. Fig. \ref{fig:label} (c) displays the example of soft-mask. Only the pixels in $B_{s}$ are set to 1, while the range of pixels between $B_{s}$ and $B_{d}$ is $(0, 1)$. The effectiveness of the soft mask is verified in Section 4.3.

\subsection{Contextual Guidance Learning}

\begin{figure}[t]
	\subfigbottomskip=-1pt
	\subfigcapskip=-1pt
	\centering
	\subfigure[Input]{
		\begin{minipage}[t]{0.14\linewidth}
			\centering
			\includegraphics[width=1.9cm,height=2cm]{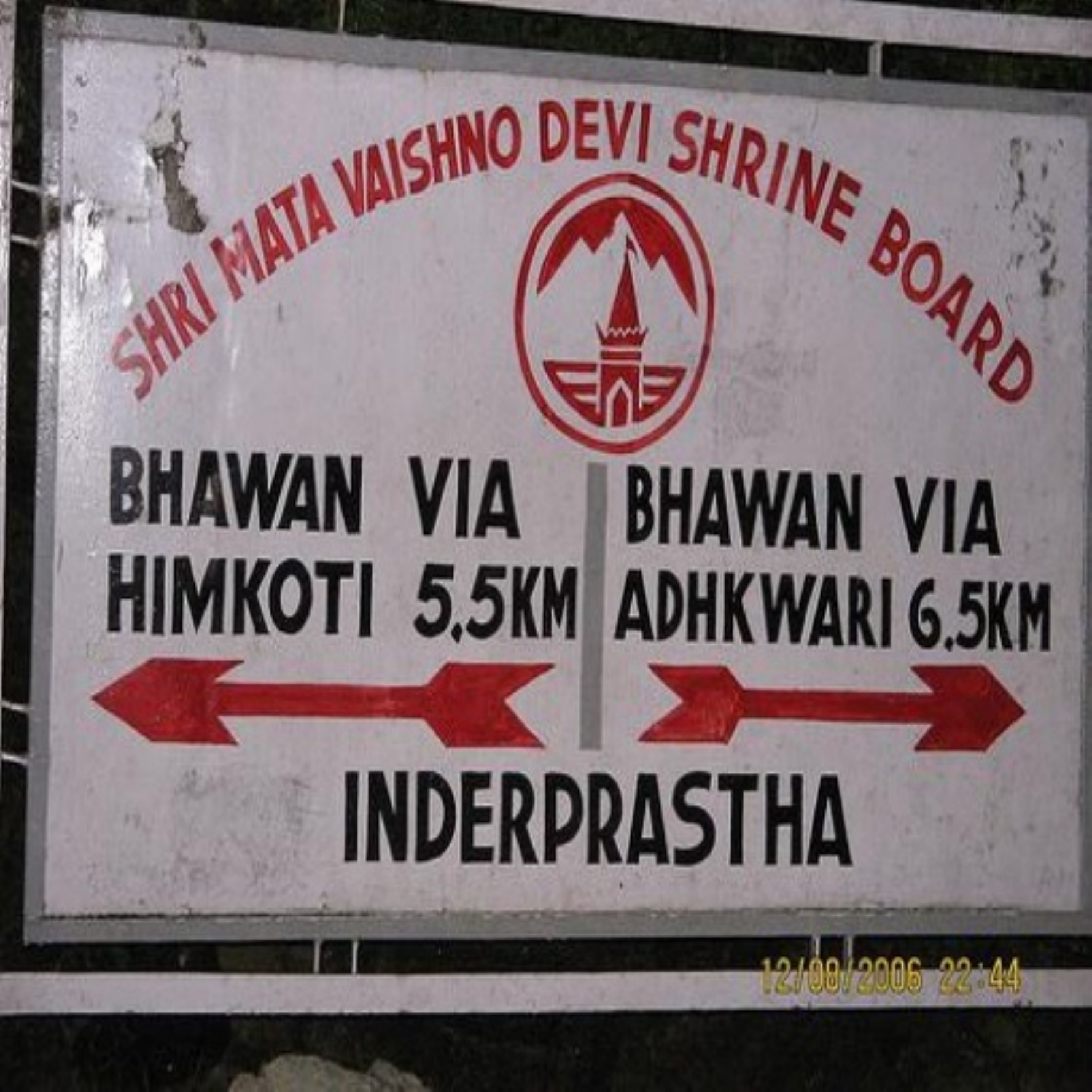}\\
		\end{minipage}%
	}
	\subfigure[HM]{
		\begin{minipage}[t]{0.14\linewidth}
			\centering
			\includegraphics[width=1.9cm,height=2cm]{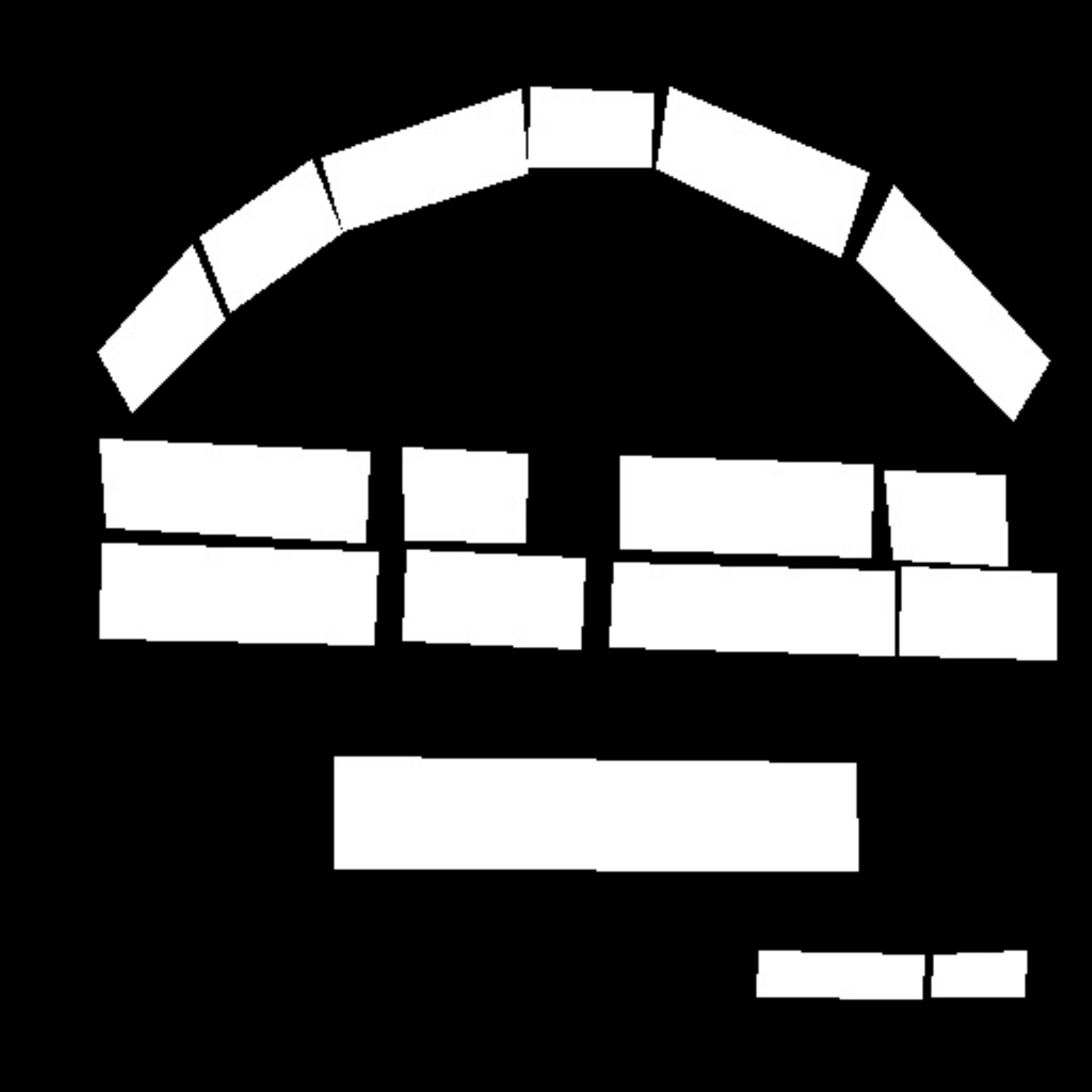}
		\end{minipage}%
	}
	\subfigure[SM]{
		\begin{minipage}[t]{0.14\linewidth}
			\centering
			\includegraphics[width=1.9cm,height=2cm]{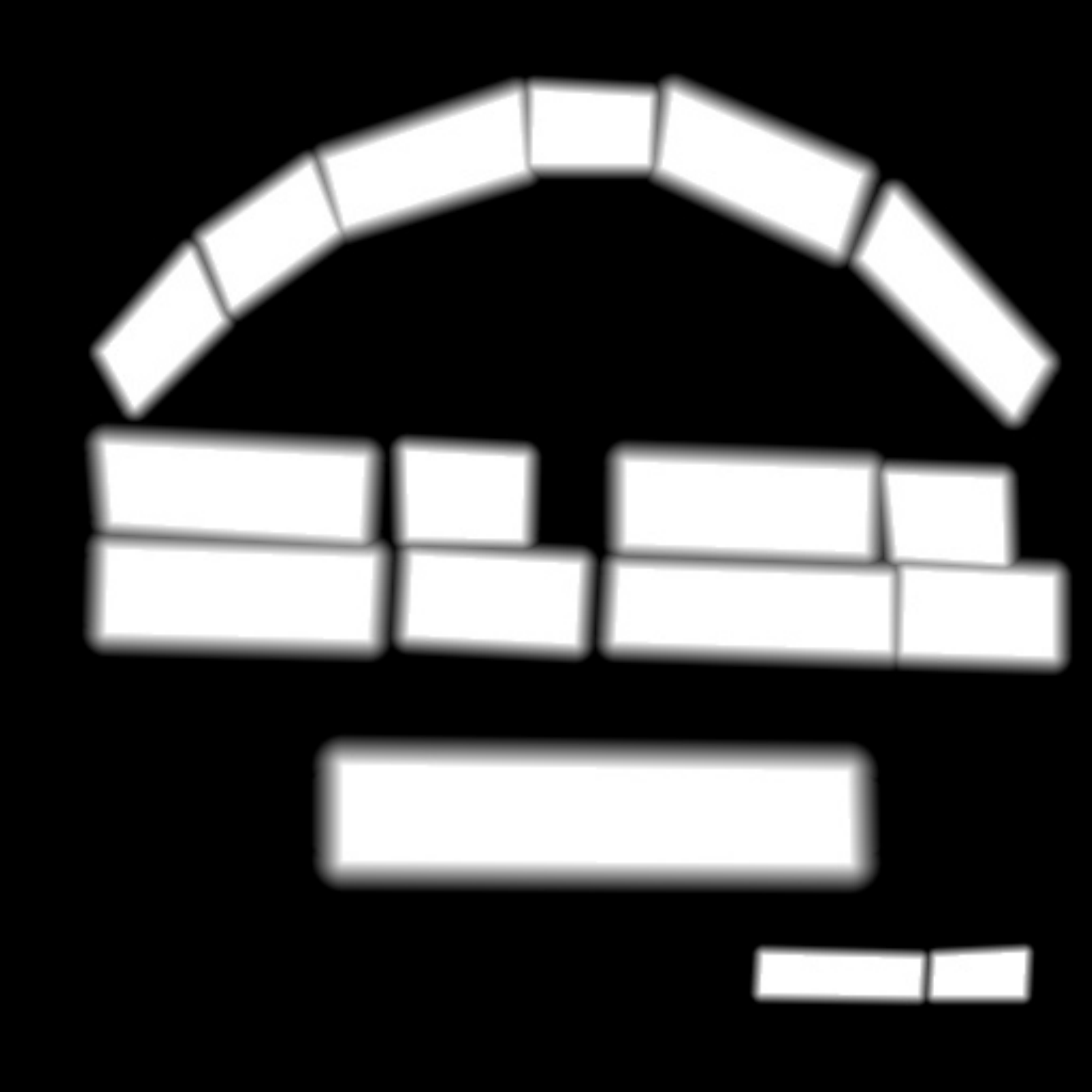}\\
		\end{minipage}%
	}
	\subfigure[$S_{in}$]{
		\begin{minipage}[t]{0.14\linewidth}
			\centering
			\includegraphics[width=1.9cm,height=2cm]{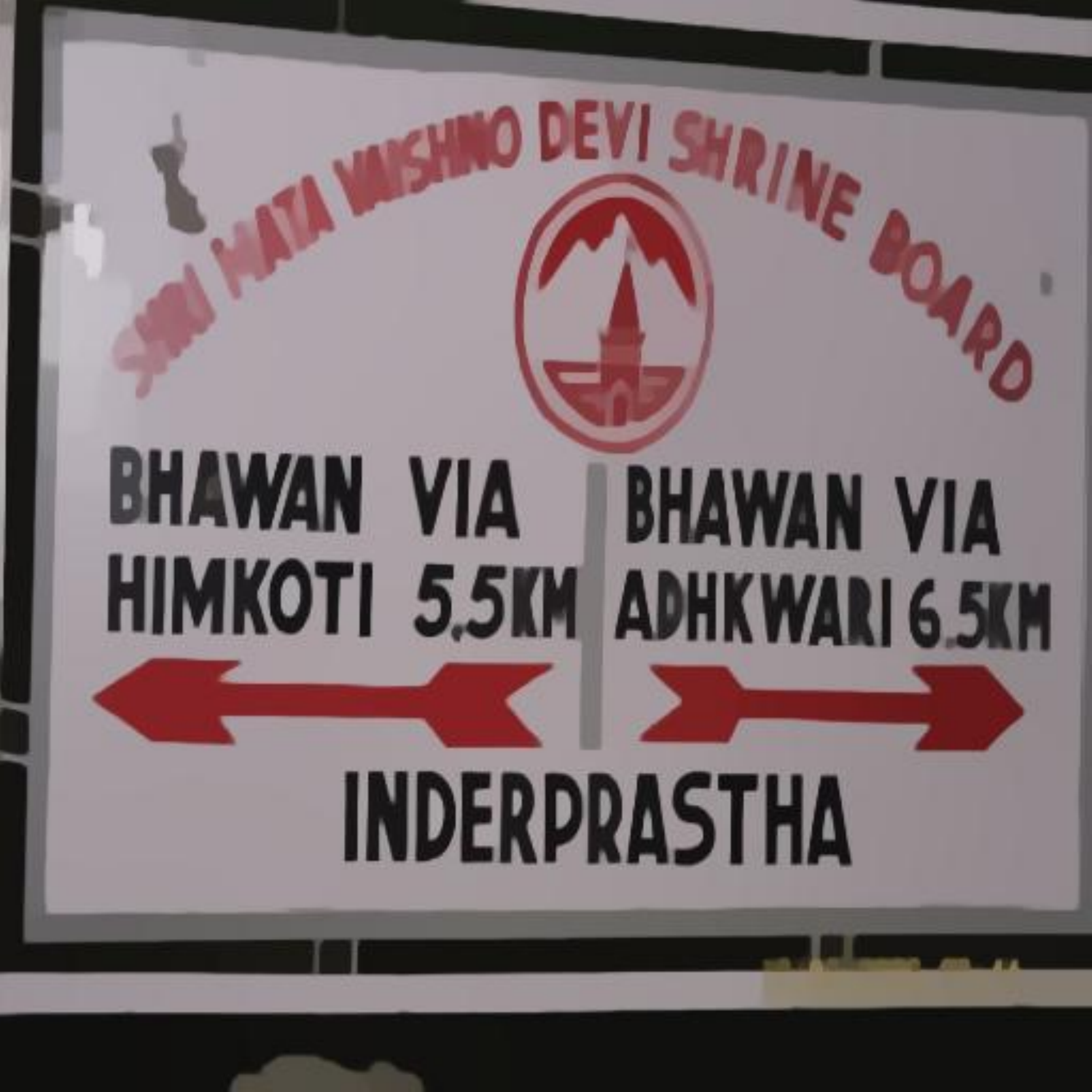}\\
		\end{minipage}%
	}
	\subfigure[$S_{gt}$]{
		\begin{minipage}[t]{0.14\linewidth}
			\centering
			\includegraphics[width=1.9cm,height=2cm]{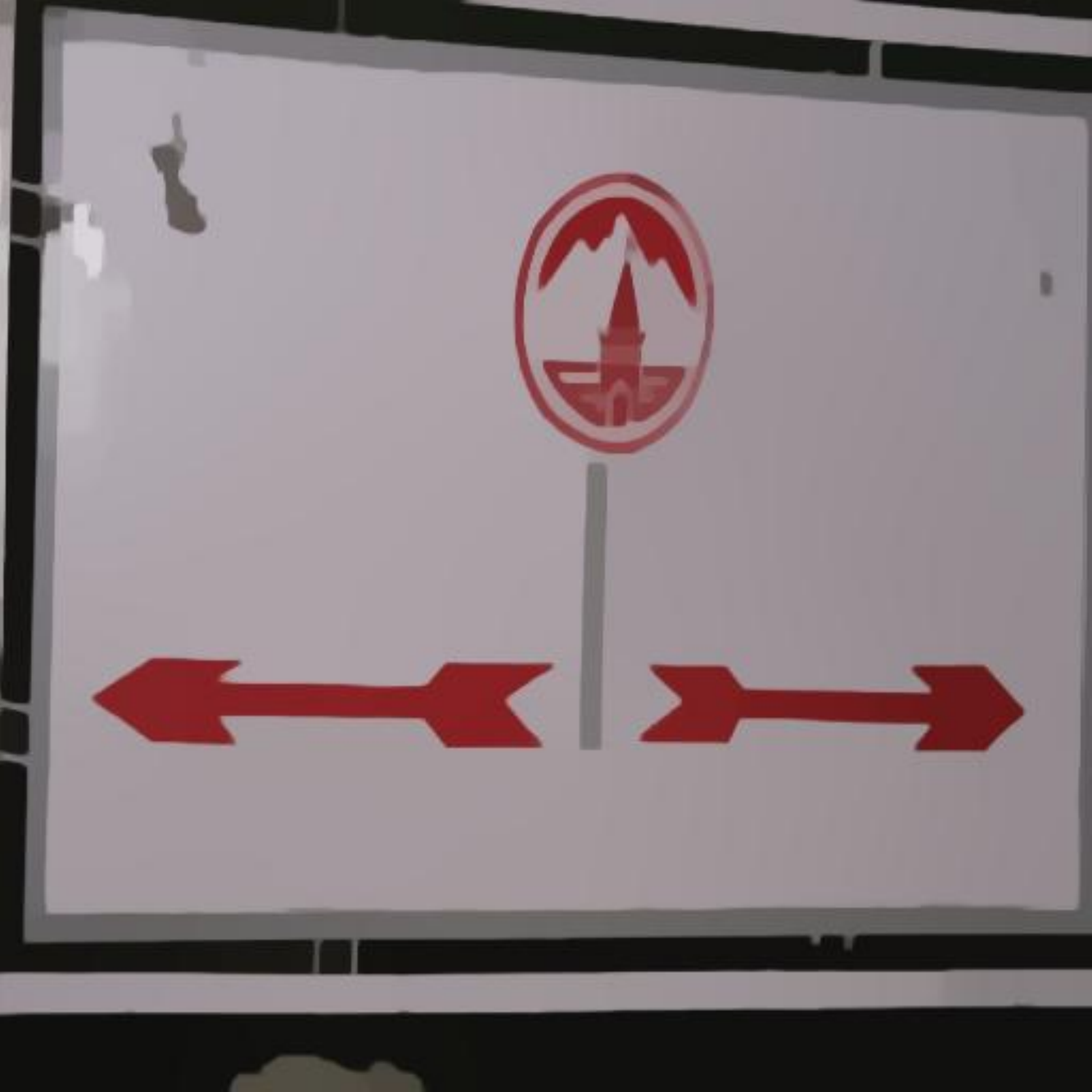}\\
		\end{minipage}%
	}
	\subfigure[GT]{
		\begin{minipage}[t]{0.14\linewidth}
			\centering
			\includegraphics[width=1.9cm,height=2cm]{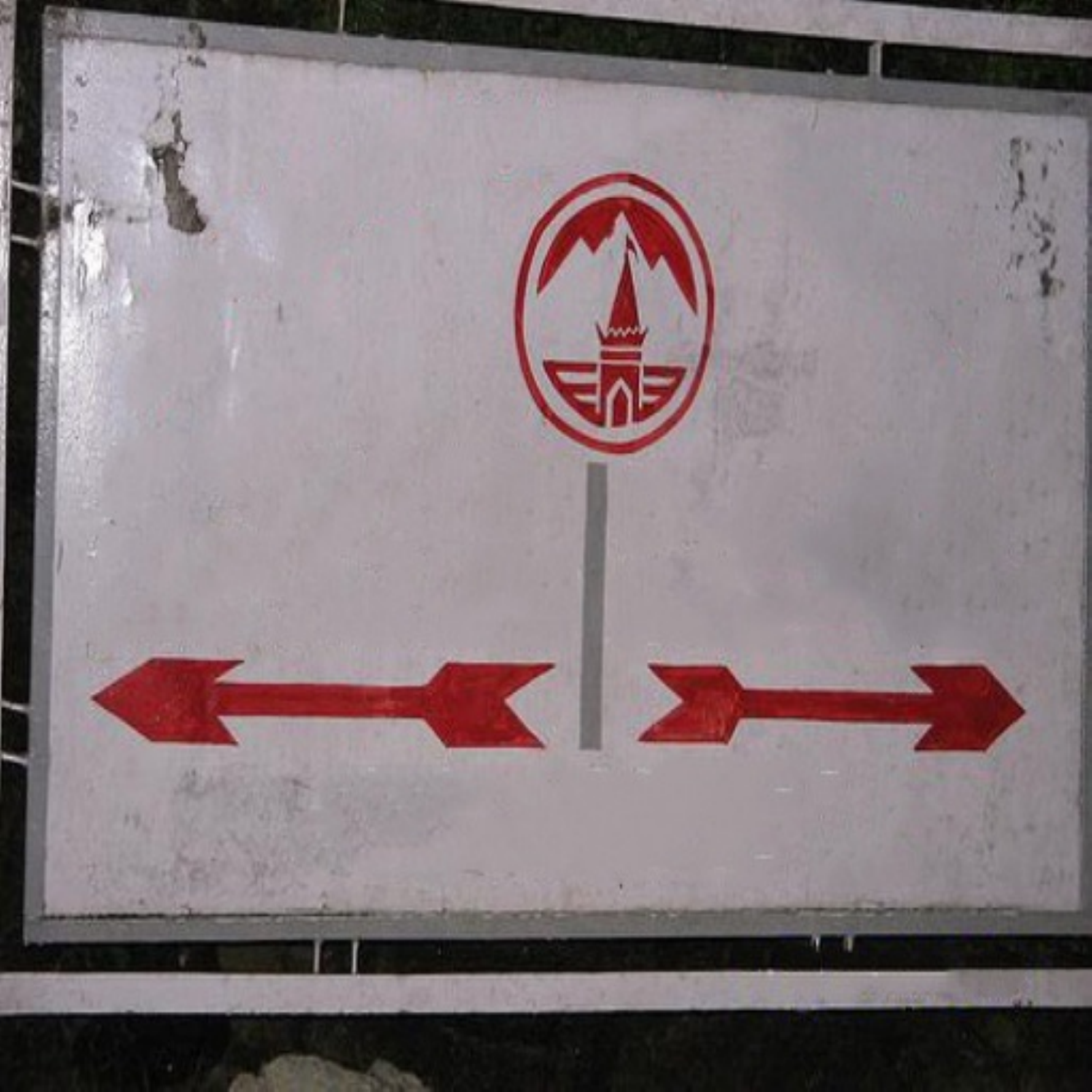}\\
		\end{minipage}%
	}
	\centering
	\caption{The basic elements of CTRNet. HM and SM denote hard mask and soft mask, respectively. $S_{in}$ and $S_{gt}$ represent the structure of the input and ground-truth. Zoom in for best view.} \label{fig:label}
\end{figure}

\noindent \textbf{Low-level Contextual Guidance (LCG) block:}
Scene text removal aims to not only erase the text, but also restore the backgrounds of text regions and synthesize their corresponding textures. Previous methods~\cite{zhang2019ensnet,liu2020erasenet,tursun2020mtrnet++,wang2021simple} follow an end-to-end training and inference procedure by directly predicting the results with scene text images as input. However, they suffer from some texture artifacts when dealing with complicated backgrounds, as shown in Fig.~\ref{fig:scut-enstext} and \ref{fig:sota_ens}. We propose to first predict the low-frequency structure of the image, and take it as low-level guidance for the subsequent network. Inspired by Ren et al. \cite{ren2019structureflow} and Liu et al. \cite{liu2020rethinking}, the structure image is constructed by the edge-preserved smooth method RTV \cite{xu2012structure}, which removes high-frequency textures with only sharp edges and smooth structure remain.
RTV consists of a pixel-wise windowed total variation measure and a windowed inherent variation to remove image texture. 
 Fig. \ref{fig:label} (d) and (e) display an example of the structure image $S_{in}$ and its ground-truth $S_{gt}$ generated from $I_{in}$ and $I_{gt}$, respectively. 
Learning a mapping between two low-frequency structures, $S_{in}$ and $S_{gt}$, is much easier than removing text directly. The structural clues for text regions can effectively simplify texture generation and enhance the performance by indicating the structure semantic of text regions, as shown in Fig. \ref{fig:scut-enstext} (e) (f) in the ablation study. 

As shown in Fig. \ref{fig:model}, LCG block consists of RTV method and a background structure generator $G_{bg\_s}$. 
$G_{bg\_s}$ is an encoder-decoder architecture that takes both the structure $S_{in}$ of scene text images and the soft mask $M_{s}$ as input, and predicts the background structure $S_{out}$ with text-erased. We take $S_{out}$ as local guidance, and directly feed it into the image encoder with $I_{in}$ to encode image features $F_{s} \in \mathbb R^{\frac{H}{4} \times \frac{W}{4} \times C}$. 

\noindent \textbf{High-level Contextual Guidance (HCG) block:}
In addition to the low-level structure priors,  we propose to explore potential high-level contextual guidance in latent feature spaces. 
Previous study~\cite{liu2020rethinking,ren2019structureflow,liu2020erasenet} with Perceptual/Style Loss~\cite{johnson2016perceptual,style}   
demonstrates the effectiveness of high-level contextual supervision for image generation and translation. Therefore, we make our CTRNet to utilize such discriminative context as additional guidance information for both text removal and background restoration, instead of taking it merely as supervision for optimization. Inspired by Zhang et al. \cite{SPL}, we incorporate an HCG block into our CTRNet to learn high-level context features.

The architecture of HCG block is illustrated in the left-bottom of Fig. \ref{fig:model}. The block consists of two feature encoders ($E_{c}(\cdot)$ and $E_{c}^{'}(\cdot)$), and a Feature Align Module (FAM), as done in \cite{SPL}. $E_{c}(\cdot)$ encodes the concatenation of the original image $I_{in}$ and its soft-mask $M_{s}$ to obtain the features $F_{hc} \in \mathbb R^{\frac{H}{4} \times \frac{W}{4} \times C}$, whereas $E_{c}^{'}(\cdot)$ extracts the context features $F_{hc}^{'} \in \mathbb R^{\frac{H}{4} \times \frac{W}{4} \times C}$ from the paired labels $I_{gt}$. Here, $E_{c}^{'}(\cdot)$ is a classification model, termed as TResNet \cite{ridnik2021asymmetric}. We directly use its pretrained model on the OpenImages datasets \cite{kuznetsova2020open} to extract $F_{hc}^{'}$ with frozen weights during the training procedure. After feature dimension mapping with $1 \times 1$ convolution layers in FAM,  feature align loss $L_{align}$ is applied to approximate the distribution of $F_{hc}$ to $F_{hc}^{'}$. 
The process can be formulated as 
\begin{equation}
	\begin{aligned}
			\label{fam}
			F_{hc}^{'} =  E_{c}^{'}(I_{gt}); F_{hc} =  E_{c}(I_{in}, M_{s})\\
			L_{align} = \left\|  F_{hc} - F_{hc}^{'} \right\|_{1} * (1 + \alpha M_{s})
		\end{aligned}
\end{equation} 
$\alpha$ is set to 2.0. In this way, $F_{hc}$ based on $I_{in}$ can be transferred to contain context information of background $I_{gt}$, which can provide a high-level global guidance for feature modeling and decoding.

\subsection{Local-global Content Modeling (LGCM)} 
While erasing text regions and filling them with reasonable textures as background, beyond considering text regions as a reference, it is necessary for a text removal method to use the pixel information from the surrounding and global backgrounds. Therefore, we propose a feature content modeling block for both local (text regions) and global (surrounding and the entire background) levels. 
As shown in Fig. \ref{fig:model}, the image content features $F_{s} \in \mathbb R^{\frac{H}{4} \times \frac{W}{4} \times C}$, incorporated with the high-level discriminative feature guidance, $F_{hc}$ are sent to LGCM to model the local-global contextual features and enhance their representations. And the right-bottom of Fig. \ref{fig:model} displays the architecture of a single LGCM block.

CNNs operate locally at a fixed size (e.g. 3$\times$3) to effectively extract features of specific regions and establish the relationship between the pixels in each local window. Therefore, four stacked vanilla $4 \times 4$ convolutions layers are utilized for local content modeling. 
In addition, features can be downsampled by CNNs to reduce the computation required for the subsequent global modeling operation.
For global content modeling, we apply Transformer-Encoder as our basic module. Transformer-Encoder \cite{vaswani2017attention}, which can effectively capture global interactions between pixels among the whole features and model their long-range dependency. Then two deconvolution layers are applied to upsample the modeled features and bring the inductive bias of CNN \cite{liang2021swinir}. 
LGCM follows an iterative process with $k$ stages ($k=8$ empirically \cite{SPL}). 
At the final convolution of each stage, $F_{hc}$ are incorporated into the LGCM with ResSPADE \cite{park2019semantic,SPL}. 
The details for LGCM and ResSPADE are presented in supplement materials. The output of the $i-th$ LGCM is denoted as $F_{l_{i}}$ ($F_{l_{0}} = F_{s}$).

Finally, Feature Decoder reconstructs the final text-erased output by decoding both features $F_{l_{8}}$ from the final LGCM ($8th$) block and shadow content features $F_{s}$.
, which can be formulated as 
\begin{equation}
	\label{feature_decoder}
	I_{out} = H_{fd}(F_{l_{8}} + F_{s})
\end{equation}

\subsection{Training Objective} 

We adopt the following losses to train our text removal network, including structure loss,  multi-scale text-aware reconstruction loss, perceptual loss, style loss, and adversarial loss.

\noindent \textbf{Structure loss} The structure loss is used to measure the $L_{1}$ distance between the background structure output $ S_{out} $  and the ground truth $ S_{gt}$, which is defined by:
\begin{equation}
	\label{structure}
	L_{str} =  \left\|  S_{gt} - S_{out}\right\|_{1} * (1 + \gamma M_{s})
\end{equation} 
\noindent $(1 + \gamma M_{s}) $ denotes higher weight for text region. $\gamma$ is set to 3.0. 

\noindent \textbf{Multi-scale text-aware reconstruction loss} The $L_{1}-norm$ difference is proposed to measure the output and the ground truth. We first predict multi-layer outputs with text removed in different sizes, then assign higher weight to text regions when computing the loss:
\begin{equation}
	\begin{aligned}
		\label{msr}
		L_{msr} = \sum_{n} \left\|  (I_{out_{n}} - I_{gt_{n}})  \right\|_{1} * (1 + \theta_{n} M_{s})
	\end{aligned}
\end{equation}
\noindent $n$ denotes $n$-th output in the scales of $\frac{1}{16}$, $\frac{1}{4}$ and 1 of the input.  $\theta_{1}, \theta_{2}, \theta_{3}$ is set to ${2,3,4}$, respectively. 

\noindent	\textbf{Perceptual loss}  Except for low-level image-to-image supervision with reconstruction loss, we also adopt perceptual loss \cite{johnson2016perceptual} to capture high-level semantics and try to simulate human perception of image quality. Both the straight output $I_{out}$ and the original image with text-removed $I_{com}$ are included as loss terms. Besides, the structure output $ S_{out} $ is also taken into consideration.
 \begin{equation}
 \setlength{\belowdisplayskip}{3pt}
	\begin{aligned}
		\label{peloss1}
		I_{com} = I_{in} * (1 - M_{s}) + I_{out} * M_{s}
	\end{aligned}
\end{equation} 
 \begin{equation}
	\begin{aligned}
		\label{peloss}
		L_{per} = \sum_{i} \sum_{j} \left\|  {\phi}_{j}(I_{i}) - {\phi}_{j}(I_{gt})\right\|_{1} 
		+ \sum_{j} \left\|  {\phi}_{j}(S_{out}) - {\phi}_{j}(S_{gt})\right\|_{1} 
	\end{aligned}
\end{equation} 
\noindent where $I_{i}$ represent $I_{out}$ and $I_{com}$. ${\phi}_{j}(.)$ denotes the activation maps of the $j$-th ($j =1,2,3$) pooling layer of VGG-16 pretrained on ImageNet \cite{deng2009imagenet}.

\noindent \textbf{Style loss} We also utilize style loss to release the artifacts of the generated results. Style loss \cite{style} construct a Gram matrix $Gr(.)$ from each selected activation map in perceptual loss. Style loss can be defined as 
 \begin{equation}
		\begin{aligned}
				\label{styloss}
				L_{style} = \sum_{i} \sum_{j} \frac {\left\|  {Gr}_{j}(I_{i}) - {Gr}_{j}(I_{gt})\right\|_{1}}{H_{j} W_{j} C_{j}} 
				+ \sum_{j} \frac {\left\|  {Gr}_{j}(S_{out}) - {Gr}_{j}(S_{gt})\right\|_{1}}{H_{j} W_{j} C_{j}}
			\end{aligned}
	\end{equation}

\noindent \textbf{Adversarial loss}
The adversarial loss encourages our model to generate more plausible details for the final results with text removed. Here we defined our adversarial loss as:
 \begin{equation}
	\begin{aligned}
		\label{ganloss}
		L_{adv} = E_{x\sim {P}_{\text {data}}(x)}\left [\log D(x) \right ] + E_{x\sim {P}_{\text {z}(z)}}\left [\log\left (1 - D(G(z)) \right ) \right ] 
	\end{aligned}
\end{equation} 
$z$ is the input $I_{in}$ and $x$ represents the corresponding ground-truth $I_{gt}$.

\noindent \textbf{Total loss} The overall loss function for our text removal network is defined as:
 \begin{equation}
	\begin{aligned}
		\label{styloss}
		L_{total} = \lambda_{al} L_{align} + \lambda_{str}L_{structure} + \lambda_{m}L_{msr} \\
		+ \lambda_{p}L_{per} +\lambda_{s}L_{style} + \lambda_{a}L_{adv} 
	\end{aligned}
\end{equation} 
\noindent $\lambda_{al}, \lambda_{str}, \lambda_{m}, \lambda_{p}, \lambda_{s}, \lambda_{a}$ are the trade-off parameters.
In our implementation, we empirically set $\lambda_{al}=1.0, \lambda_{str}=2.0, \lambda_{m}=10.0, \lambda_{p}=0.01, \lambda_{s}=120, \lambda_{a}=1.0$.

\section{Experiments}

\subsection{Datasets and Evaluation Metrics}
\noindent{\textbf{Datasets}}
To evaluate the effectiveness of our proposed CTRNet, we conduct experiments on the two widely used benchmarks, SCUT-Syn \cite{zhang2019ensnet} and SCUT-EnsText~\cite{liu2020erasenet}. 

\noindent \textbf{(1) SCUT-Syn}: 
SCUT-Syn contains a training set of 8,000 images and a testing set of 800 images. It is a synthetic dataset with \cite{gupta2016synthetic}. The background images are mainly collected from ICDAR-2013 \cite{ic13} and MLT-2017 \cite{nayef2017icdar2017}, and the text instances are manually erased.

\noindent \textbf{(2) SCUT-EnsText}: 
SCUT-EnsText is a comprehensive real-world dataset with 2,749 images for training and 813 images for testing. These images are collected from public scene text benchmark, including
ICDAR-2013 \cite{ic13}, ICDAR-2015 \cite{karatzas2015icdar}, MS COCO-Text \cite{veit2016coco},
SVT \cite{svt}, MLT-2017 \cite{nayef2017icdar2017}, MLT-2019 \cite{mlt2019}, and ArTs \cite{chng2019icdar2019},
which consists of SCUT-CTW1500 \cite{yuliang2017detecting} and Total-Text \cite{ch2017total}. All the images are carefully annotated with Photoshop.

~\\
\noindent{\textbf{Evaluation metrics:}}
To comprehensively evaluate the performance of our CTRNet, we utilize both Image-Eval and Detection-Eval as used in EraseNet \cite{liu2020erasenet}. (1) Image-Eval includes the following metrics for image quality evaluation.  (1) Peak signal
to noise ratio (PSNR); (2) Multi-scale Structural Similarity (MSSIM); (3) Mean Square Error (MSE); (4) Fréchet Inception Distance (FID) \cite{heusel2017gans}.
A higher PSNR, MSSIM and lower MSE, FID denotes better results. (2) Detection-Eval evaluates the Recall (R), Precision (P), F-measure (F), TIoU-Recall (TR), TIoU-Precision (TP), and TIoU-F-measure (TF) for the results under the protocols of ICDAR 2015 \cite{karatzas2015icdar} and T-IoU \cite{liu2019tightness}. CRAFT \cite{Baek2019Character} is served as the text detector for evaluation. The lower R, P and F indicate that more text can be removed.

\subsection{\textbf{Implement details}}
We utilize Pixel Aggregation Network (PAN)~\cite{wang2019efficient} as text perception head for CTRNet. The input size is set to $512 \times 512$. Adam solver \cite{kingma2014adam} is used to optimized our method, and the $\beta$ is set to (0.0, 0.9) as default.The batch size is set to 2. All experiments are conducted on a workstation with two NVIDIA 2080TI GPUs. More training details and the architectures of each component are provided in the supplementary materials.

\begin{table*}[t]
	\caption{Ablation Study on SCUT-EnsText. MSSIM and MSE are represented by $\%$ in the table.}
	\label{table:aba}
	\centering
	\newcommand{\tabincell}[2]{\begin{tabular}{@{}#1@{}}#2\end{tabular}}
	\small
	\begin{tabular}{c|cccc|c|c|c|c|c|c|c|c}
		\hline \Xhline{0.3pt}
		\multirow{2}*{ }  & \multicolumn{4}{c|}{Components} & \multicolumn{4}{c|}{Evaluation on $I_{out}$} &  \multicolumn{4}{c}{Evaluation on $I_{com}$}\\
		\cline{2-13}
		& HCG & LGCM & SM & LCG  & PSNR & MSSIM & MSE & FID & PSNR & MSSIM & MSE & FID \\
		\Xhline{0.3pt}
		\hline
		baseline &  - & - & - & - & 32.39 & 95.45 & 0.13 & 20.75 & 33.21 & 95.52 & 0.11 & 22.15 \\
		\hline
		Ours+  & $\checkmark$ &   &  & & 32.90 & 96.62 & 0.11 & 17.40 & 34.88 & 97.09 & 0.10 & 19.42   \\
		Ours+  & $\checkmark$ & $\checkmark$  &  & & 35.10  & 97.36 & 0.09 &  14.36 & 35.30 & 97.20 & 0.09 & 17.91  \\
		Ours+  & $\checkmark$ &  $\checkmark$ & $\checkmark$ & & 35.16 &  \textbf{97.38} & 0.09 & 14.33 &  35.83 & \textbf{97.42} & 0.09 &  15.02   \\
		Ours+  & $\checkmark$ &  $\checkmark$ & $\checkmark$ & $\checkmark$ & \textbf{35.20} & 97.36 &  \textbf{0.09}  &  \textbf{13.99}  &  \textbf{35.85} &   97.40  &  \textbf{0.09} &  \textbf{14.57}     \\ \hline \Xhline{0.3pt}
	\end{tabular}
\end{table*}

\begin{figure*}[t]
	\subfigbottomskip=2pt
	\subfigcapskip=2pt
	\setlength{\abovecaptionskip}{-0.0cm}
	\setlength{\belowcaptionskip}{-0.4cm} 
	\centering
	\subfigure[]{
		\begin{minipage}[t]{0.127\linewidth}
			\centering
			\includegraphics[width=1.6cm,height=1.8cm]{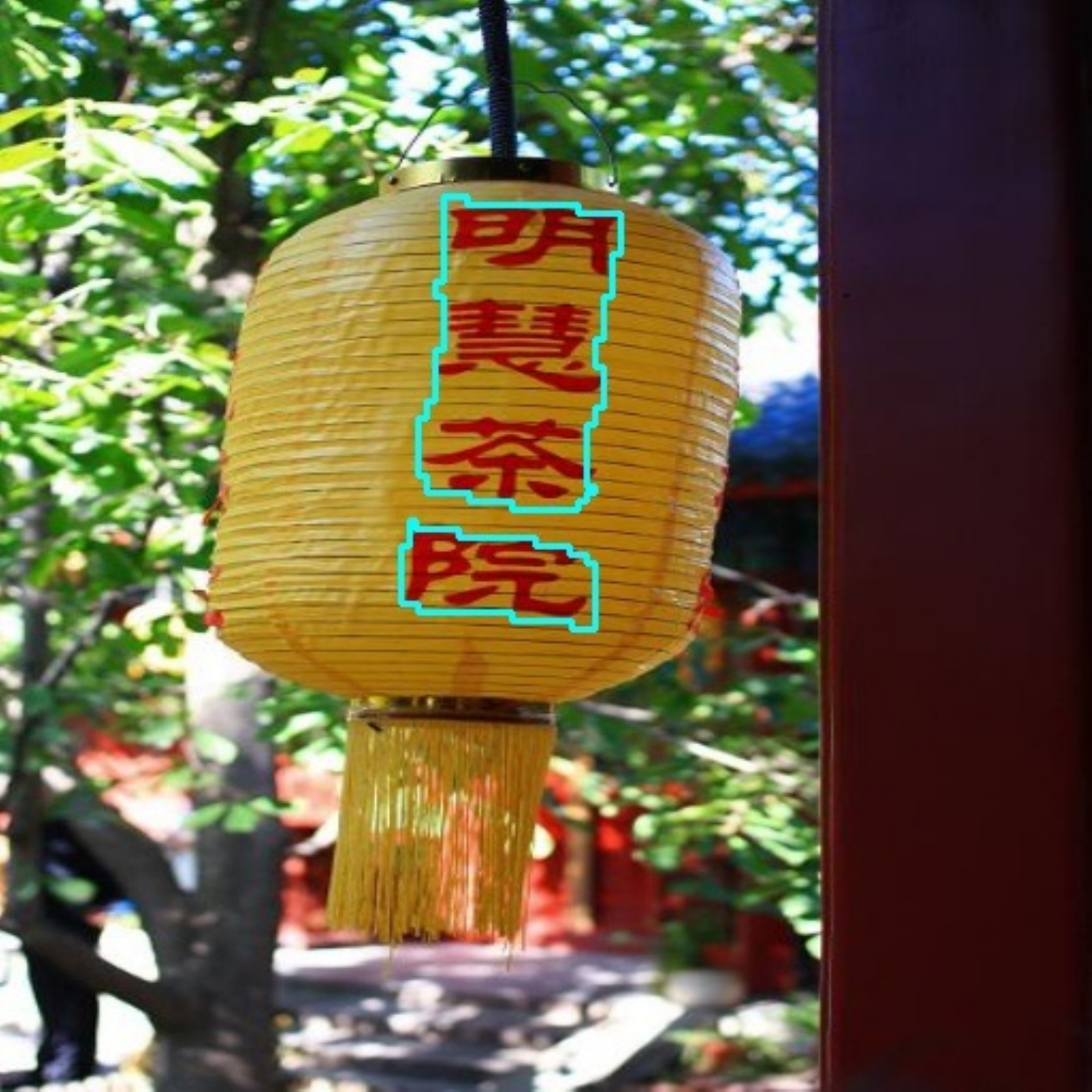}\\
			\includegraphics[width=1.6cm,height=1.8cm]{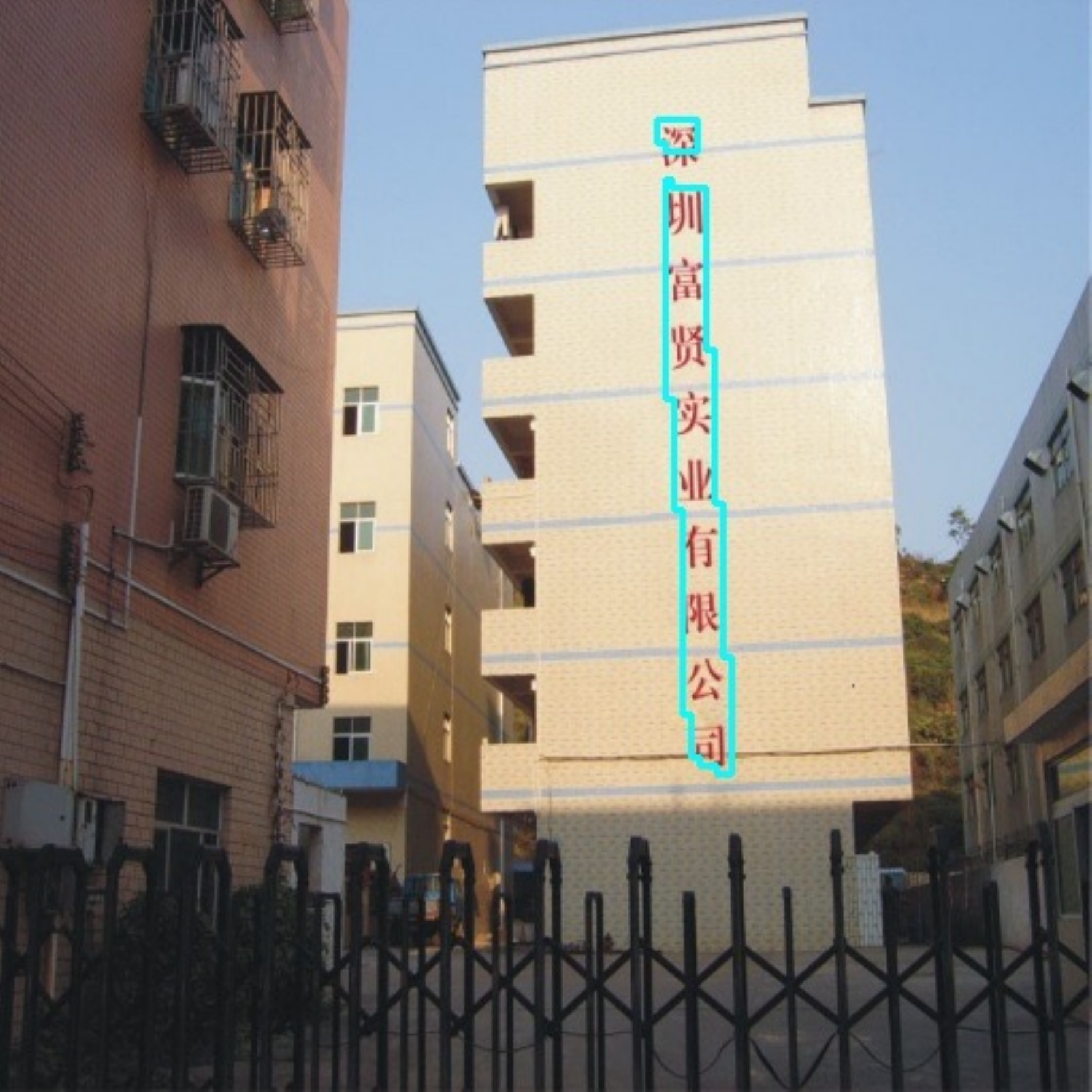}\\
			\includegraphics[width=1.6cm,height=1.8cm]{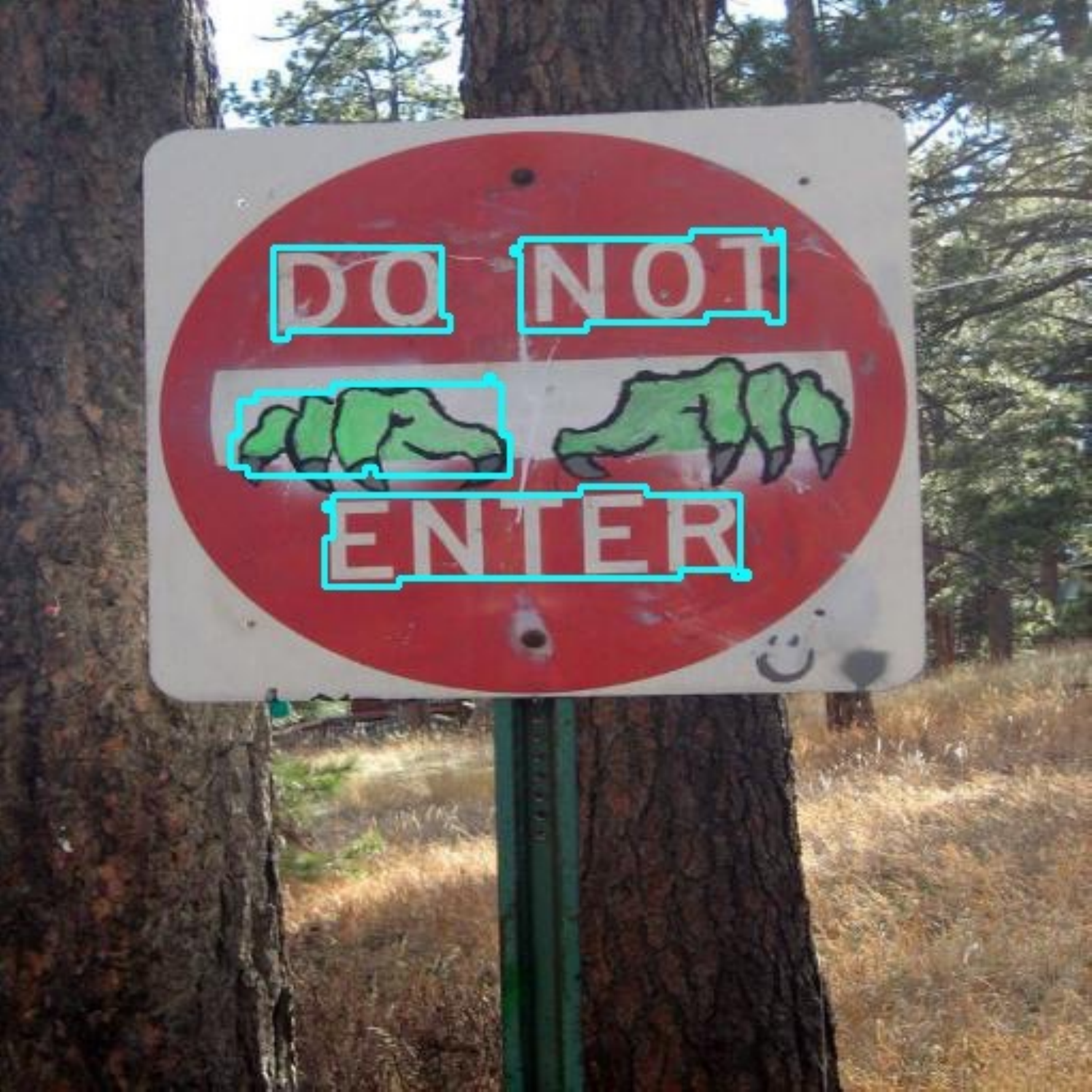}\\
		\end{minipage}%
	}%
	\subfigure[]{
		\begin{minipage}[t]{0.121\linewidth}
			\centering
			\includegraphics[width=1.6cm,height=1.8cm]{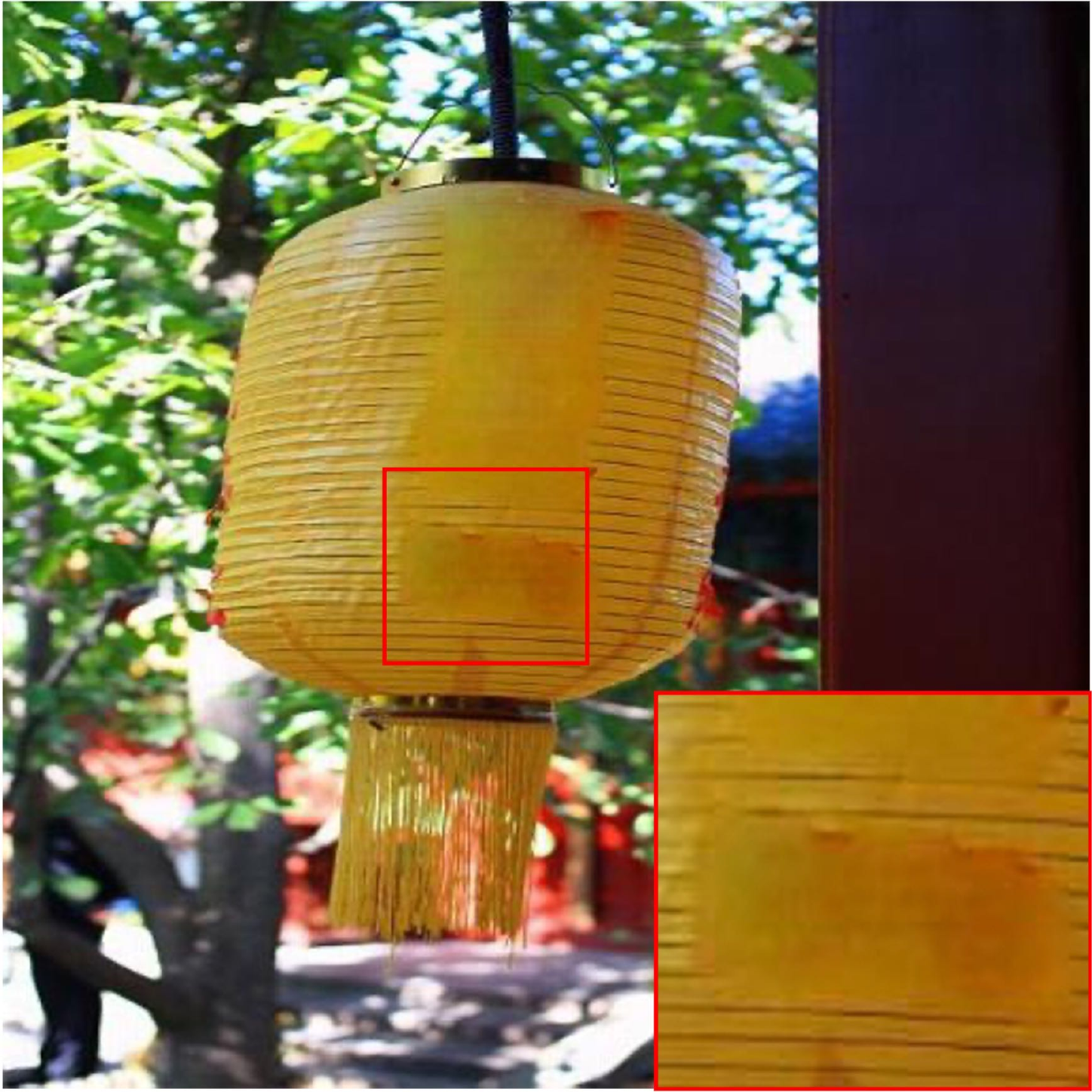}\\
			\includegraphics[width=1.6cm,height=1.8cm]{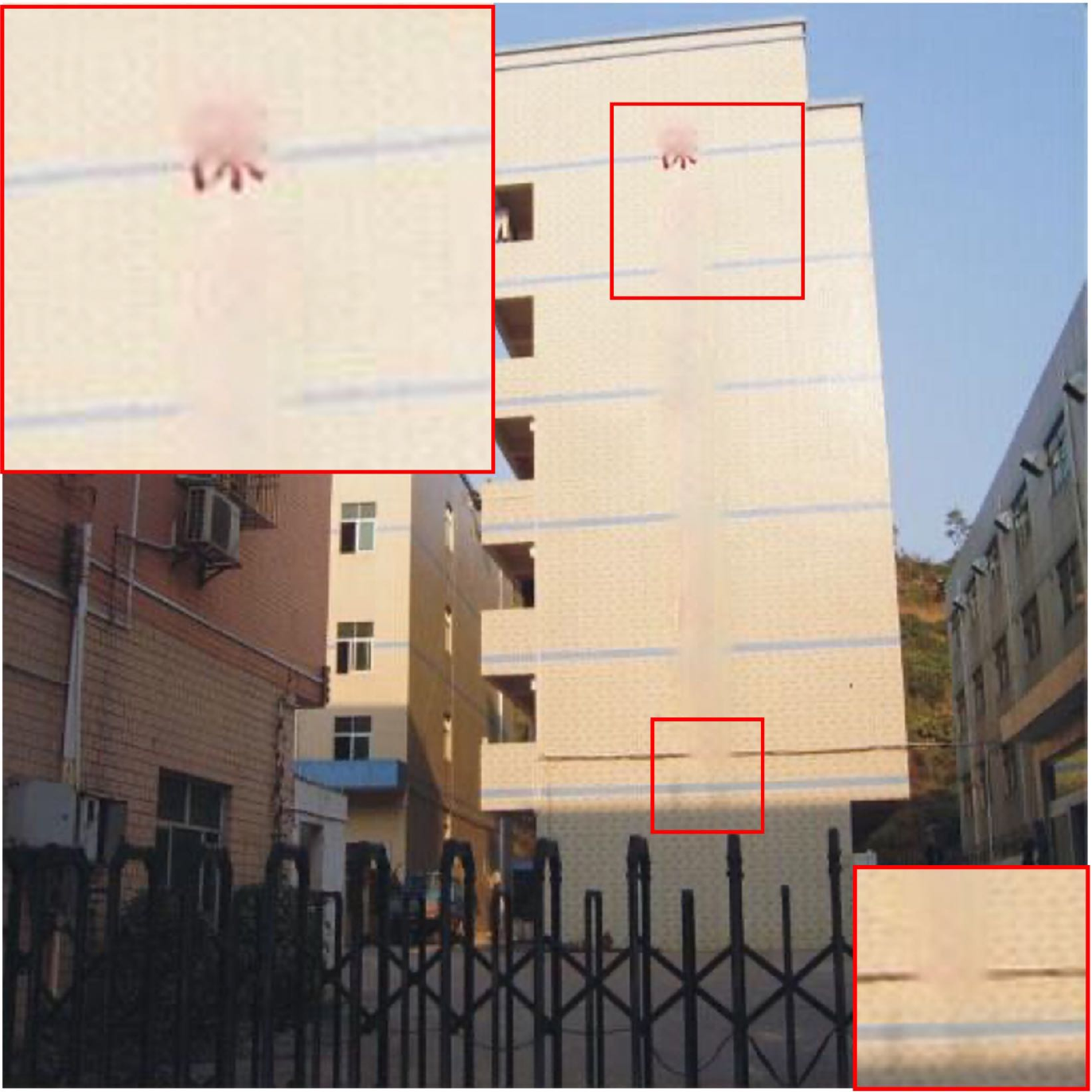}\\
			\includegraphics[width=1.6cm,height=1.8cm]{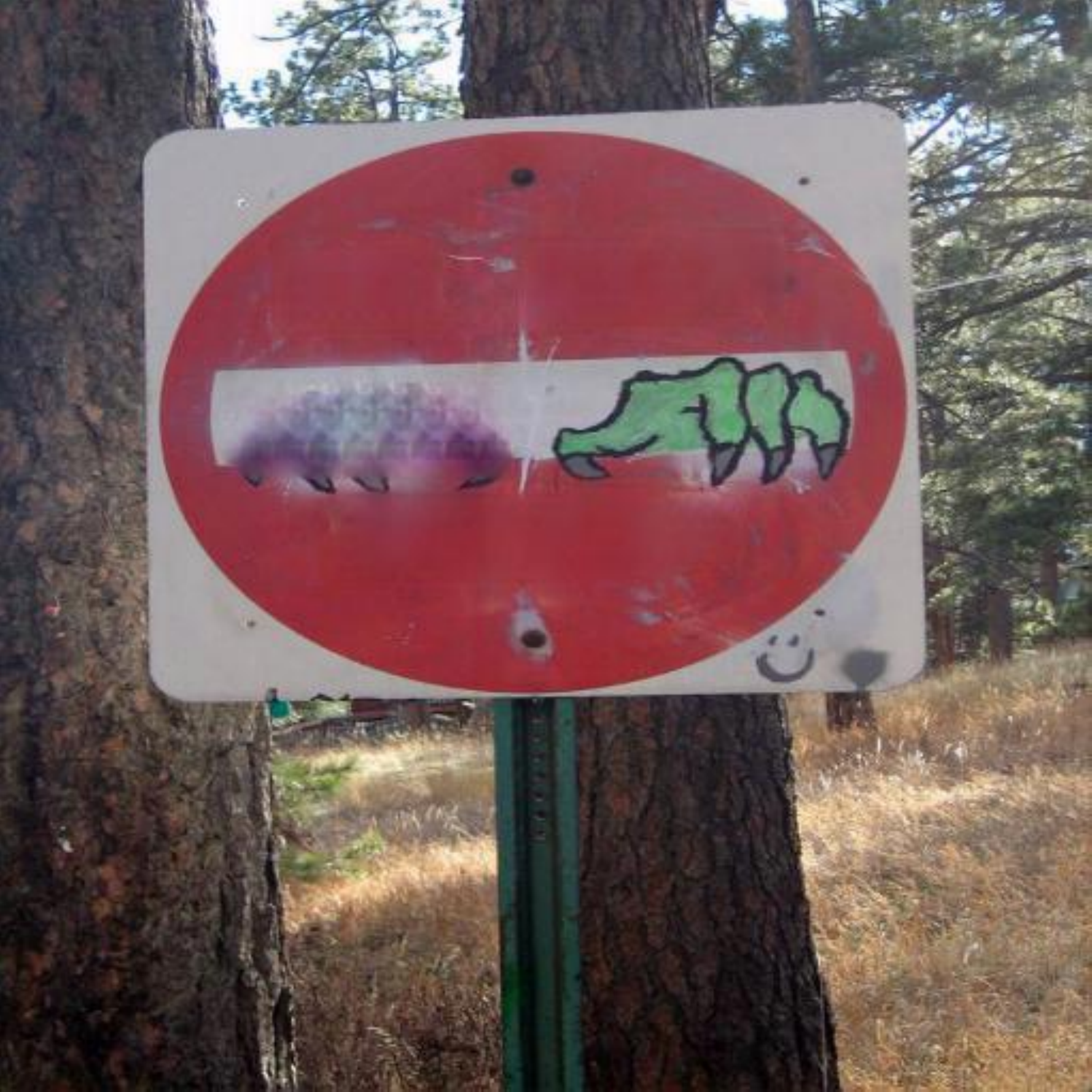}\\
		\end{minipage}%
	}
	\subfigure[]{
		\begin{minipage}[t]{0.121\linewidth}
			\centering
			\includegraphics[width=1.6cm,height=1.8cm]{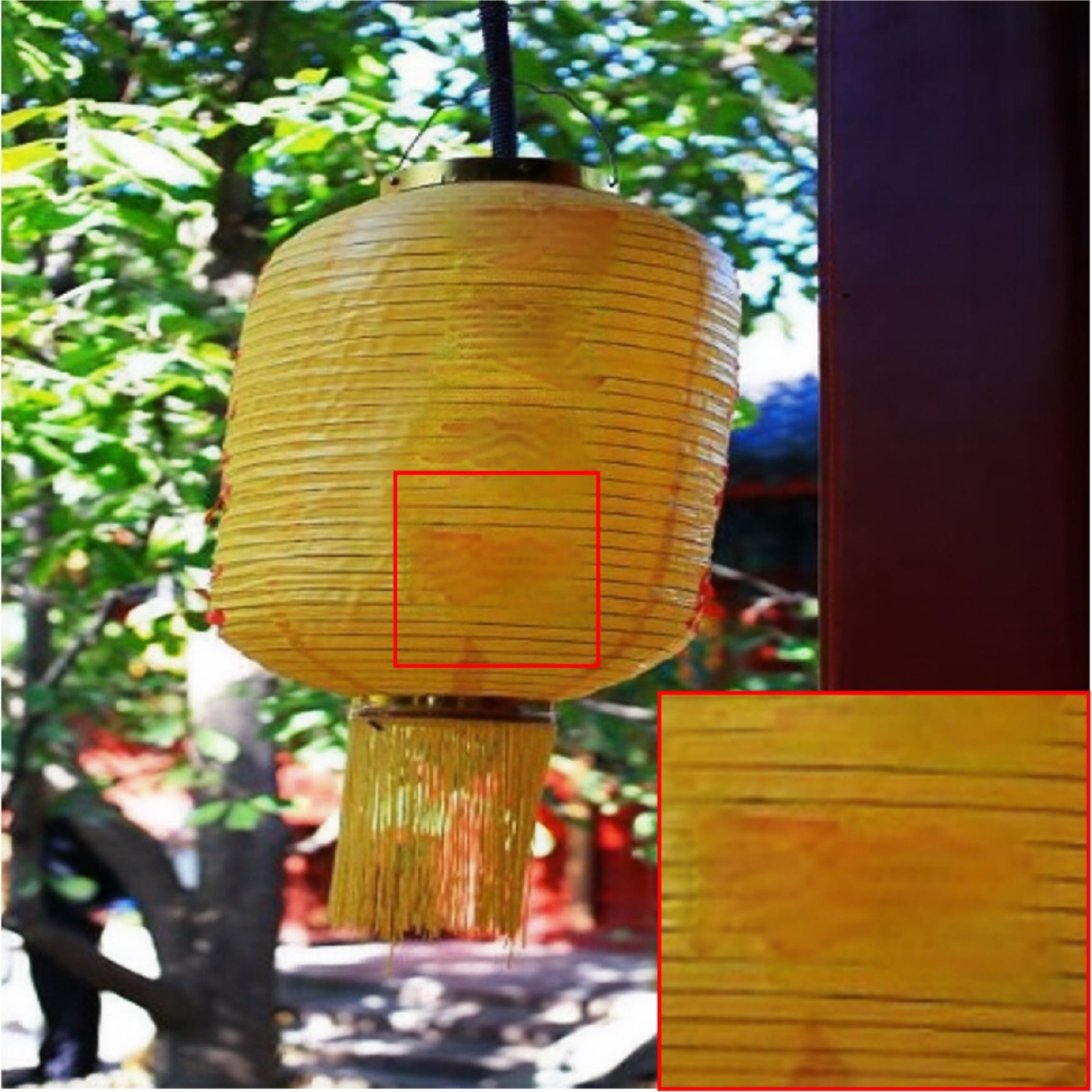}\\
			\includegraphics[width=1.6cm,height=1.8cm]{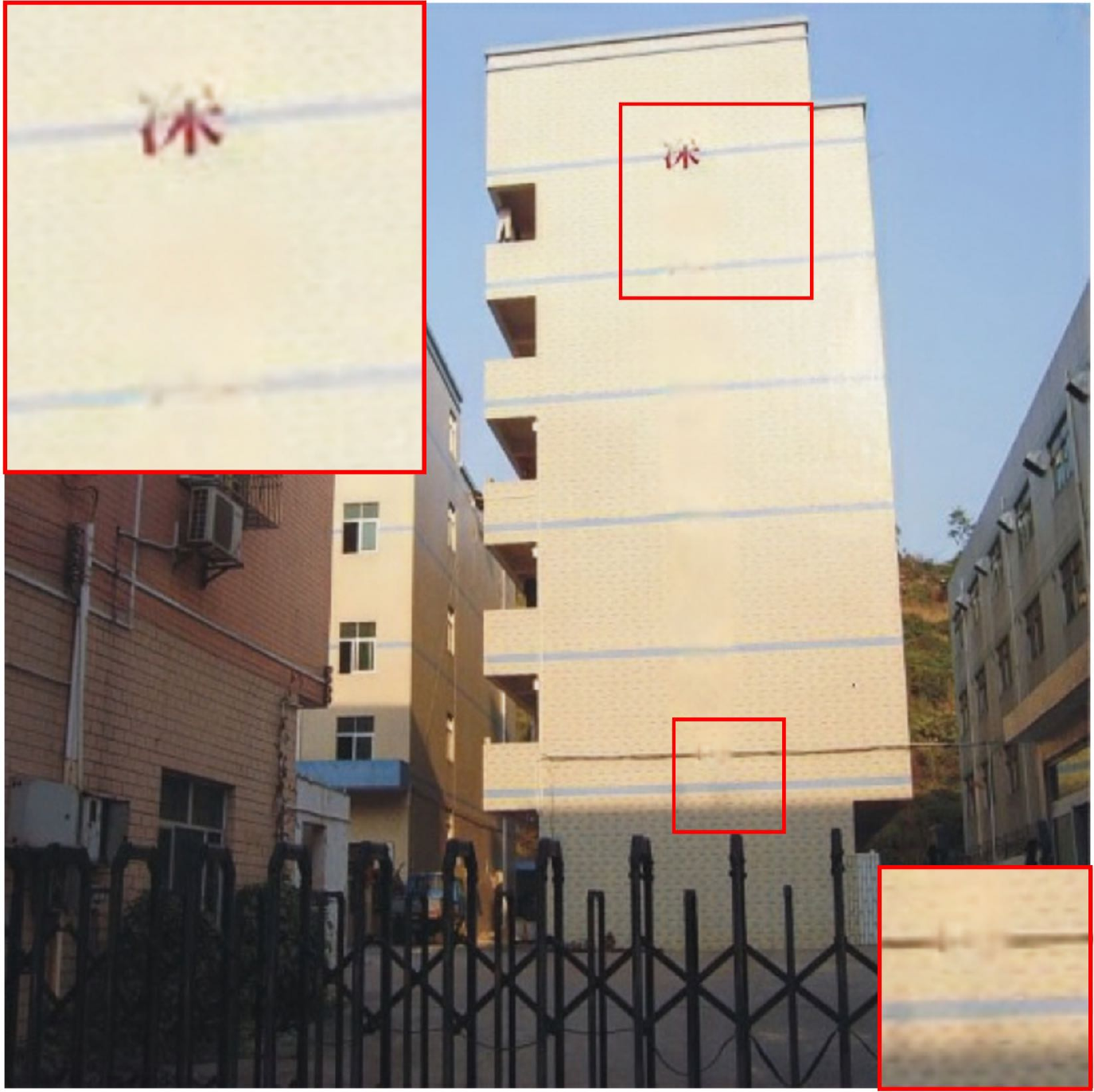}\\
			\includegraphics[width=1.6cm,height=1.8cm]{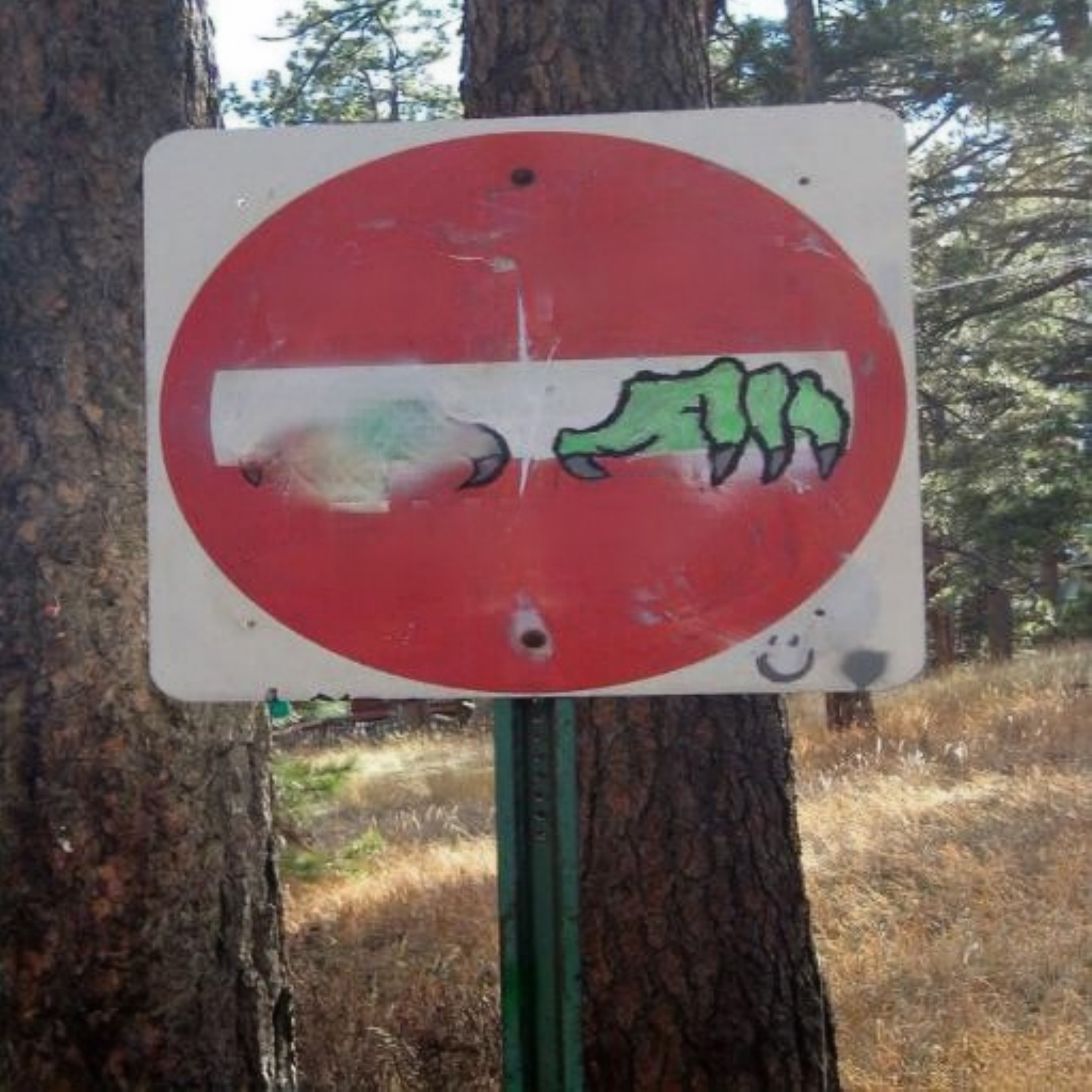}\\
		\end{minipage}%
	}
	\subfigure[]{
		\begin{minipage}[t]{0.121\linewidth}
			\centering
			\includegraphics[width=1.6cm,height=1.8cm]{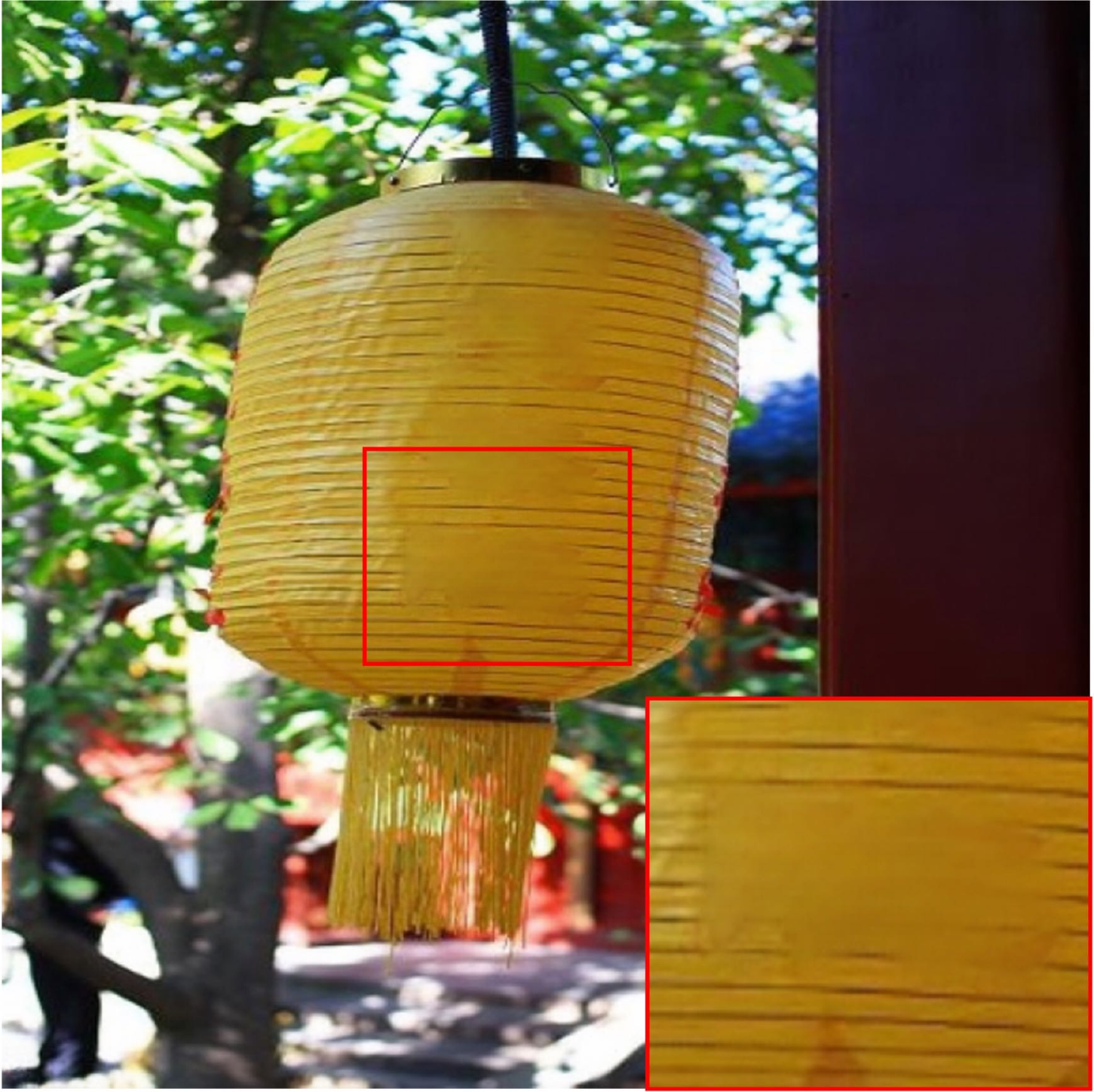}\\
			\includegraphics[width=1.6cm,height=1.8cm]{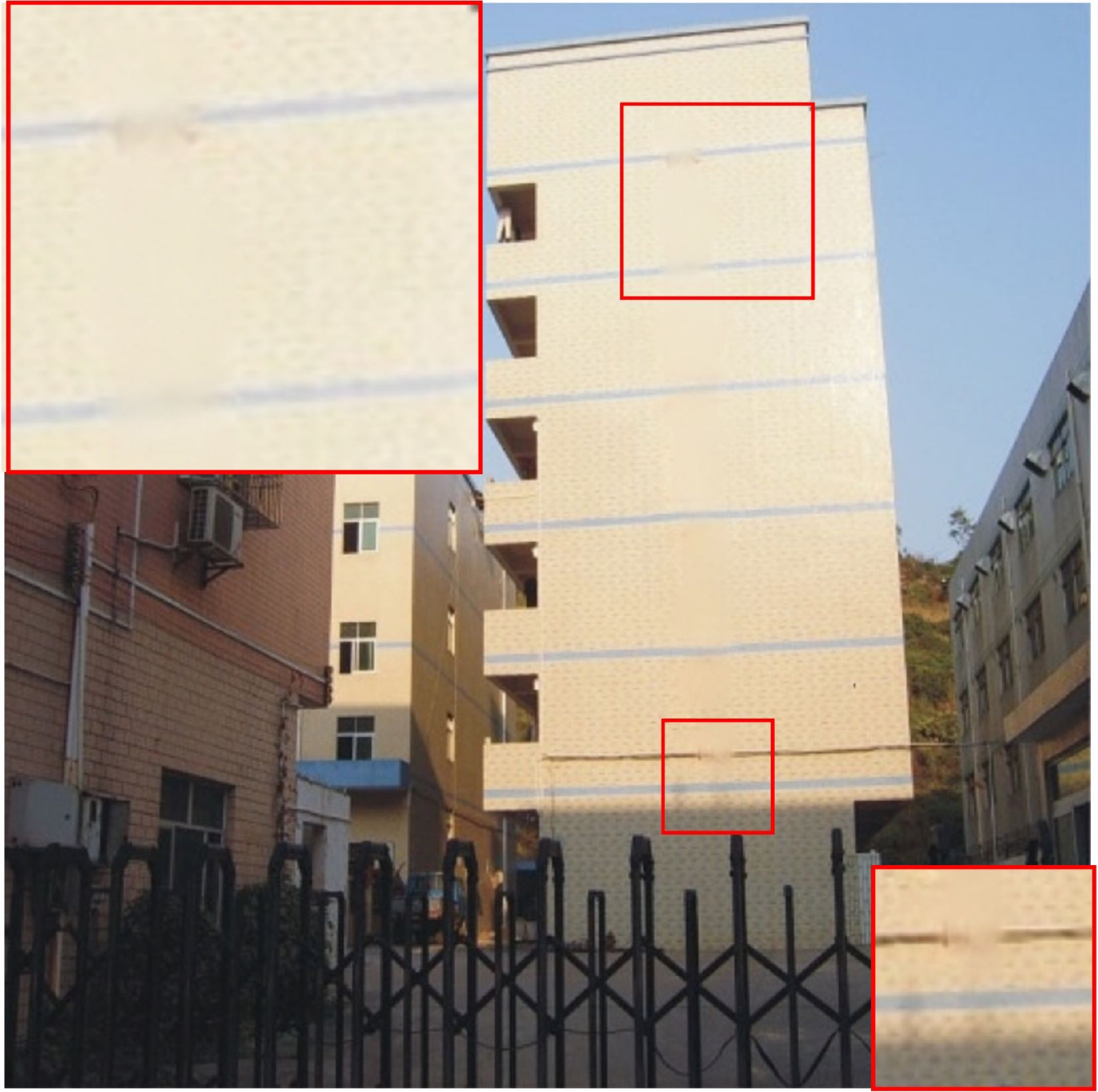}\\
			\includegraphics[width=1.6cm,height=1.8cm]{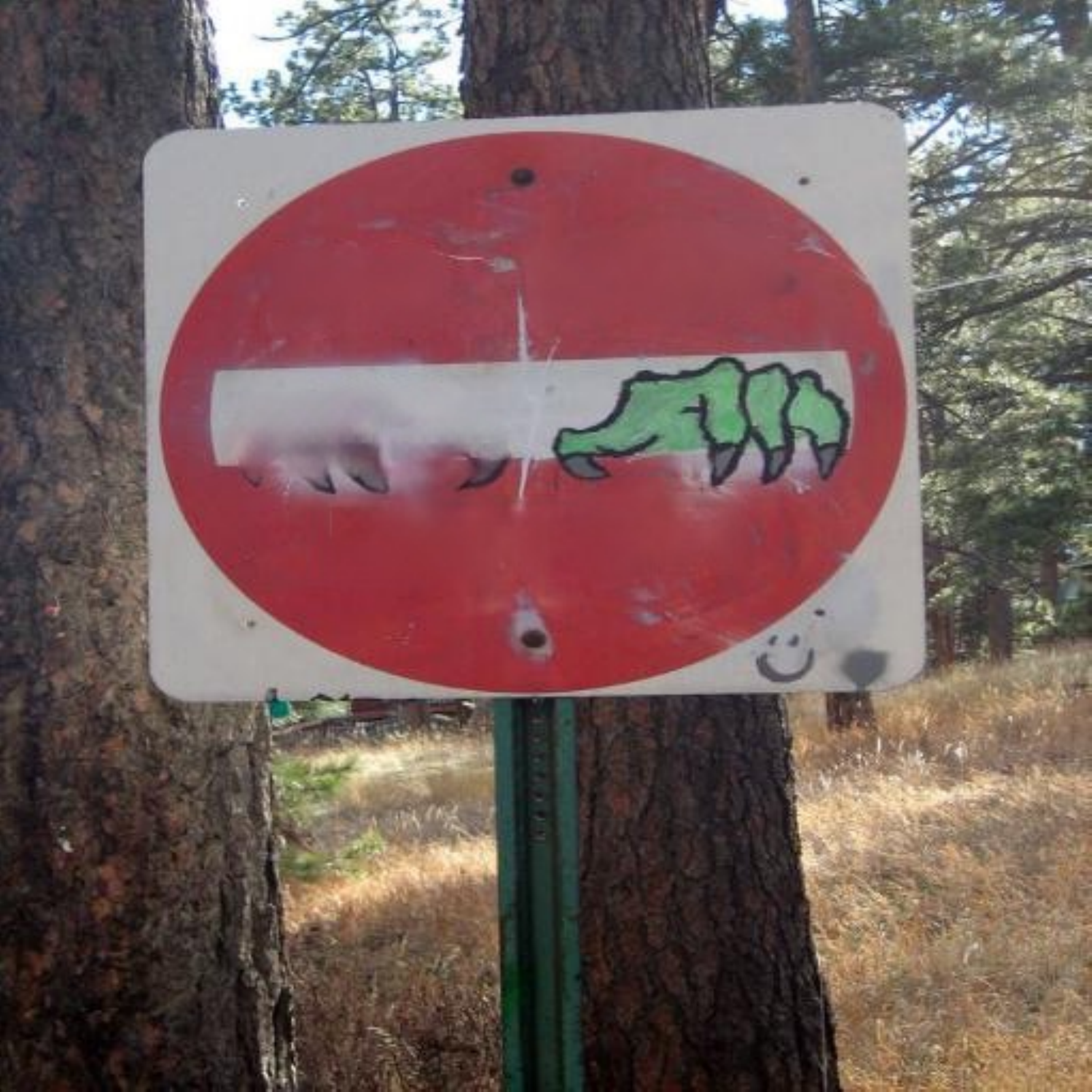}\\
		\end{minipage}%
	}
	\subfigure[]{
		\begin{minipage}[t]{0.121\linewidth}
			\centering
			\includegraphics[width=1.6cm,height=1.8cm]{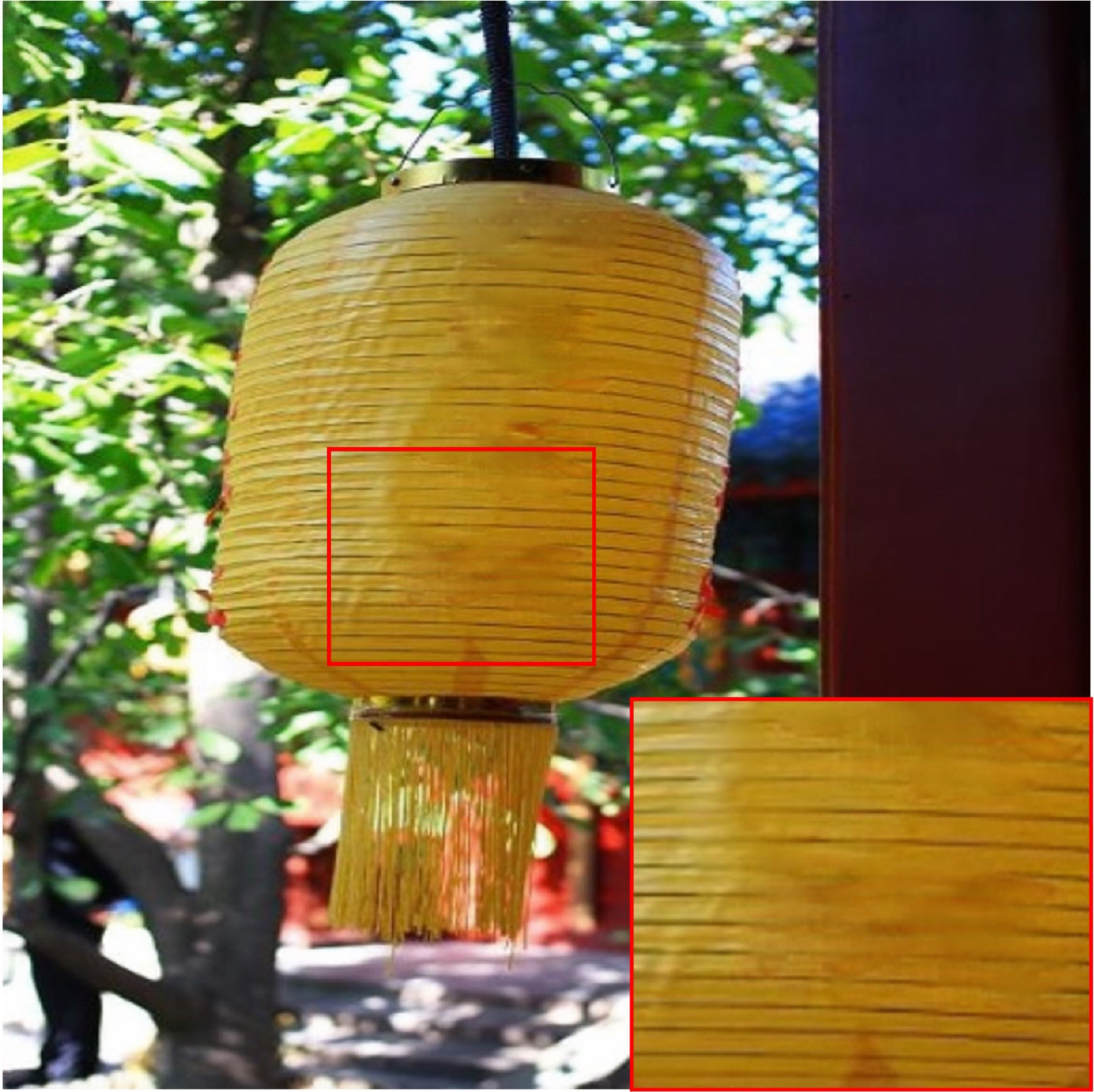}\\
			\includegraphics[width=1.6cm,height=1.8cm]{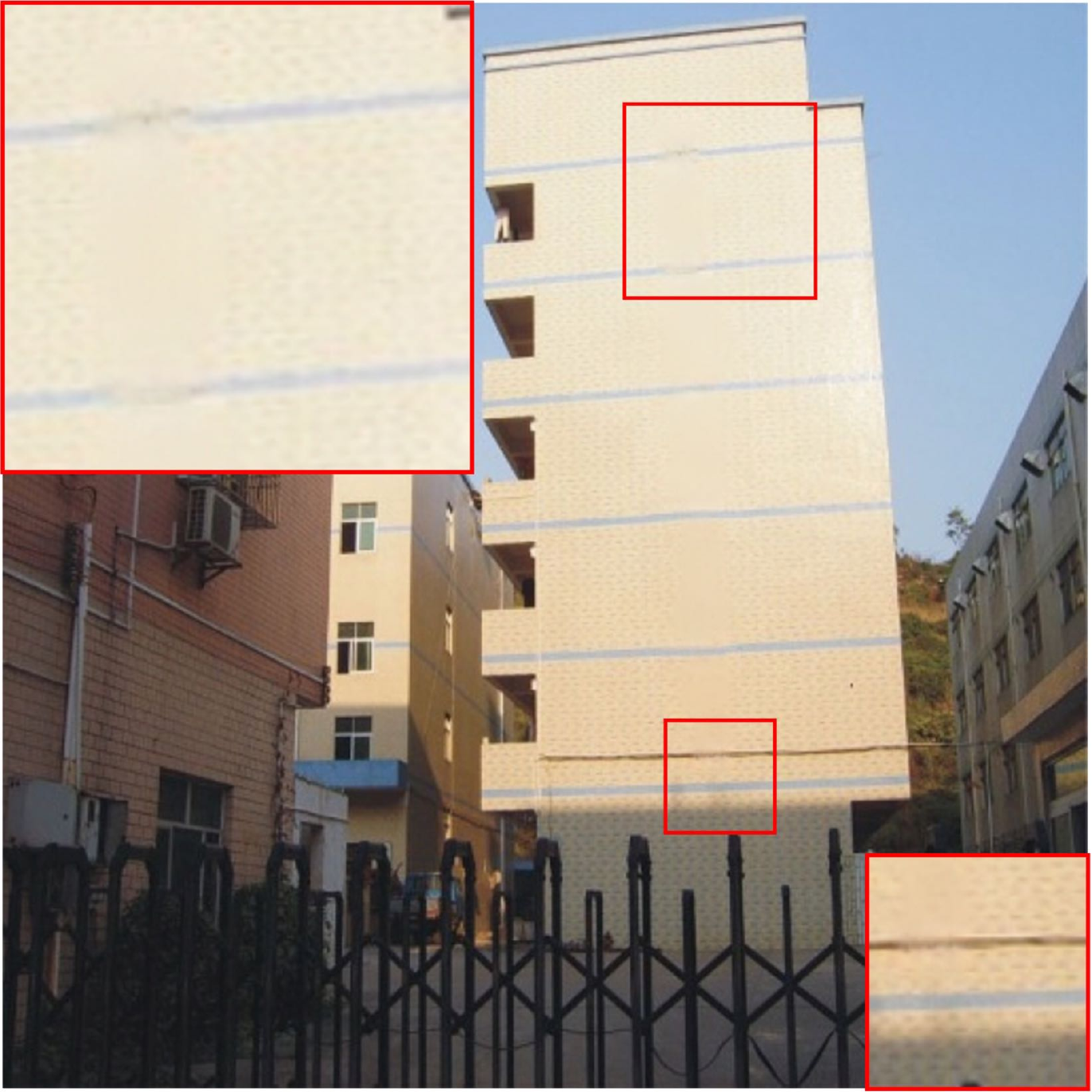}\\
			\includegraphics[width=1.6cm,height=1.8cm]{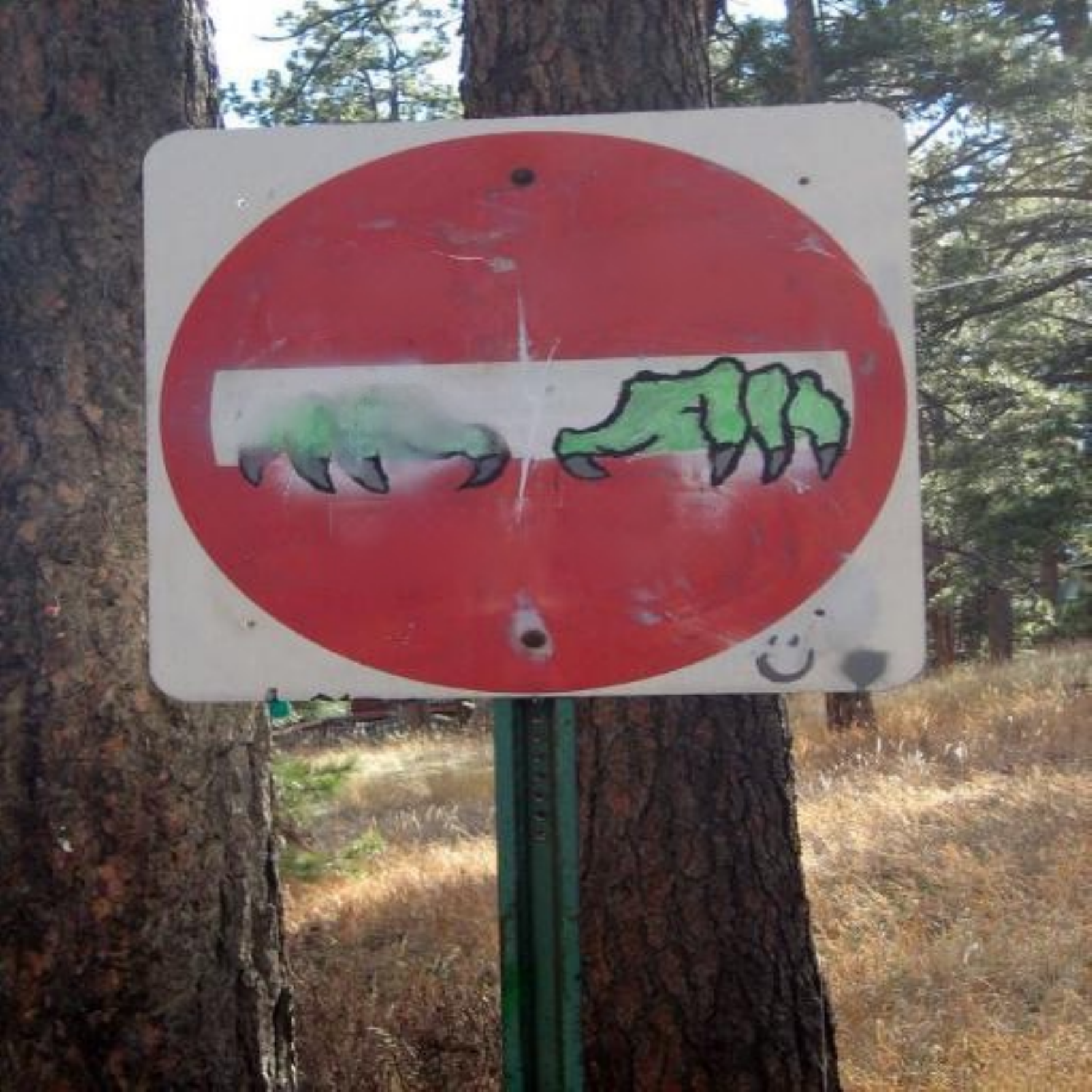}\\
		\end{minipage}%
	}
	\subfigure[]{
		\begin{minipage}[t]{0.121\linewidth}
			\centering
			\includegraphics[width=1.6cm,height=1.8cm]{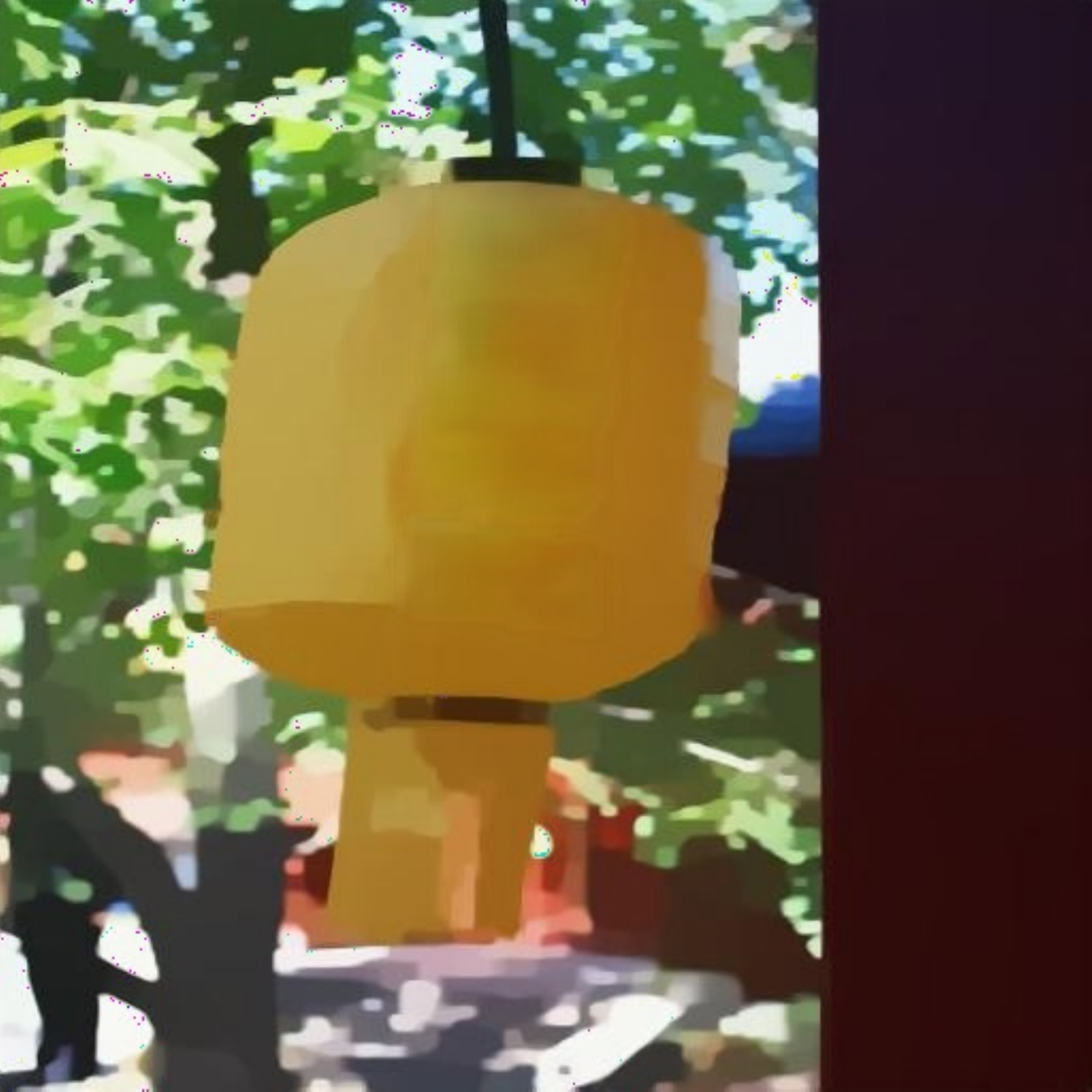}\\
			\includegraphics[width=1.6cm,height=1.8cm]{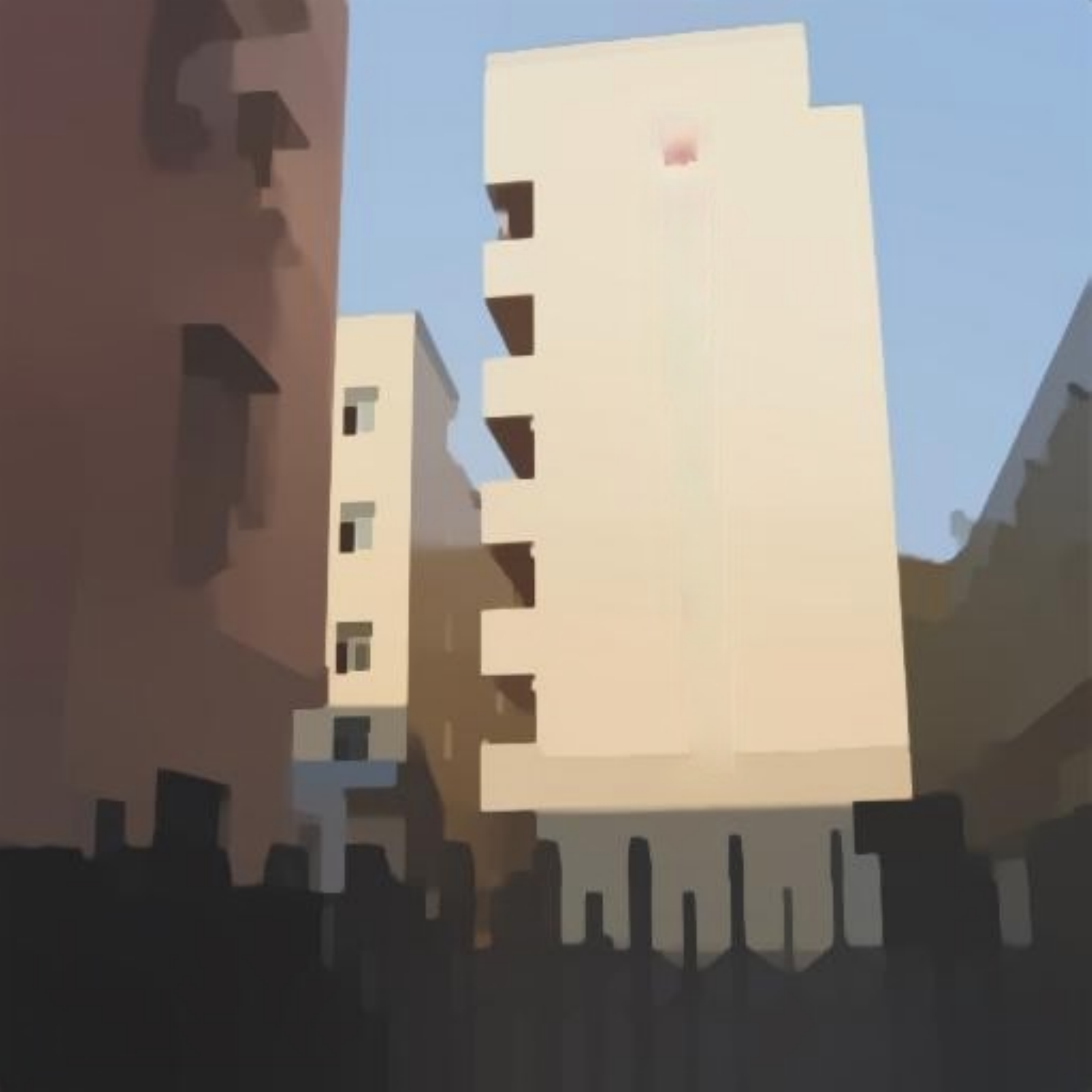}\\
			\includegraphics[width=1.6cm,height=1.8cm]{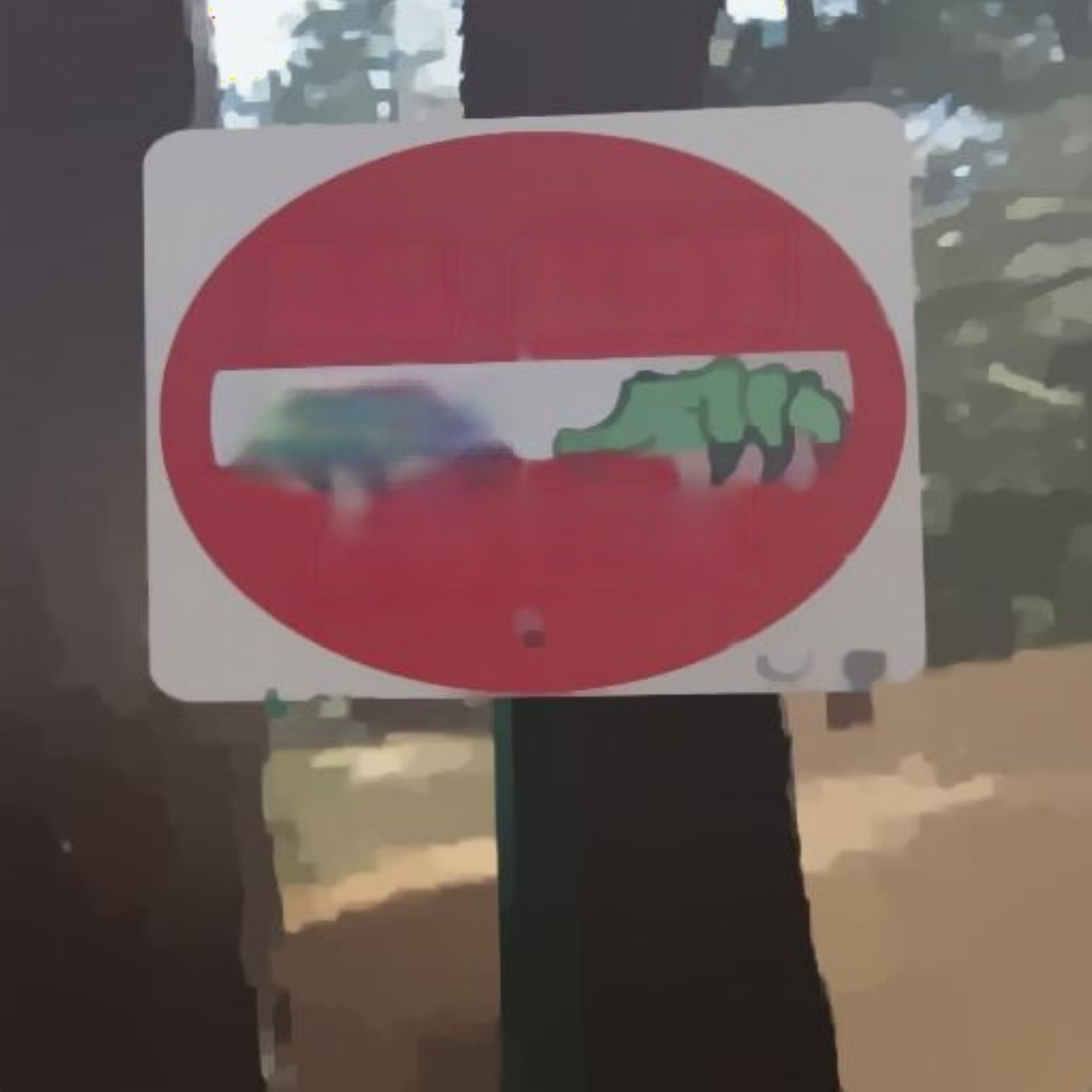}\\
		\end{minipage}%
	}
	\subfigure[]{
		\begin{minipage}[t]{0.12\linewidth}
			\centering
			\includegraphics[width=1.6cm,height=1.8cm]{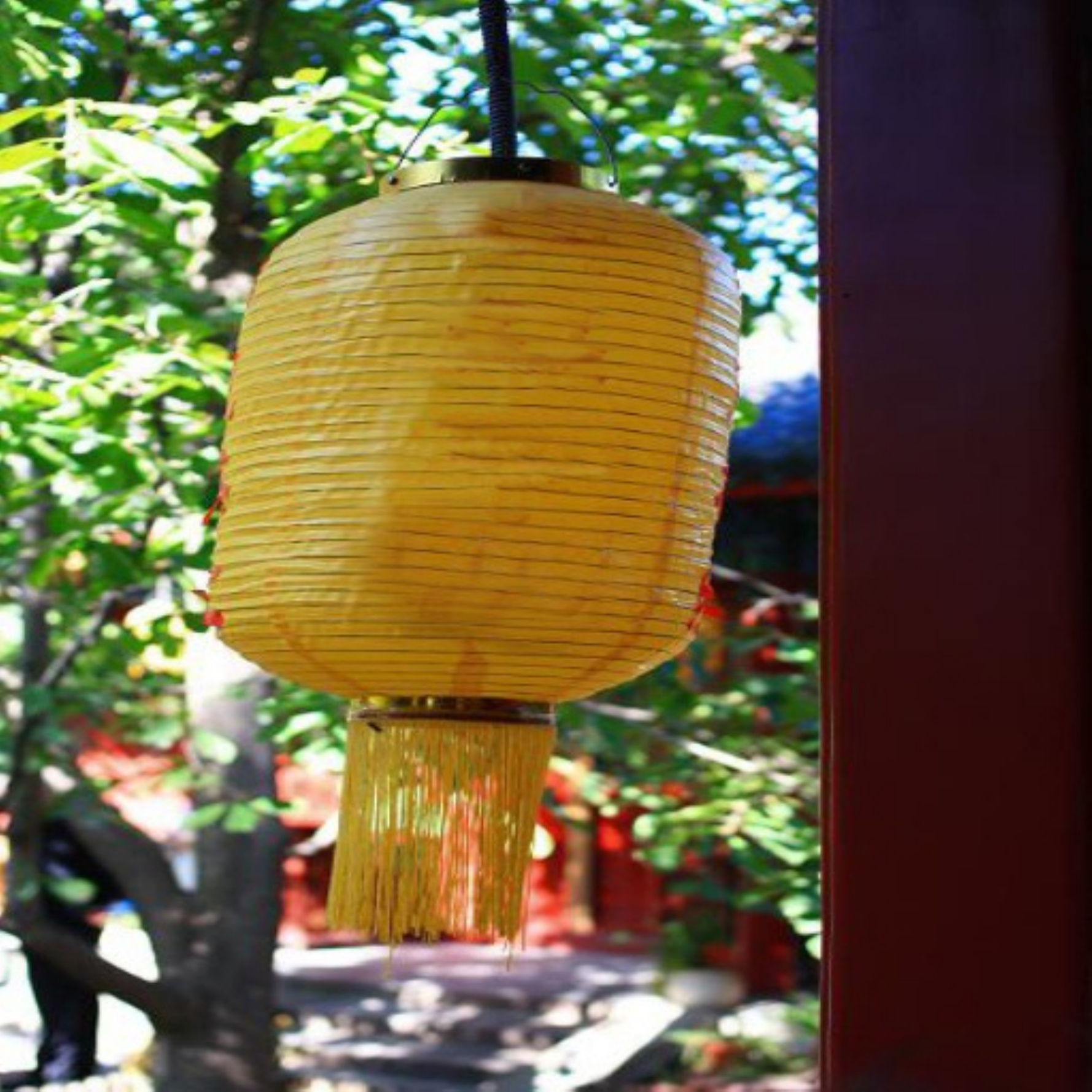}\\
			\includegraphics[width=1.6cm,height=1.8cm]{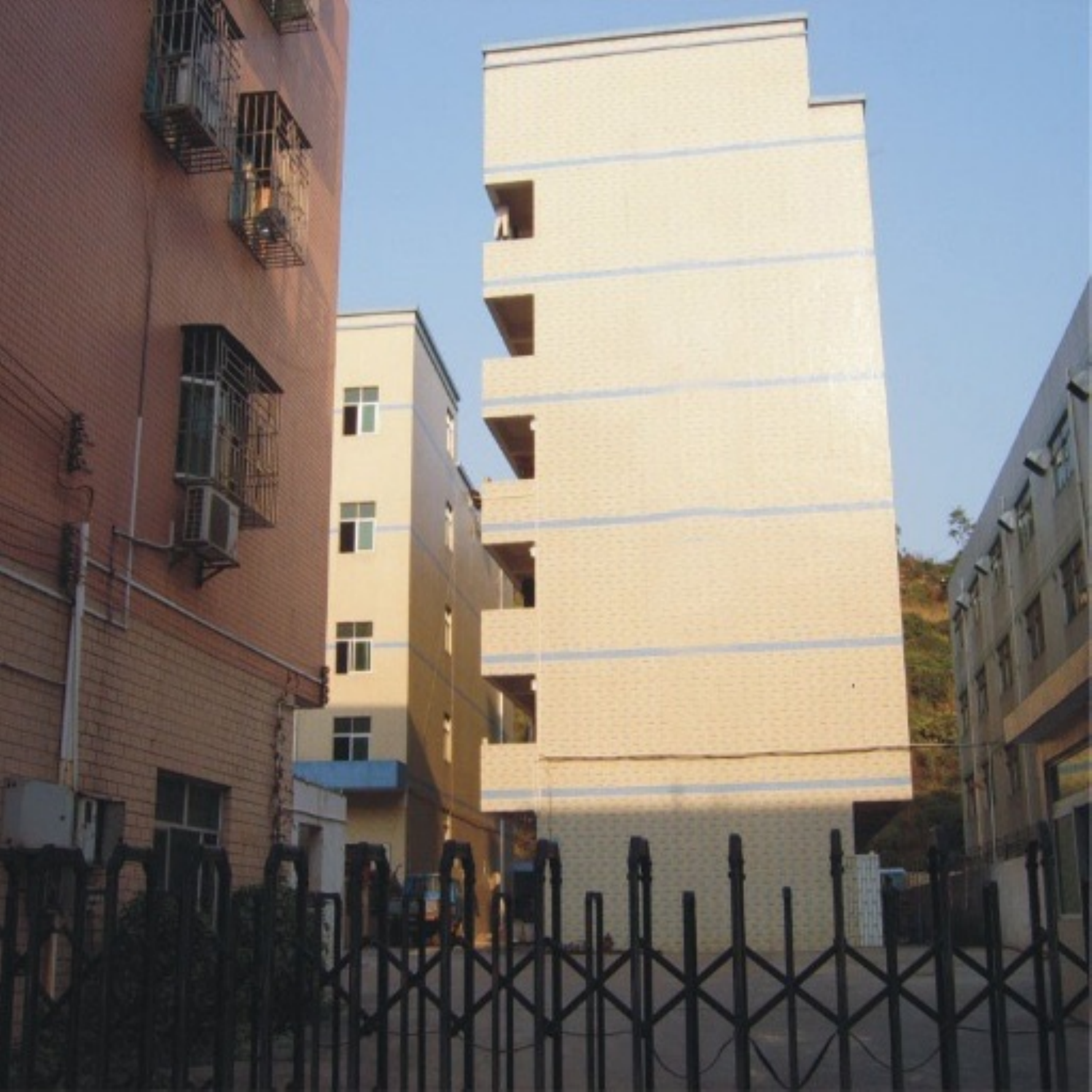}\\
			\includegraphics[width=1.6cm,height=1.8cm]{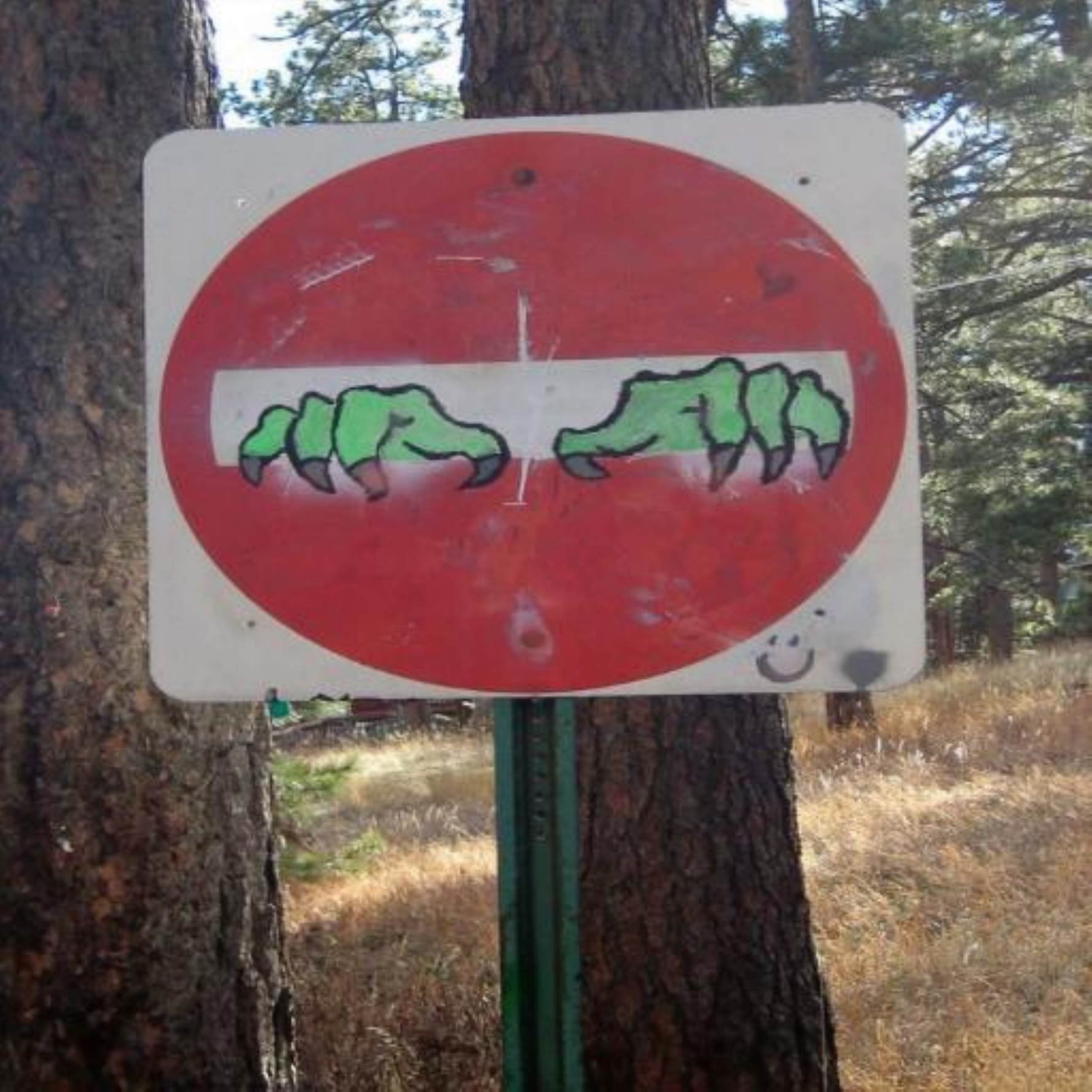}\\
		\end{minipage}%
	}
	\centering
	\caption{Qualitative results for ablation studies on HCG, LGCM, LCG. (a) The input images;  (b) Baseline results; (c) Baseline + HCG; (d) Baseline + HCG + LGCM; (e) Baseline + HCG + LGCM + LCG; (f) The structure output from LCG for the model (e);(g) The ground-truth.
		Zoom in for best view.} \label{fig:scut-enstext}
\end{figure*}

\subsection{Ablation Study}
In this section, we conduct experiments on SCUT-EnsText to verify the contributions of different components in CTRNet. 
Our baseline model is implemented by a Pix2pix-based model, which takes both the images and the corresponding masks as input.
As the text perception head is frozen when training the other components, the detected text regions remain the same in each experiment during inference; thus, we only employ Image-Eval to evaluate the performance.
Apart from the direct output $I_{out}$, we also paste the erased text regions back to the input images based on the detected results to obtain $I_{com}$. The quantitative results for both outputs are presented in Table \ref{table:aba}. The qualitative results are displayed in Fig. \ref{fig:scut-enstext}. 
Besides, we evaluate each loss item and their corresponding hyper-parameters, and the results are presented in our supplement materials.

\noindent \textbf{HCG:} HCG block aims to learn high-level discriminative context feature representation, which can effectively guide the process of feature modeling and decoding. As shown in Table \ref{table:aba}, 
the incorporation of HCG into the modeling and decoding phase with ResSPADE blocks yields significant improvement on all metrics,
with the increases of 0.51, 1.17, 0.02, 3.35 for $I_{out}$ and 1.67, 1.57, 0.01, 2.73 for $I_{com}$ in PSNR, MSSIM, MSE, and FID, respectively. The qualitative results shown in Fig. \ref{fig:scut-enstext} also illustrate the effect of this component. 
Comparing with the results of the baseline model in Fig. \ref{fig:scut-enstext} (b),  the results in Fig. \ref{fig:scut-enstext} (c) indicate that the HCG block can help generate a more plausible background and release more artifacts in the output.

\noindent \textbf{LGCM:} As shown in Table \ref{table:aba},  the incorporation of our LGCM significantly facilitates performance improvement of 2.20, 0.74, 0.02, 3.04 for $I_{out}$ in PSNR, MSSIM, MSE, and FID, respectively, while 0.42, 0.11, 0.01, 1.51 for $I_{com}$. Such a remarkable promotion benefits a lot from both the local and global modeling for the features and the learned context prior, which can capture not only the long-range dependency among pixels around the feature maps but their relationship at a fixed window as well. Therefore, LGCM enables our CTRNet to take advantage of both local and global information. The qualitative results are presented in Fig. \ref{fig:scut-enstext} (d). 
In comparison, the outputs of our model without LGCM exhibit some obvious defects on the text regions (the up/bottom row of Fig. \ref{fig:scut-enstext} (c)), while those with LGCM are more favorable, though there still exist mistaken erasure (e.g. the bottom of Figure \ref{fig:scut-enstext} (d), but the restored background is smoother than (c)).
Besides, with the long-rang dependency, the text can be removed more thoroughly with incorrect detection results, which is presented in the middle row of Fig. \ref{fig:scut-enstext} (c) and (d). Furthermore, we discuss the number of LGCM blocks in the supplementary materials.

\begin{figure}[t]
	\subfigbottomskip=2pt
	\subfigcapskip=2pt
	\setlength{\belowcaptionskip}{-0.1cm} 
	\setlength{\abovecaptionskip}{-0.0cm}
	\centering
	\subfigure[Input]{
		\begin{minipage}[t]{0.18\linewidth}
			\centering
			\includegraphics[width=2.3cm,height=1.7cm]{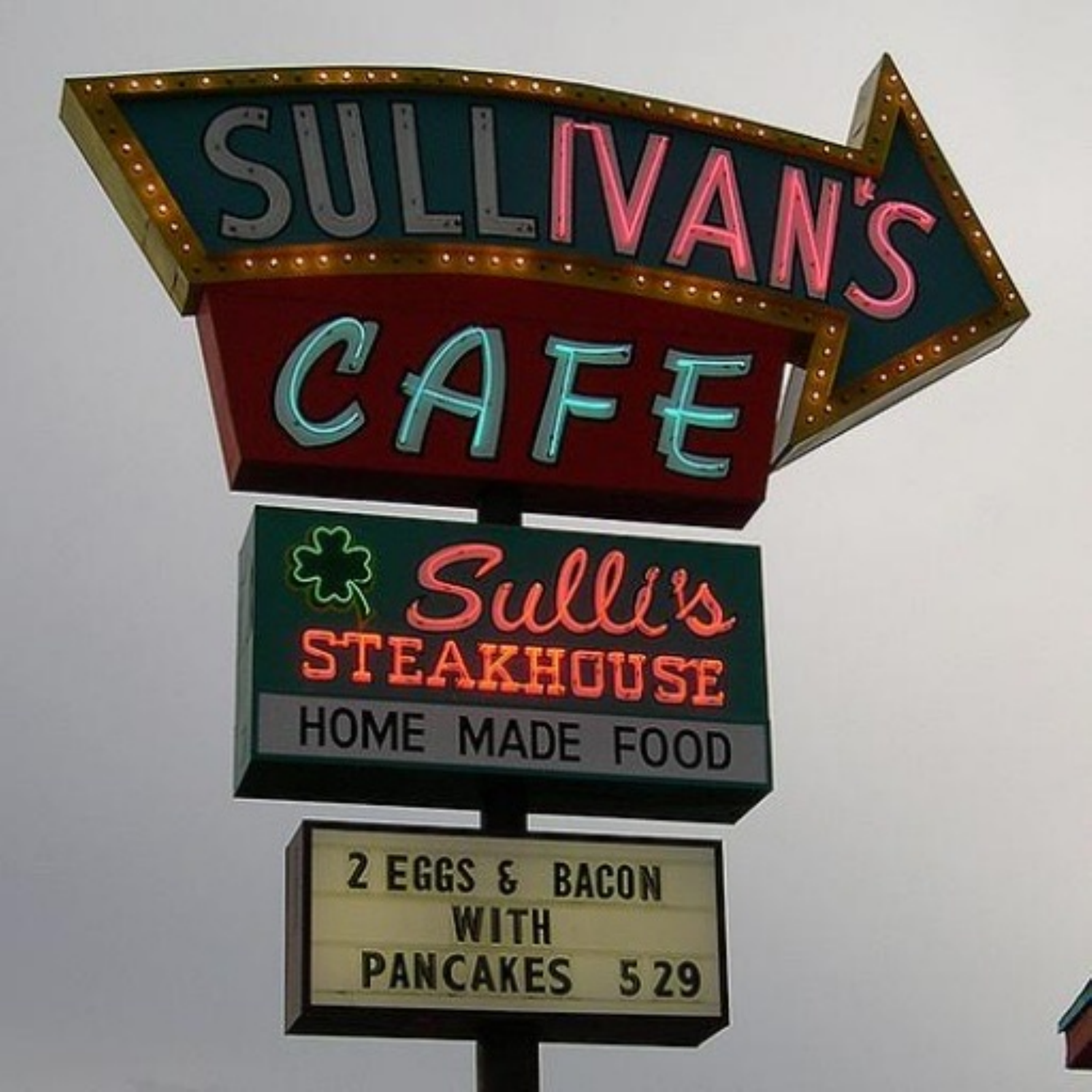}\\
			\includegraphics[width=2.3cm,height=1.7cm]{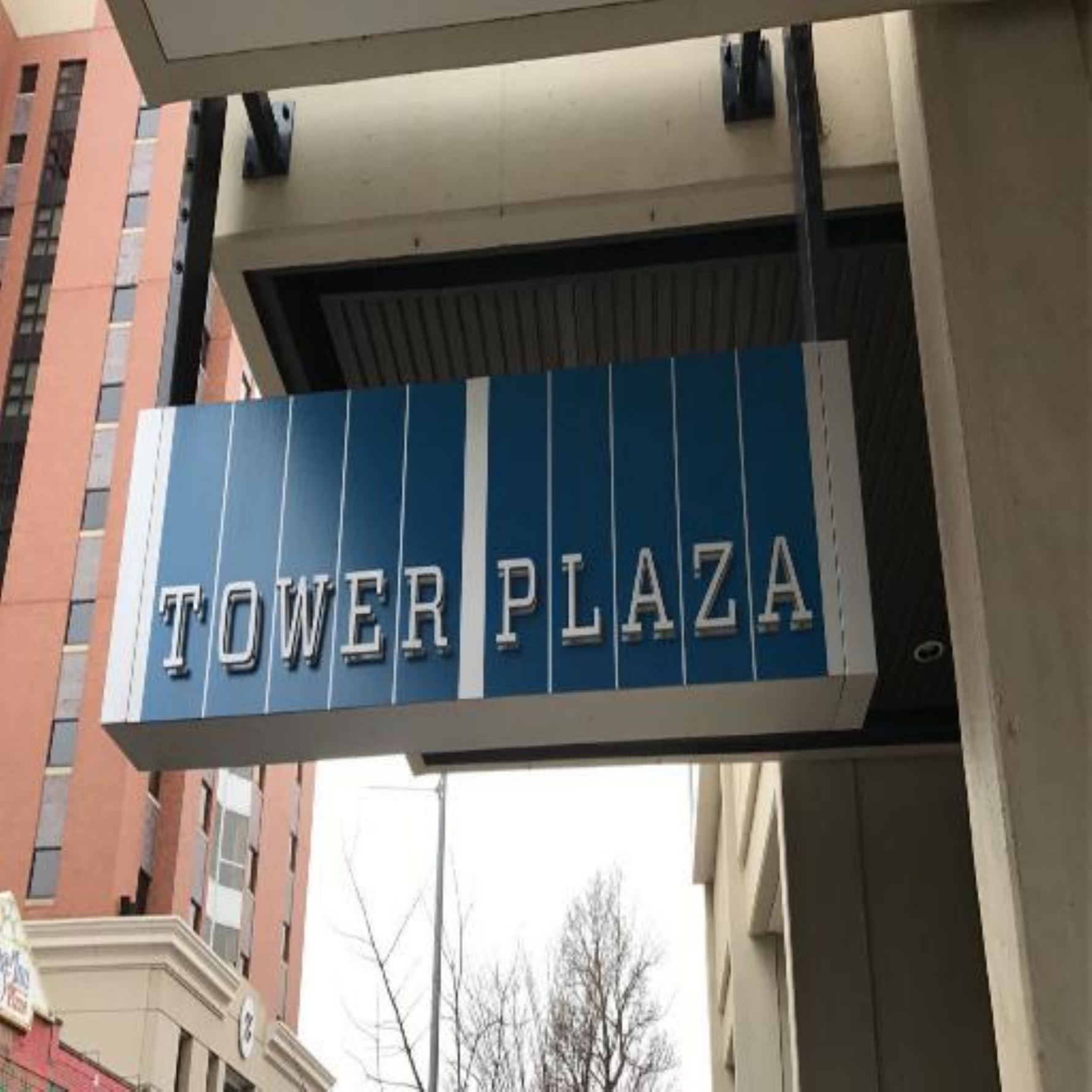}\\
		\end{minipage}%
	}
	\subfigure[GT]{
		\begin{minipage}[t]{0.18\linewidth}
			\centering
			\includegraphics[width=2.3cm,height=1.7cm]{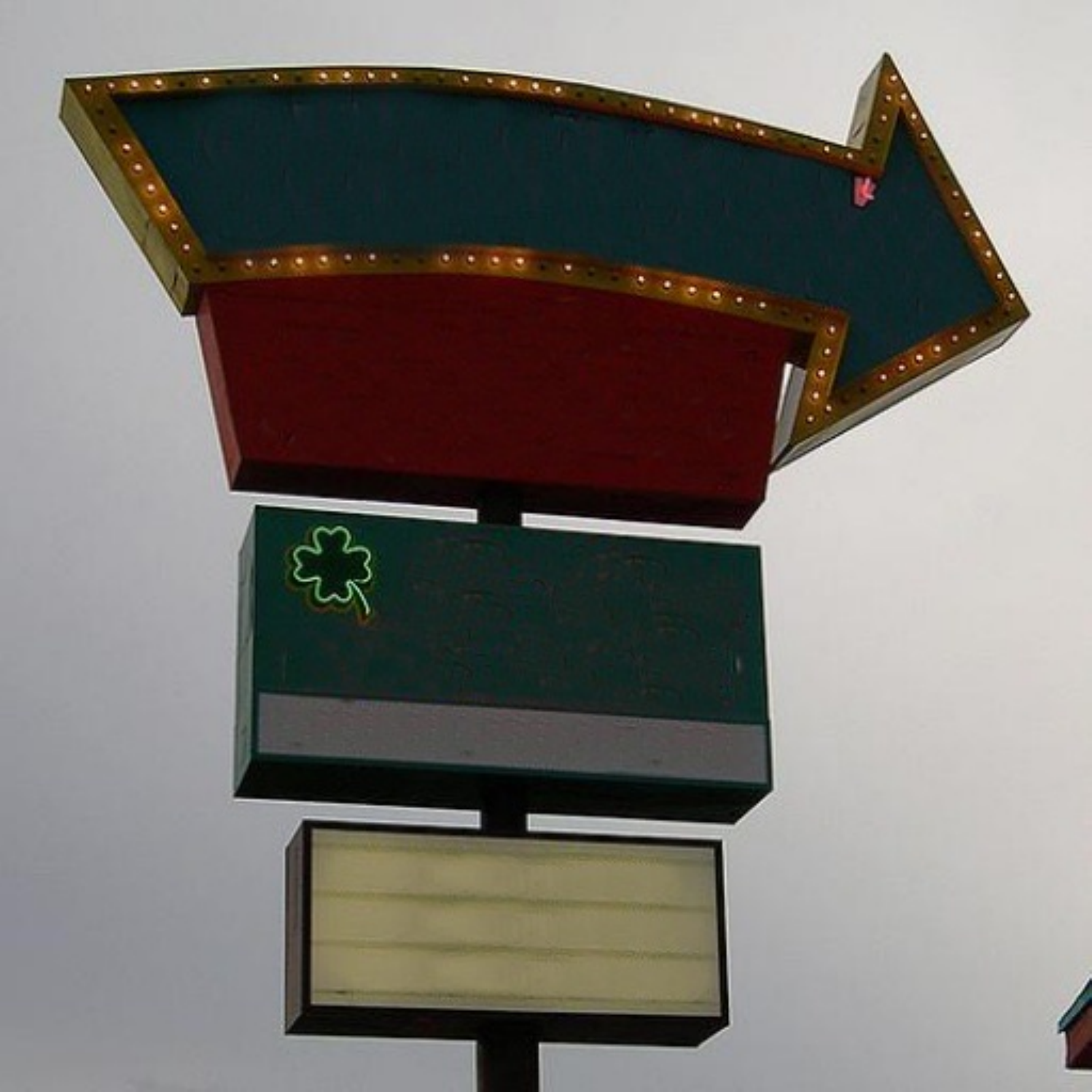}\\
			\includegraphics[width=2.3cm,height=1.7cm]{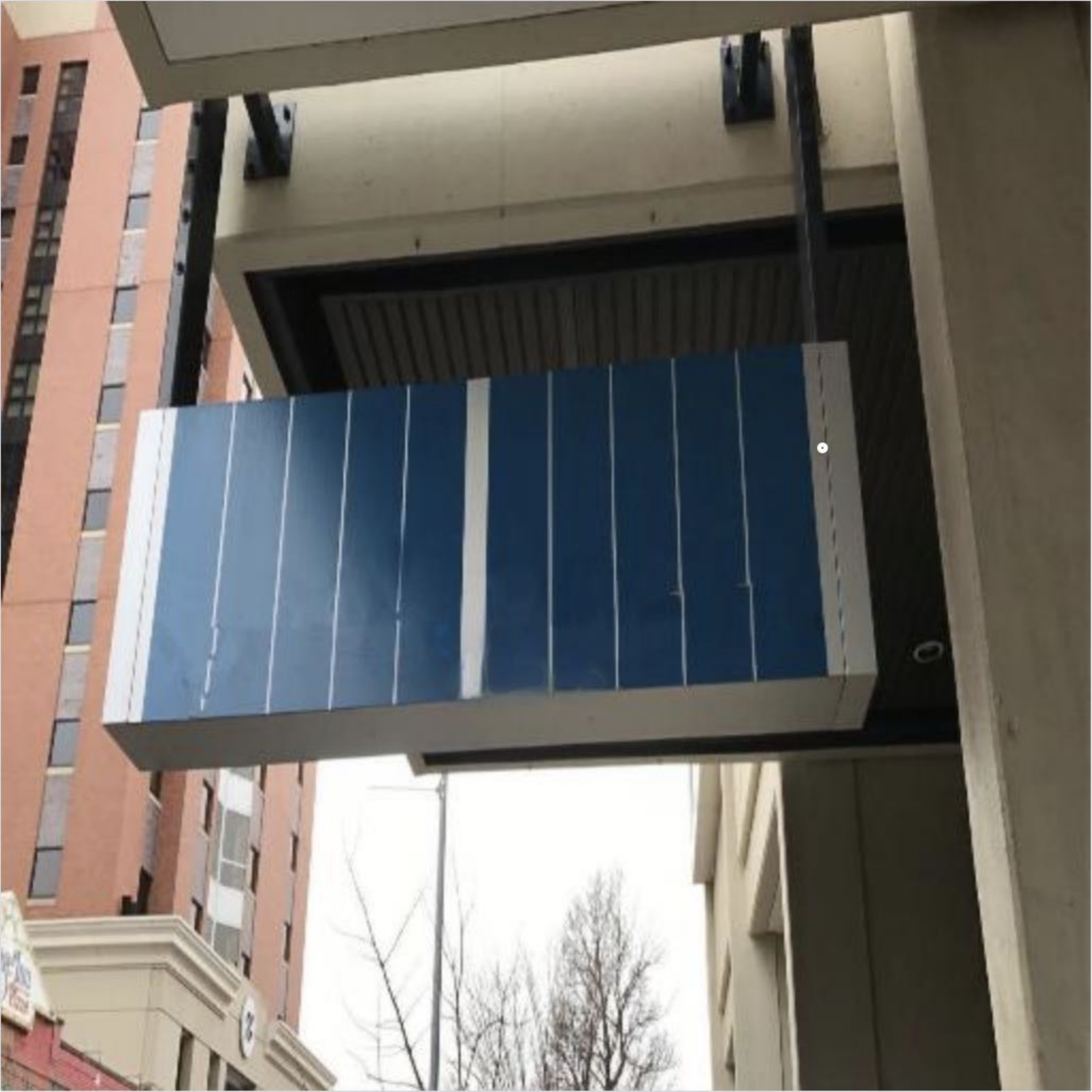}\\
		\end{minipage}%
	}
	\subfigure[EN\cite{liu2020erasenet}]{
		\begin{minipage}[t]{0.188\linewidth}
			\centering
			\includegraphics[width=2.4cm,height=1.7cm]{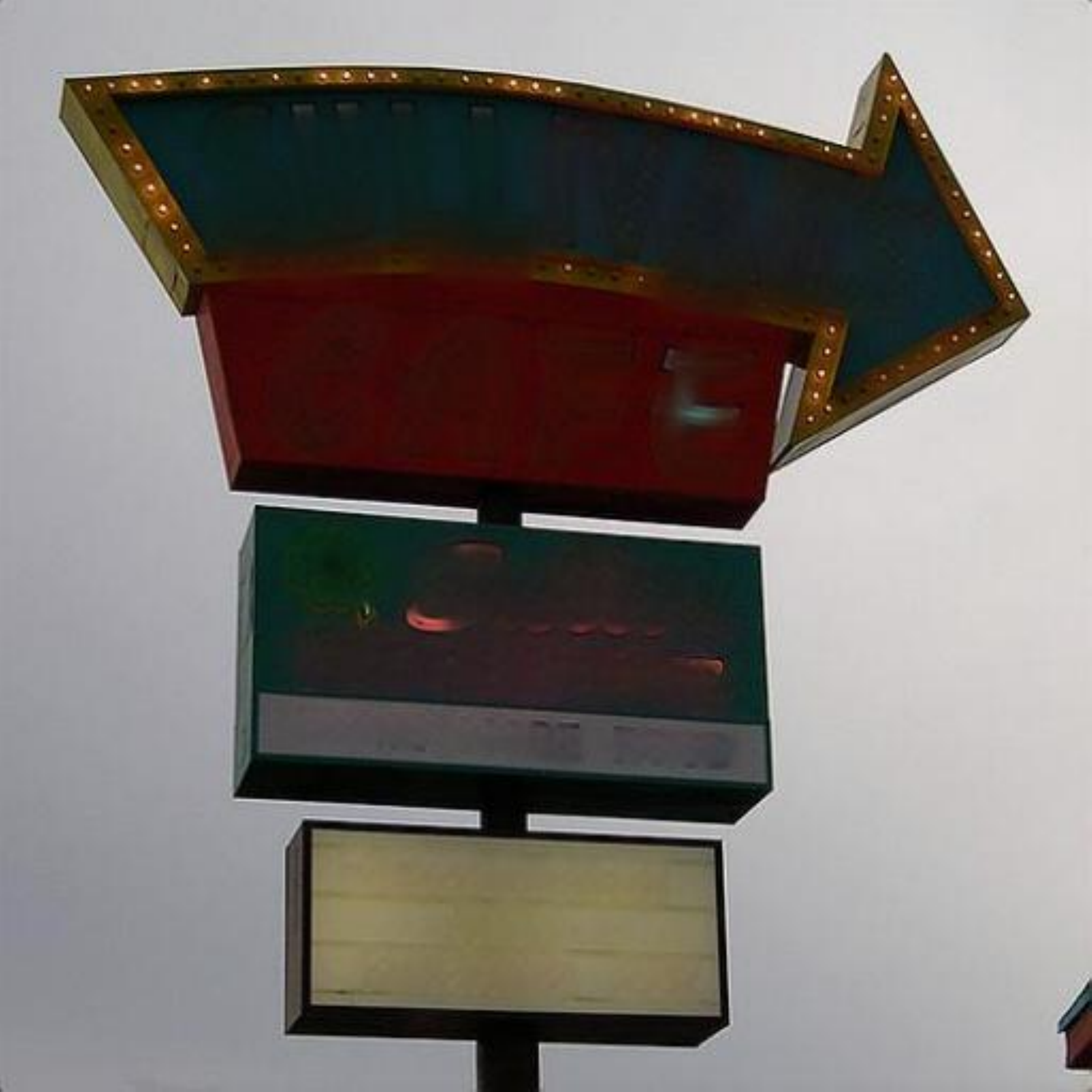}\\
			\includegraphics[width=2.4cm,height=1.7cm]{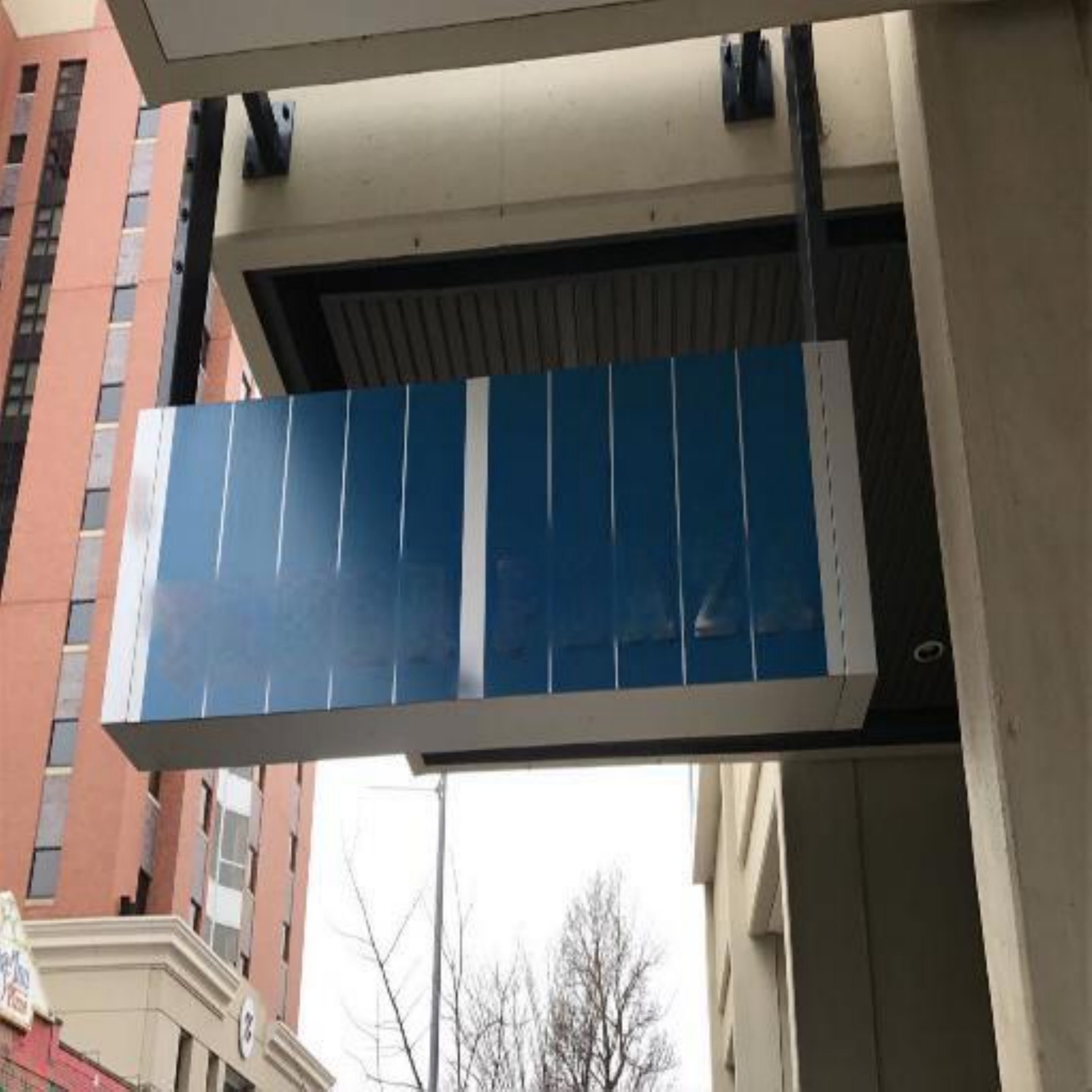}\\
		\end{minipage}%
	}
	\subfigure[Stroke\cite{tang2021stroke}]{
		\begin{minipage}[t]{0.188\linewidth}
			\centering
			\includegraphics[width=2.4cm,height=1.7cm]{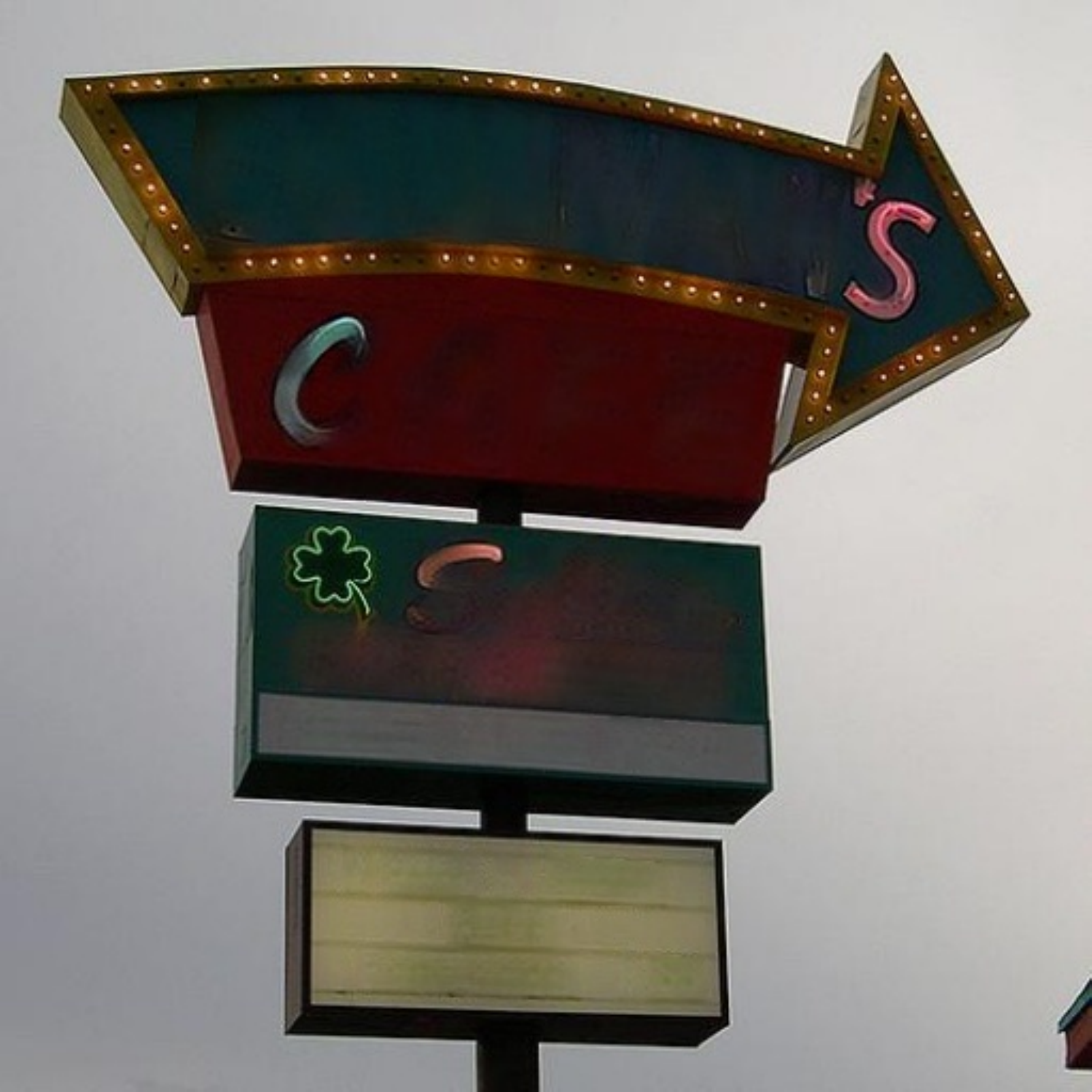}\\
			\includegraphics[width=2.4cm,height=1.7cm]{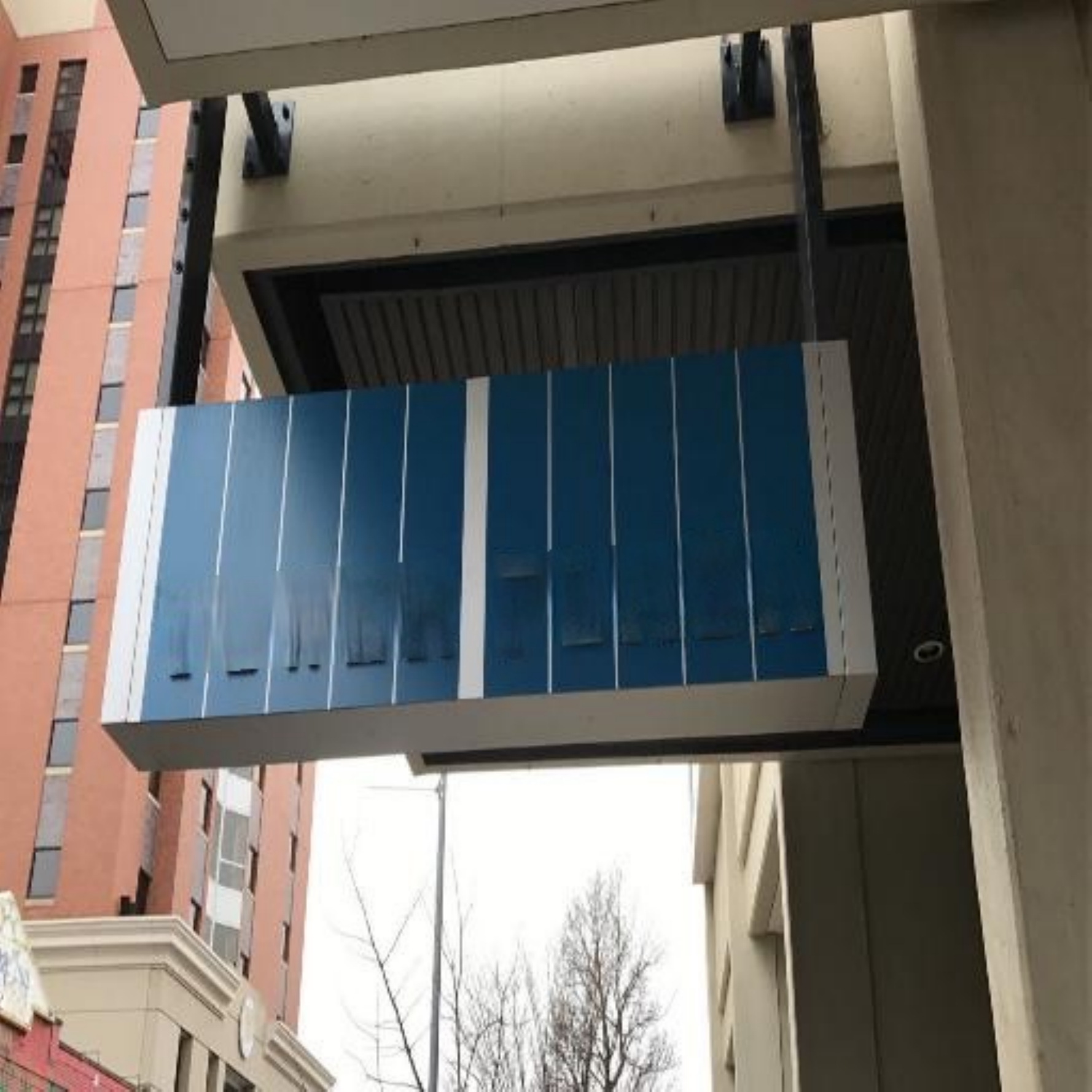}\\
		\end{minipage}%
	}
	\subfigure[Ours]{
		\begin{minipage}[t]{0.18\linewidth}
			\centering
			\includegraphics[width=2.3cm,height=1.7cm]{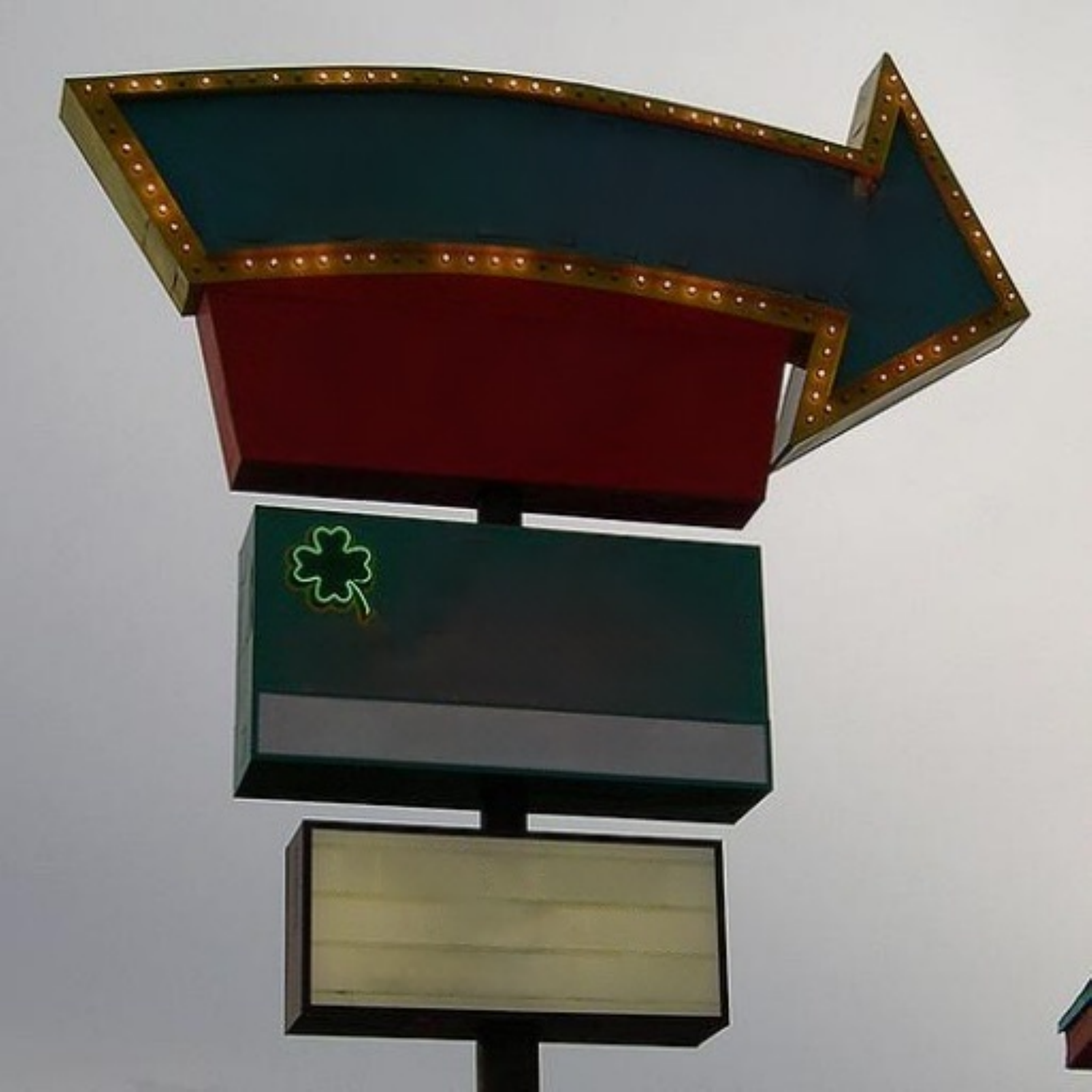}\\
			\includegraphics[width=2.3cm,height=1.7cm]{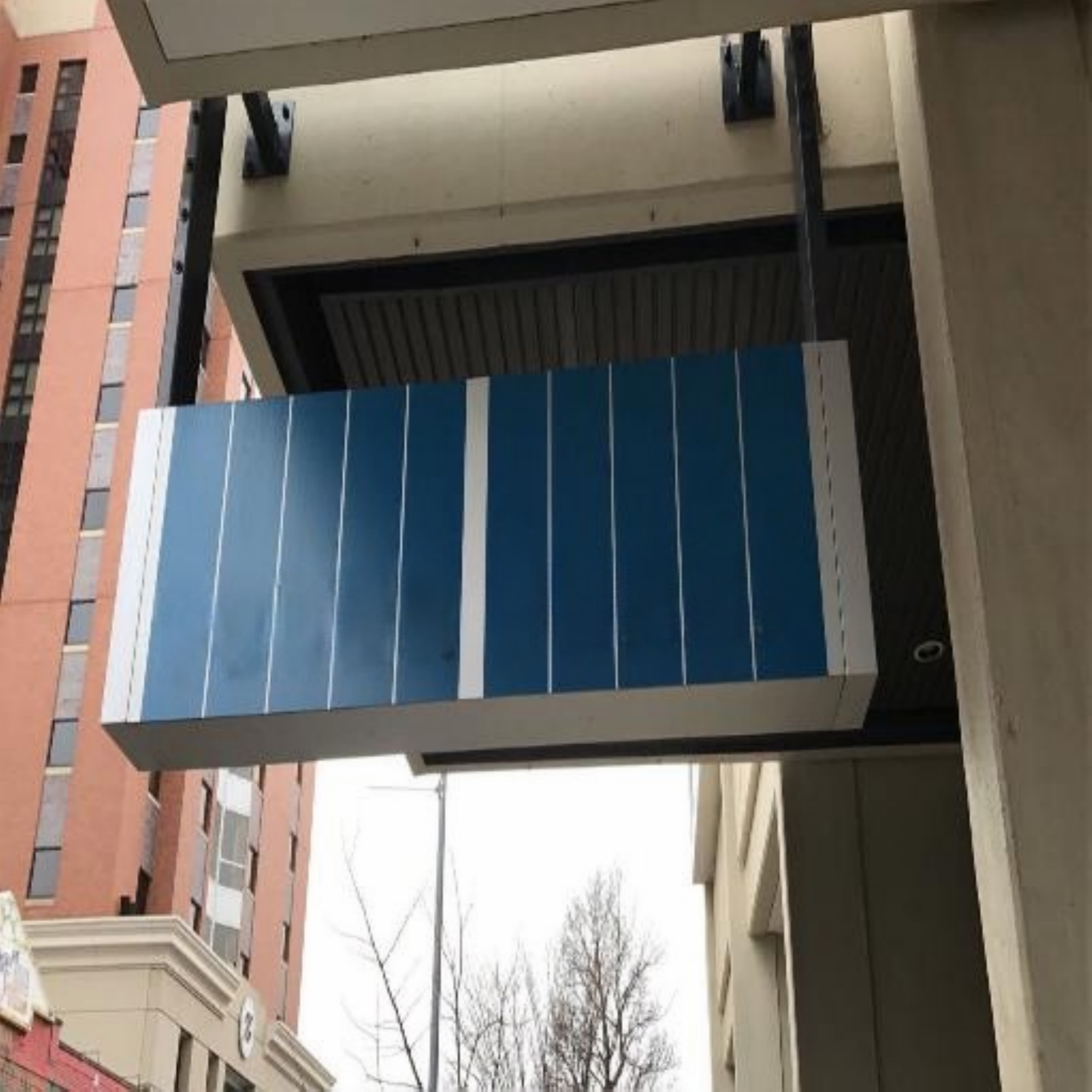}\\
		\end{minipage}%
	}
	\centering
	\caption{Qualitative results on SCUT-EnsText for comparing our model with previous scene text removal methods. EN denotes EraseNet~\cite{liu2020erasenet} and Stroke denotes the method proposed by Tang et al.~\cite{tang2021stroke}. Zoom in for best view.} \label{fig:sota_ens}
\end{figure}

\begin{table*}[t]
	\centering
	\setlength{\belowcaptionskip}{0.1cm} 
	\caption{Comparison with state-of-the-art methods on SCUT-EnsText. The methods with ``*'' denote that the text mask are generated with the GT instead of the detected results. MSSIM and MSE are represented by $\%$ in the table. Bold indicates SOTA, while Underline indicates second best. }
	\label{table:sota_enstext}
	\setlength{\tabcolsep}{1mm}{
		\begin{tabular}{ c | c c c c | c c c | c c c }
			\hline \Xhline{0.3pt}
			\multirow{2}{*}{Methods} & \multicolumn{4}{c|}{Image-Eval} & \multicolumn{6}{c}{Detection-Eval}  \\
			\cline{2-11} 
			& PSNR & MSSIM & MSE & FID & R & P & H & T-R & T-P & T-H  \\
			\hline \Xhline{0.3pt}
			Original & - & - & - &  - & 69.5 & 79.4 & 74.1 & 50.9 & 61.4 & 55.7   \\
			\hline
			Pix2pix\cite{isola2017image} & 26.70 & 88.56 & 0.37 &  46.88 &35.4 & 69.7 & 47.0 & 24.3 & 52.0 & 33.1    \\
			\hline
			STE\cite{ste} & 25.46 & 90.14 & 0.47 &  43.39 & 5.9 & \underline{40.9} & 10.2 & 3.6 & \underline{28.9} & 6.4    \\
			\hline
			EnsNet\cite{zhang2019ensnet} & 29.54 & 92.74 & 0.24 &  32.71 &32.8 & 68.7 & 44.4 & 50.7 & 22.1 & 30.8    \\
			\hline
			EraseNet\cite{liu2020erasenet} & 32.30 & 95.42 & 0.15 &  19.27 & 4.6 & 53.2 & 8.5 & 2.9 & 37.6 & 5.4    \\
			\hline
			PERT\cite{wang2021simple} & \underline{33.25} & \underline{96.95} & \underline{0.14} &  - & \underline{2.9}  & 52.7 & \underline{5.4} & \underline{1.8} & 38.7 & \underline{3.5}    \\ \hline
			Ours ($I_{out}$) & \textbf{35.20} & \textbf{97.36} & \textbf{0.09} & \textbf{13.99} & \textbf{1.4} & \textbf{38.4} & \textbf{2.7} & \textbf{0.9} & \textbf{28.3} & \textbf{1.7}   \\
			\hline \Xhline{0.3pt}
			Tang et al.\cite{tang2021stroke} & \underline{35.34} & 96.24 & 0.09 &  - & 3.6 & - & - & - & - & -    \\
			\hline 
			Ours ($I_{com}$) & \textbf{35.85} & \textbf{97.40} & \textbf{0.09} & \textbf{14.57} & \textbf{1.7} & \textbf{40.1} & \textbf{3.3} & \textbf{1.1} & \textbf{29.4} & \textbf{2.1}   \\
			\hline \Xhline{0.3pt}
			Tang et al.*\cite{tang2021stroke} & \underline{37.08} & 96.54 & \textbf{0.05} &  - & - & - & - & - & - & -    \\
			\hline 
			Ours* ($I_{com}$) & \textbf{37.20} & \textbf{97.66} & \underline{0.07} & \textbf{11.72} & - & - & - & - & - & -    \\
			\hline  \Xhline{0.3pt}
	\end{tabular}}
\end{table*}

\noindent \textbf{LCG:} According to the results shown in Table \ref{table:aba}, under the same experimental setting, CTRNet with LCG yields slightly improvement in PSNR and FID by approximately 0.03 and 0.40 on average, respectively for both $I_{out}$ and $I_{com}$, while the other metrics remain comparable. One of the basic challenges of scene text removal is the background restoration, the generated structure can indicate the region clues of text-erased so that provide the guidance for texture synthesis of the background. Fig. \ref{fig:scut-enstext} (e) and (f) show the outputs of CTRNet with LCG and the generated background structure. With the background structure, our text removal network can restore a more natural background texture, as indicated by the red boxes in the figures. Besides, as shown in the bottom row of Fig. \ref{fig:scut-enstext} (e), there exist wrong detected results, but LCG can still help retain the corresponding region and predict more reasonable contents than others.

\noindent \textbf{Soft Mask:} 
The application of soft-mask mainly aims to eliminate the discontinuity and inconsistency at the junction of text/non-text regions on the output. Soft mask only achieves only a slight improvement on $I_{out}$, but a significant promotion on  $I_{com}$ by 0.53 in PSNR and 0.22 in MSSIM, respectively. Qualitative results of $I_{com}$  for the comparison on hard-mask (0-1) and soft-mask are shown in the supplement file. With soft-mask, the output can preserve smoother edges between text and non-text regions. Besides, as the soft-mask is expanded, the texts are removed more completely and thoroughly.

\begin{figure}[t]
	\subfigbottomskip=2pt
	\subfigcapskip=2pt
	\setlength{\belowcaptionskip}{-0.1cm} 
	\setlength{\abovecaptionskip}{-0.0cm}
	\centering
	\subfigure[Input]{
		\begin{minipage}[t]{0.18\linewidth}
			\centering
			\includegraphics[width=2.3cm,height=1.8cm]{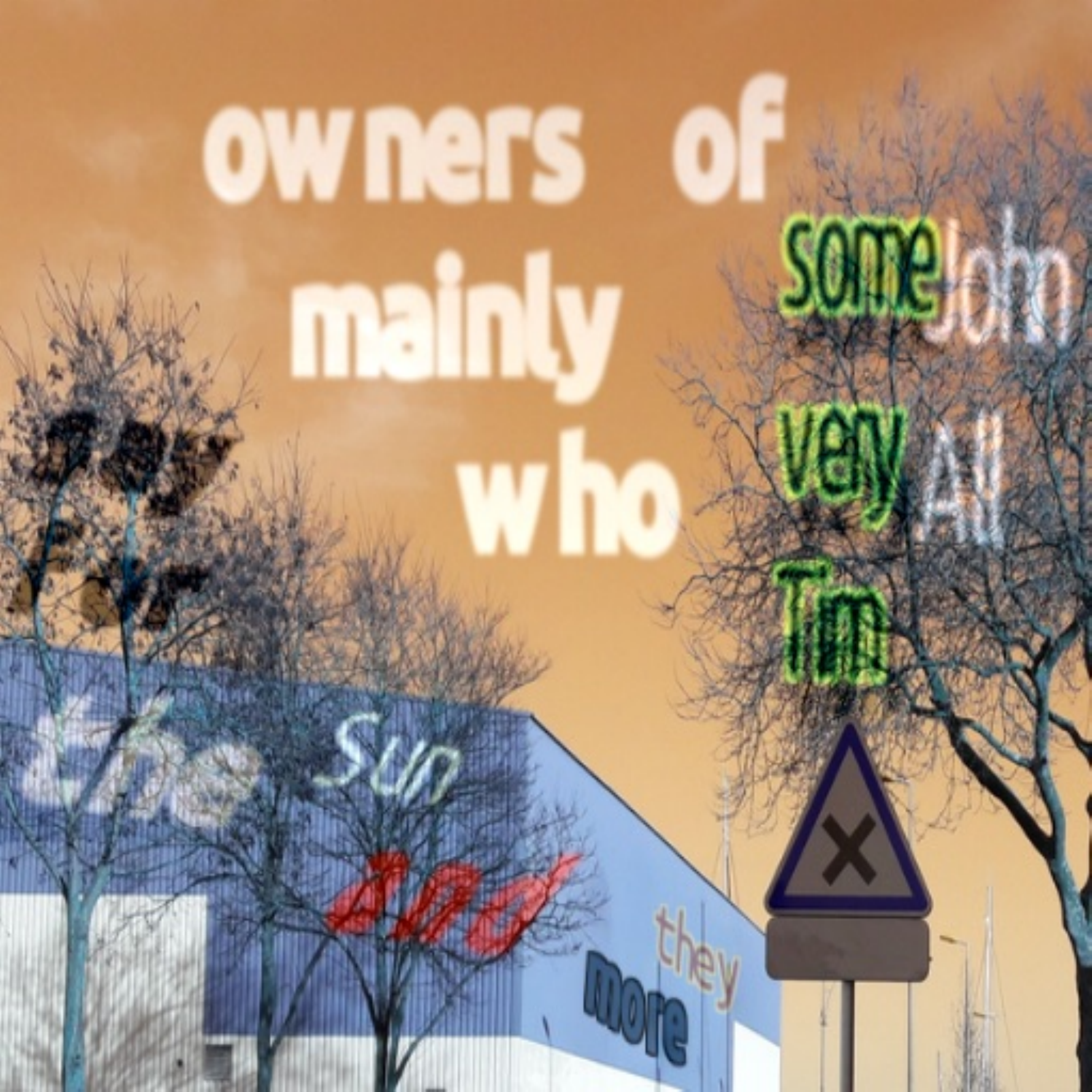}\\
			\includegraphics[width=2.3cm,height=1.8cm]{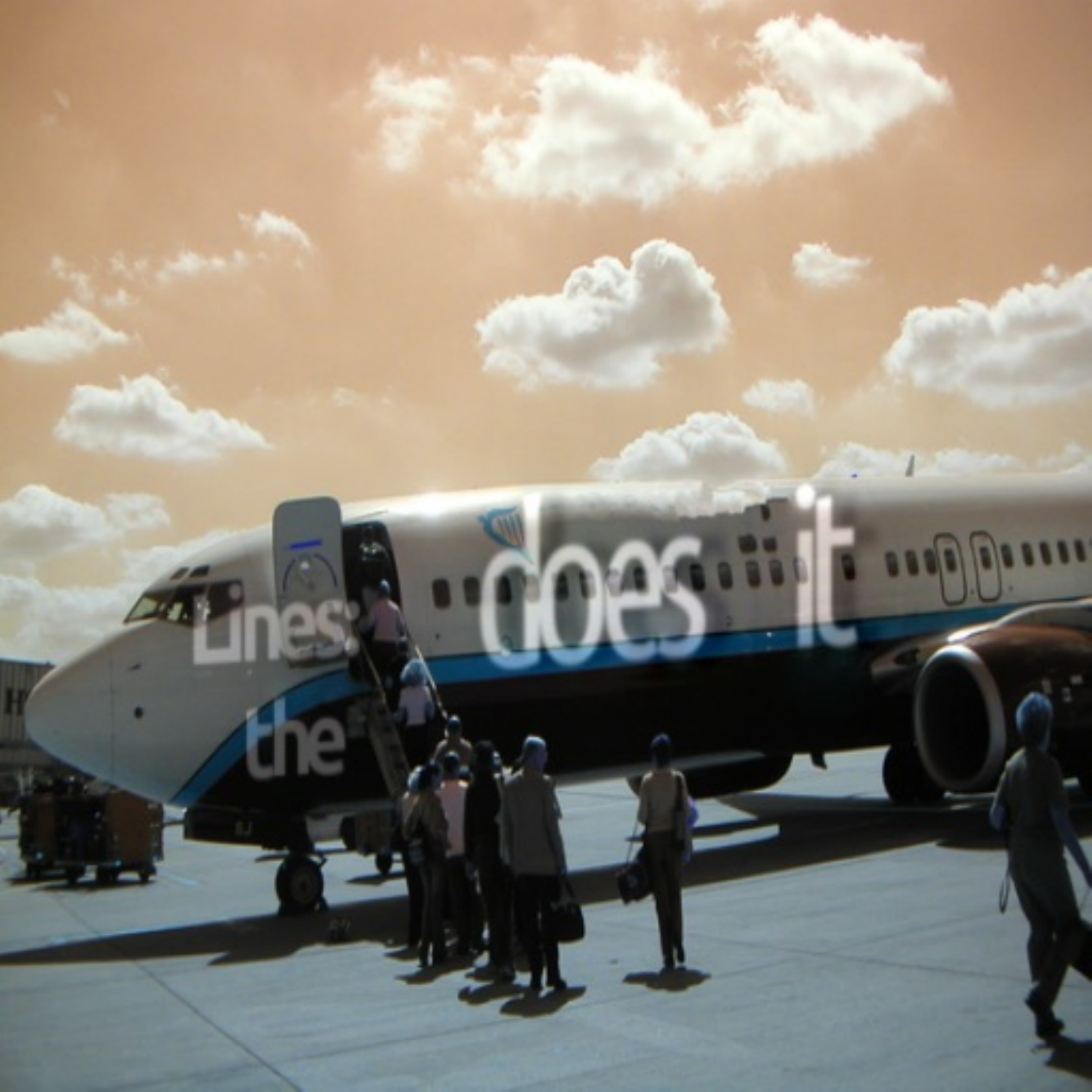}\\
		\end{minipage}%
	}
	\subfigure[GT]{
		\begin{minipage}[t]{0.18\linewidth}
			\centering
			\includegraphics[width=2.3cm,height=1.8cm]{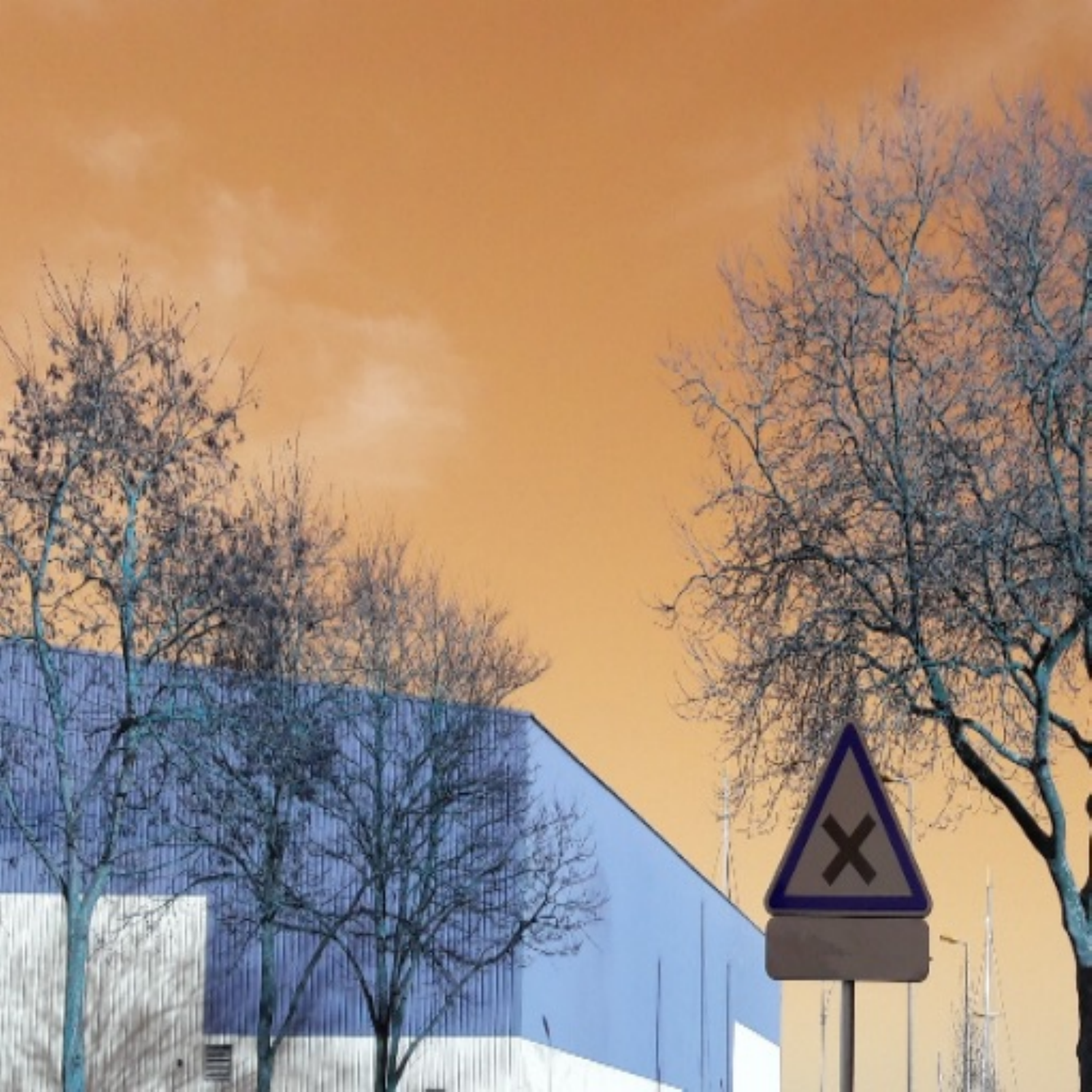}\\
			\includegraphics[width=2.3cm,height=1.8cm]{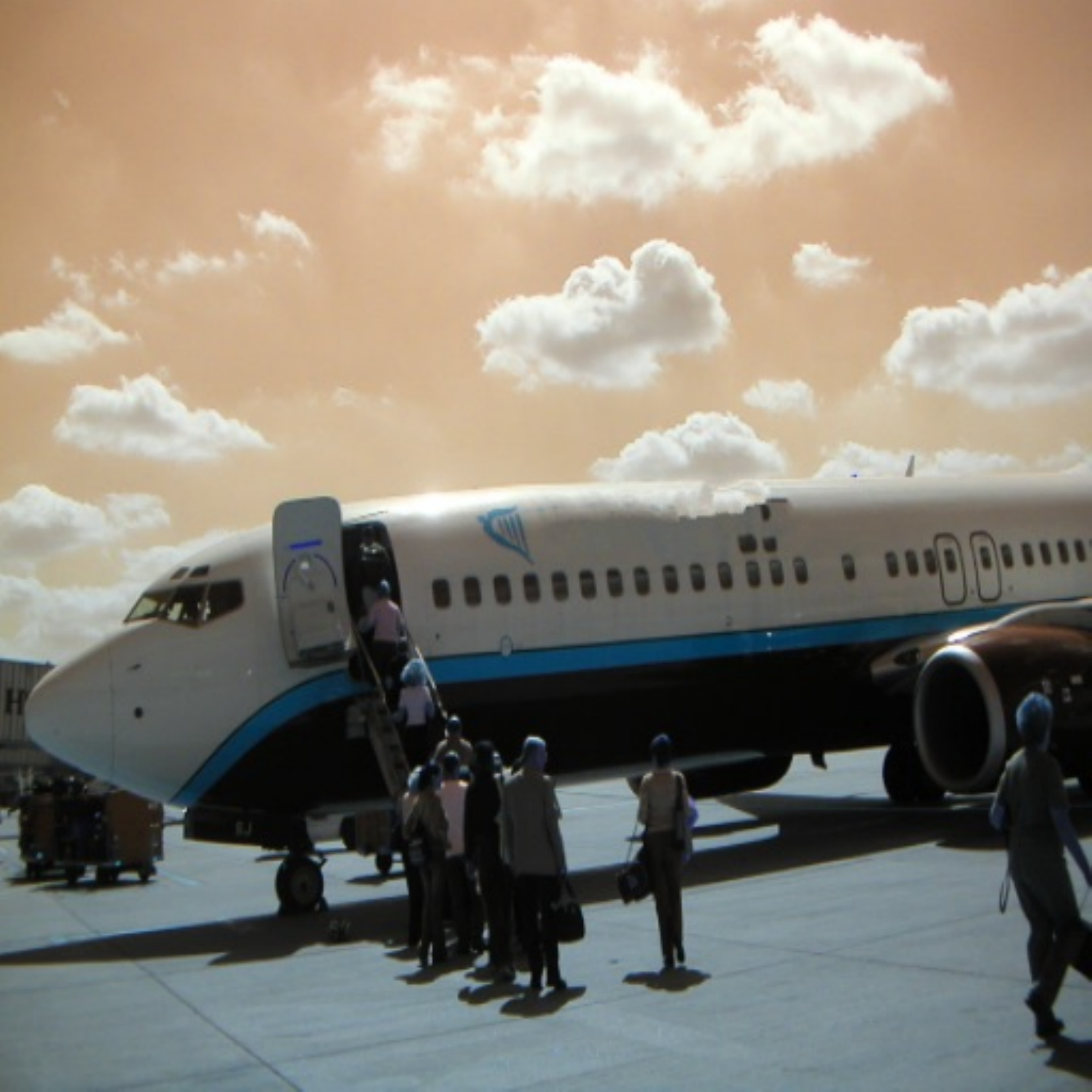}\\
		\end{minipage}%
	}
	\subfigure[EN\cite{liu2020erasenet}]{
		\begin{minipage}[t]{0.188\linewidth}
			\centering
			\includegraphics[width=2.4cm,height=1.8cm]{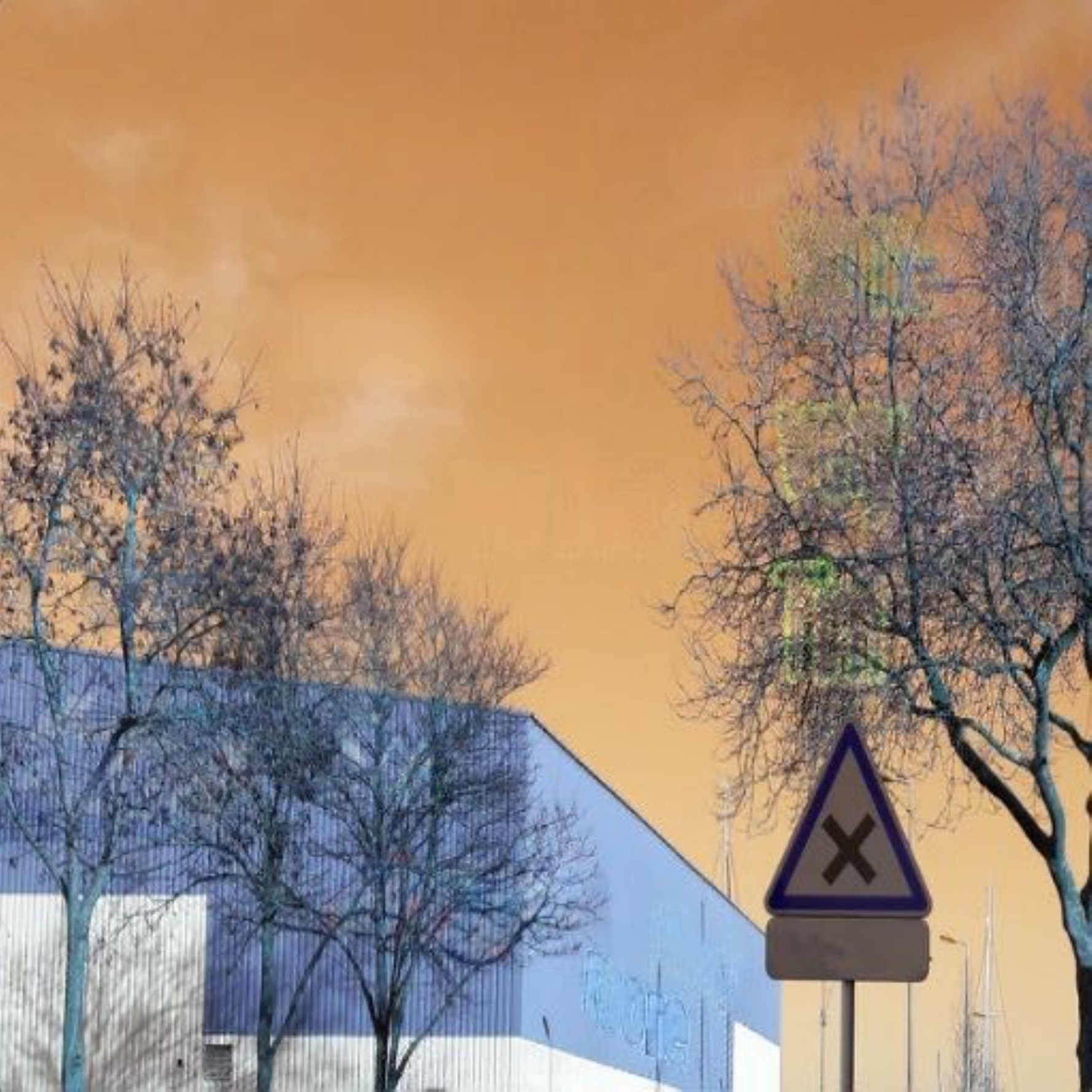}\\
			\includegraphics[width=2.4cm,height=1.8cm]{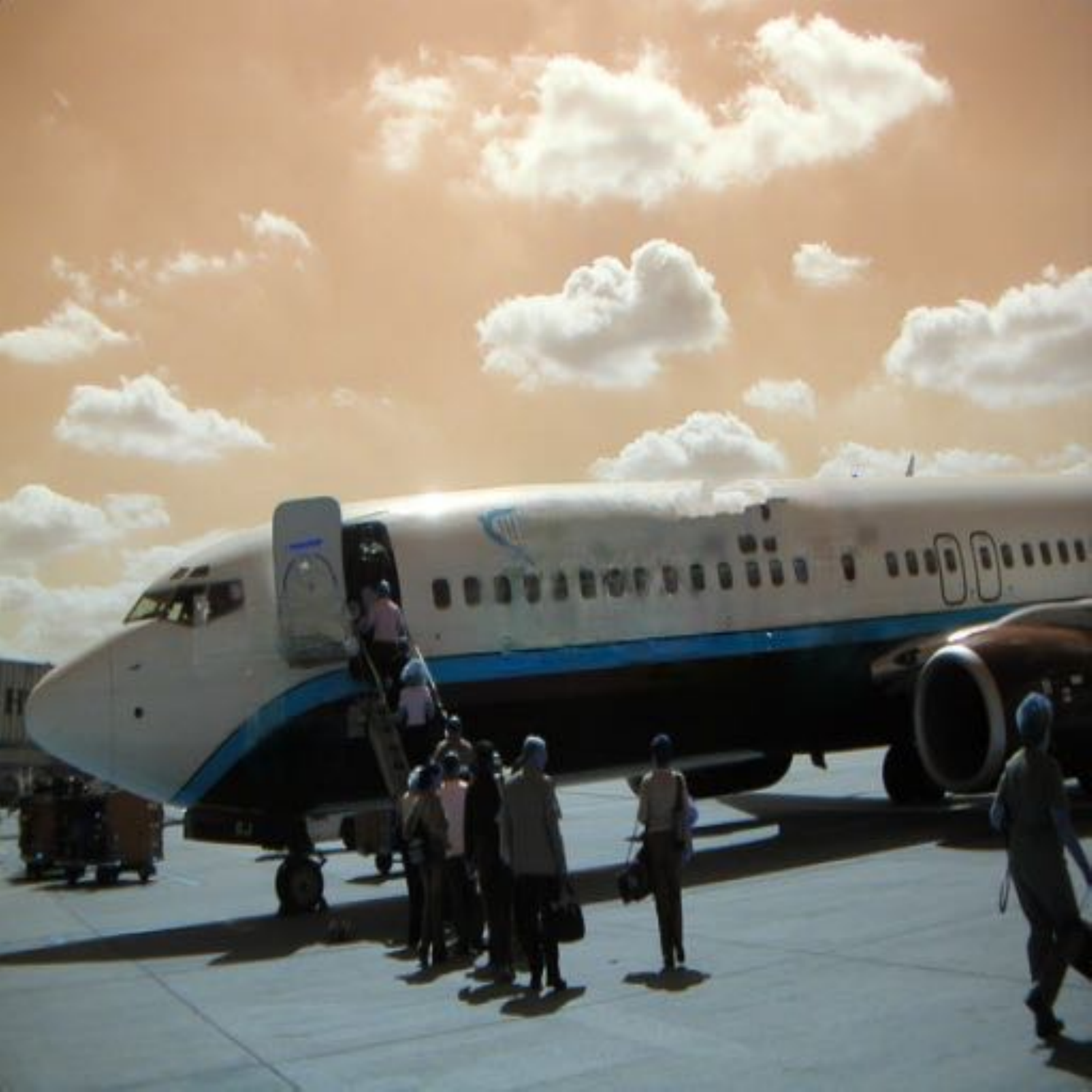}\\
		\end{minipage}%
	}
	\subfigure[Stroke\cite{tang2021stroke}]{
		\begin{minipage}[t]{0.188\linewidth}
			\centering
			\includegraphics[width=2.4cm,height=1.8cm]{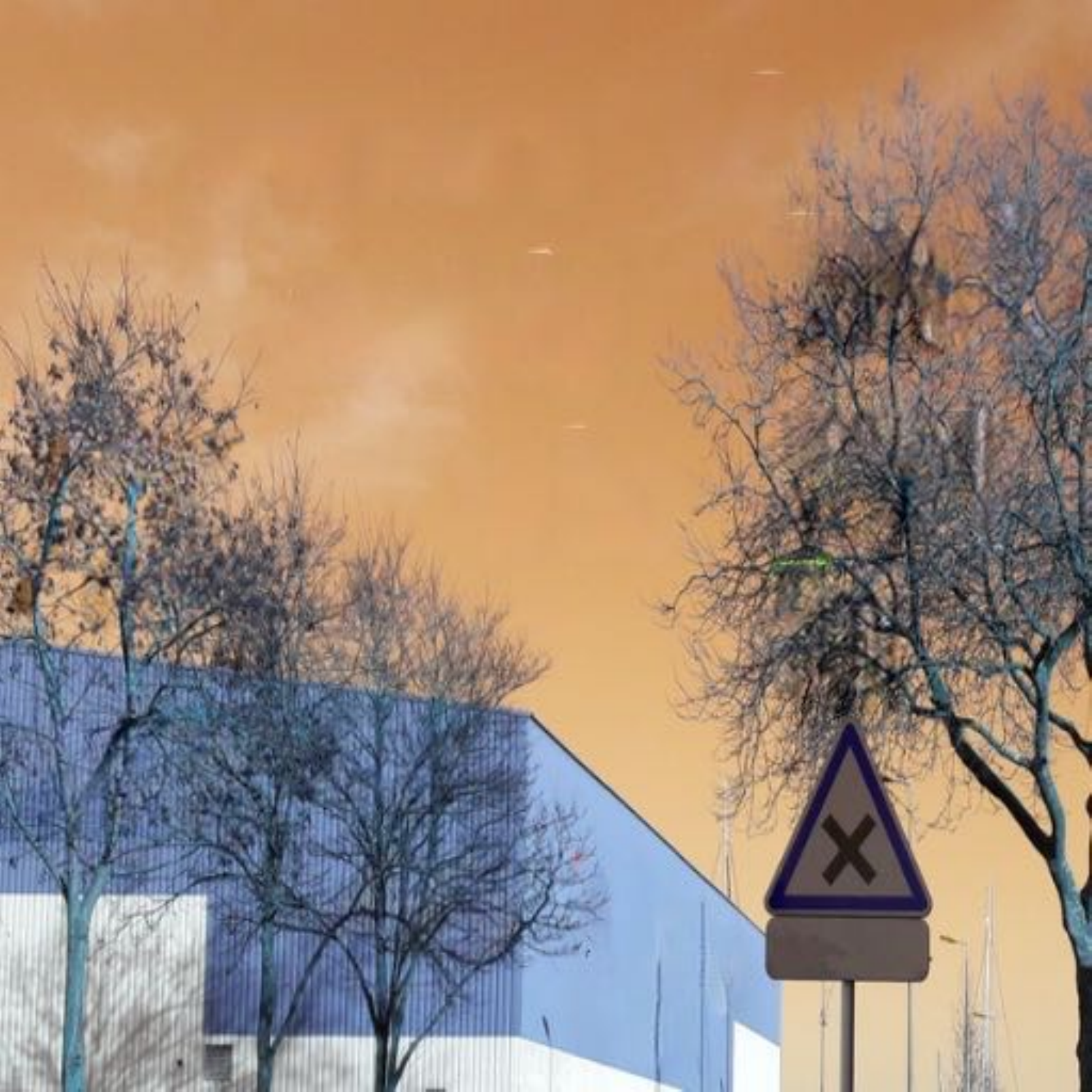}\\
			\includegraphics[width=2.4cm,height=1.8cm]{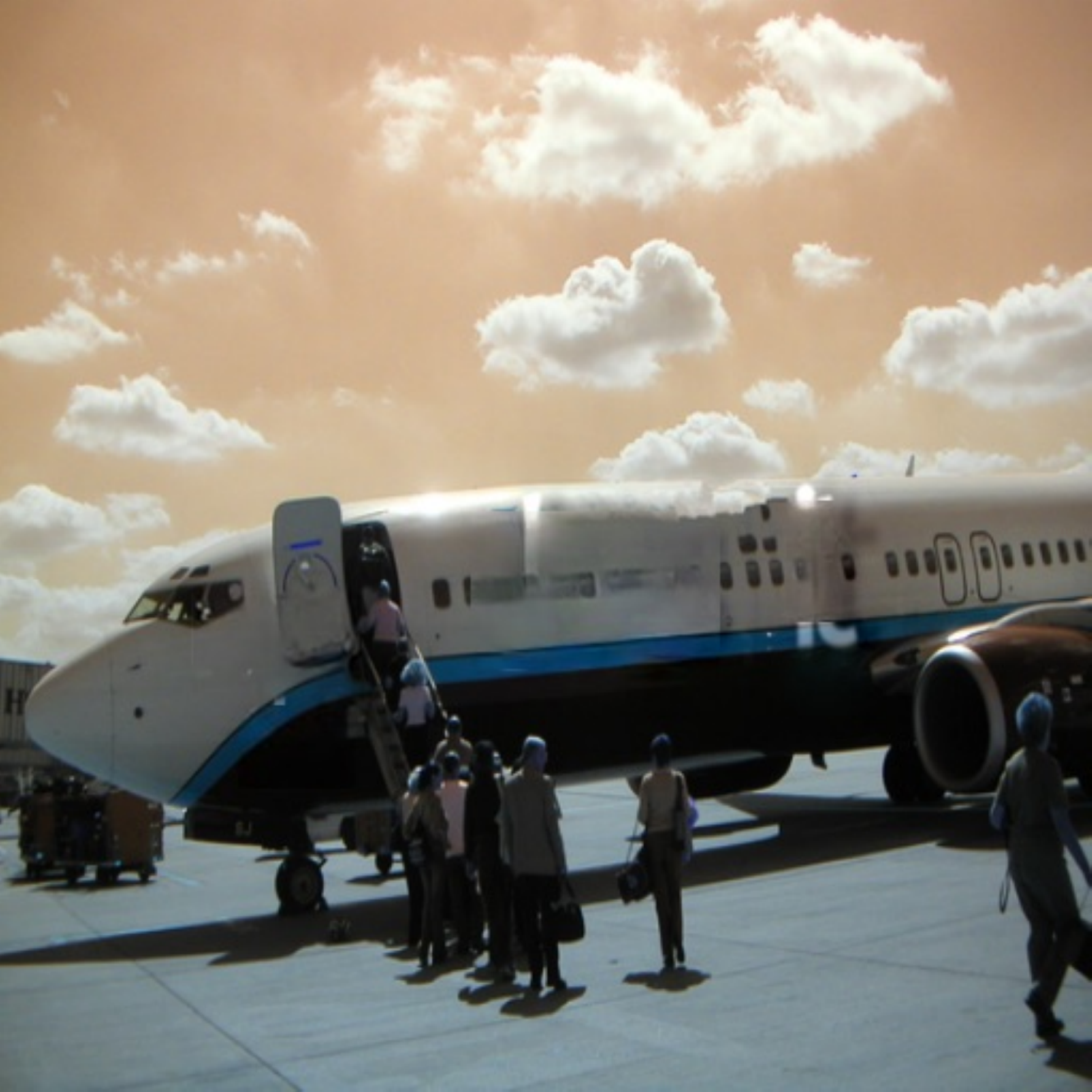}\\
		\end{minipage}%
	}
	\subfigure[Ours]{
		\begin{minipage}[t]{0.18\linewidth}
			\centering
			\includegraphics[width=2.4cm,height=1.8cm]{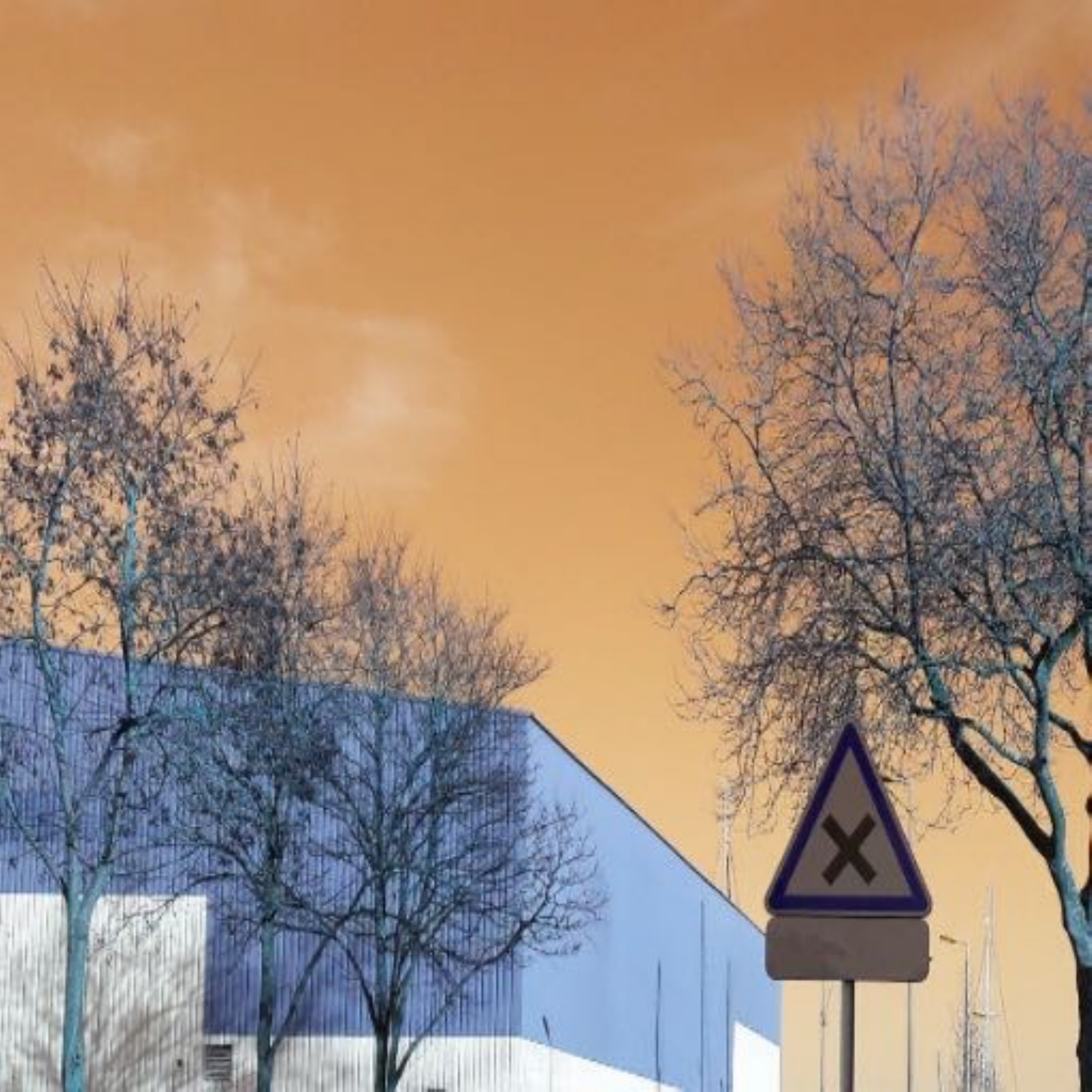}\\
			\includegraphics[width=2.4cm,height=1.8cm]{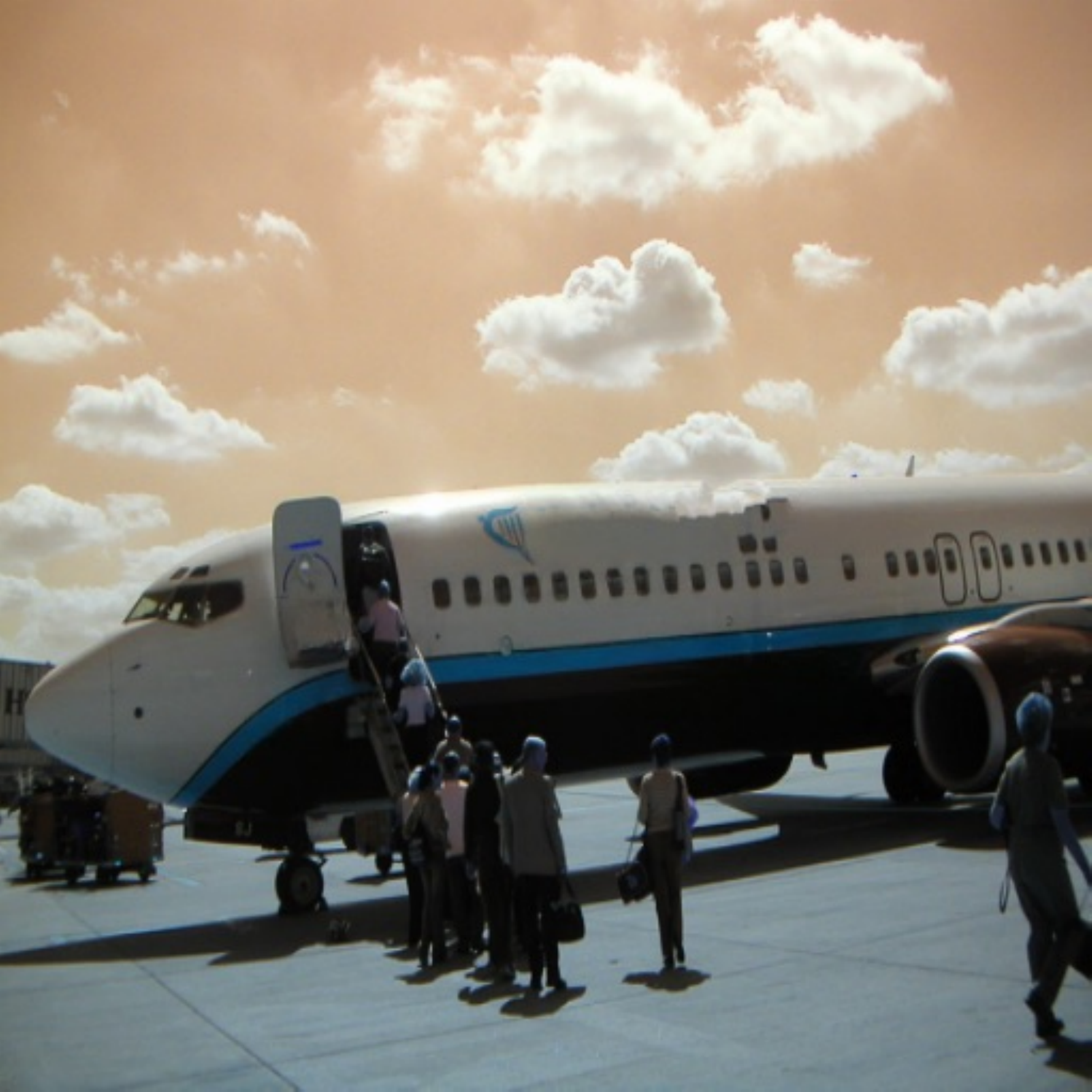}\\
		\end{minipage}%
	}
	\centering
	\caption{Qualitative results on SCUT-Syn for comparing our model with previous scene text removal methods. EN denotes EraseNet~\cite{liu2020erasenet} and Stroke denotes the method proposed by Tang et al.~\cite{tang2021stroke}.  Zoom in for best view.} \label{fig:sota_syn}
\end{figure}

\begin{table*}[t]
	\centering
	\caption{Comparison with state-of-the-art methods on SCUT-Syn. MSSIM and MSE are represented by ($\%$) in the table. Bold indicates SOTA, while Underline indicates second best.}
	\label{table:sota_syn}
	\setlength{\tabcolsep}{1.3mm}{
		\begin{tabular}{ c | c c c c }
			\hline \Xhline{0.3pt}
			\multirow{2}{*}{Methods} & \multicolumn{4}{c}{Image-Eval}   \\
			\cline{2-5} 
			& PSNR & MSSIM & MSE & FID   \\
			\hline \Xhline{0.3pt}
			Pix2pix\cite{isola2017image} & 26.76 & 91.08 & 0.27 &  47.84     \\
			\hline
			STE\cite{ste} & 25.40 & 90.12 & 0.65 &  46.39     \\
			\hline
			EnsNet\cite{zhang2019ensnet} & 37.36 & 96.44 & 0.21 &  -     \\
			\hline
			EnsNet (reimplemented)\cite{zhang2019ensnet} & 36.23 & 96.76 & 0.04  &  19.96    \\
			\hline
			EraseNet\cite{liu2020erasenet} & 38.32 & 97.67 & 0.02 &  9.53     \\
			\hline
			MTRNet++\cite{tursun2020mtrnet++} & 34.55 & 98.45 & \underline{0.04} &  -     \\
			\hline
			Zdenek et al.\cite{Zdenek_2020_WACV} & 37.46 & 93.64 & - &  -     \\
			\hline
			Conrad et al.\cite{conrad2021two} & 32.97 & 94.90 & - &  -     \\
			\hline
			PERT\cite{wang2021simple} & \underline{39.40} & \underline{97.87} & 0.02 &  -     \\
			\hline
			Tang et al.\cite{tang2021stroke} & 38.60 & 97.55 & 0.02 &  -     \\
			\hline 
			Ours & \textbf{41.28} & \textbf{98.50} & \textbf{0.02} & \textbf{3.84}    \\
			\hline \Xhline{0.3pt}
	\end{tabular}}
\end{table*}

\subsection{Comparison with the State-of-the-arts}

In this section, we conduct experiments to evaluate the performance of CTRNet and the relevant SOTA methods on scene text removal on both SCUT-EnsText and SCUT-Syn. The quantitative results on SCUT-EnsText are given in Table \ref{table:sota_enstext}, and 
the quantitative results on SCUT-Syn are given in Table \ref{table:sota_syn}.
The results for these two datasets demonstrate that our proposed model outperforms existing state-of-the-art methods on both Image-Eval and Detection-Eval, indicating that our model can effectively remove the text on th images and meanwhile restore more reasonable background textures. Only the results of Tang et al.~\cite{tang2021stroke} preserve non-text regions of the input (i.e. $I_{com}$) while the others are direct model output $I_{out}$, we evaluate all their performance for fair comparisons.

The qualitative comparison with other methods on SCUT-EnsText is shown in Fig. \ref{fig:sota_ens}, and for SCUT-Syn in Fig. \ref{fig:sota_syn}. 
In Fig. \ref{fig:sota_ens}, 
the results generated by EraseNet~\cite{liu2020erasenet} and Tang et al.~\cite{tang2021stroke} still contain artifacts and discontinuities on the output when dealing with complex text images.
 By utilizing local-global content modeling and different level contextual guidance, our model can predict more realistic textures for text regions and obtain significantly fewer noticeable inconsistencies. 
 Besides, in Fig. \ref{fig:sota_syn}, our model can also generate results of higher quality with more visually pleasing and plausible contents for synthetic data. More cases for comparison are given in the supplement materials.

\begin{table}[t]
	\centering
	\caption{Comparison with state-of-the-art image inpainting methods on SCUT-EnsText. MSSIM and MSE are represented by ($\%$) in the table. }
	\label{table:sota_inpaint}
	\setlength{\tabcolsep}{1.3mm}{
		\begin{tabular}{ c | c c c c }
			\hline \Xhline{0.3pt}
			\multirow{2}{*}{Methods} & \multicolumn{4}{c}{Image-Eval}   \\
			\cline{2-5} 
			& PSNR & MSSIM & MSE & FID   \\
			\hline \Xhline{0.3pt}
			 CTSDG\cite{Guo_2021_ICCV} & 33.10  & 95.55 & 0.14 & 20.01     \\
			\hline
			SPL\cite{SPL} &35.41 & 97.39 & 0.07 &  17.85    \\
			\hline 
			Ours & \textbf{37.20} & \textbf{97.66} & \textbf{0.07} & \textbf{11.72}  \\
			\hline  \Xhline{0.3pt}
	\end{tabular}}
\end{table}

\begin{figure}[t]
	\subfigbottomskip=2pt
	\subfigcapskip=2pt
	\setlength{\abovecaptionskip}{-0.0cm}
	\setlength{\belowcaptionskip}{-0.2cm}
	\centering
	\subfigure[Input]{
		\begin{minipage}[t]{0.17\linewidth}
			\centering
			\includegraphics[width=2.2cm,height=1.7cm]{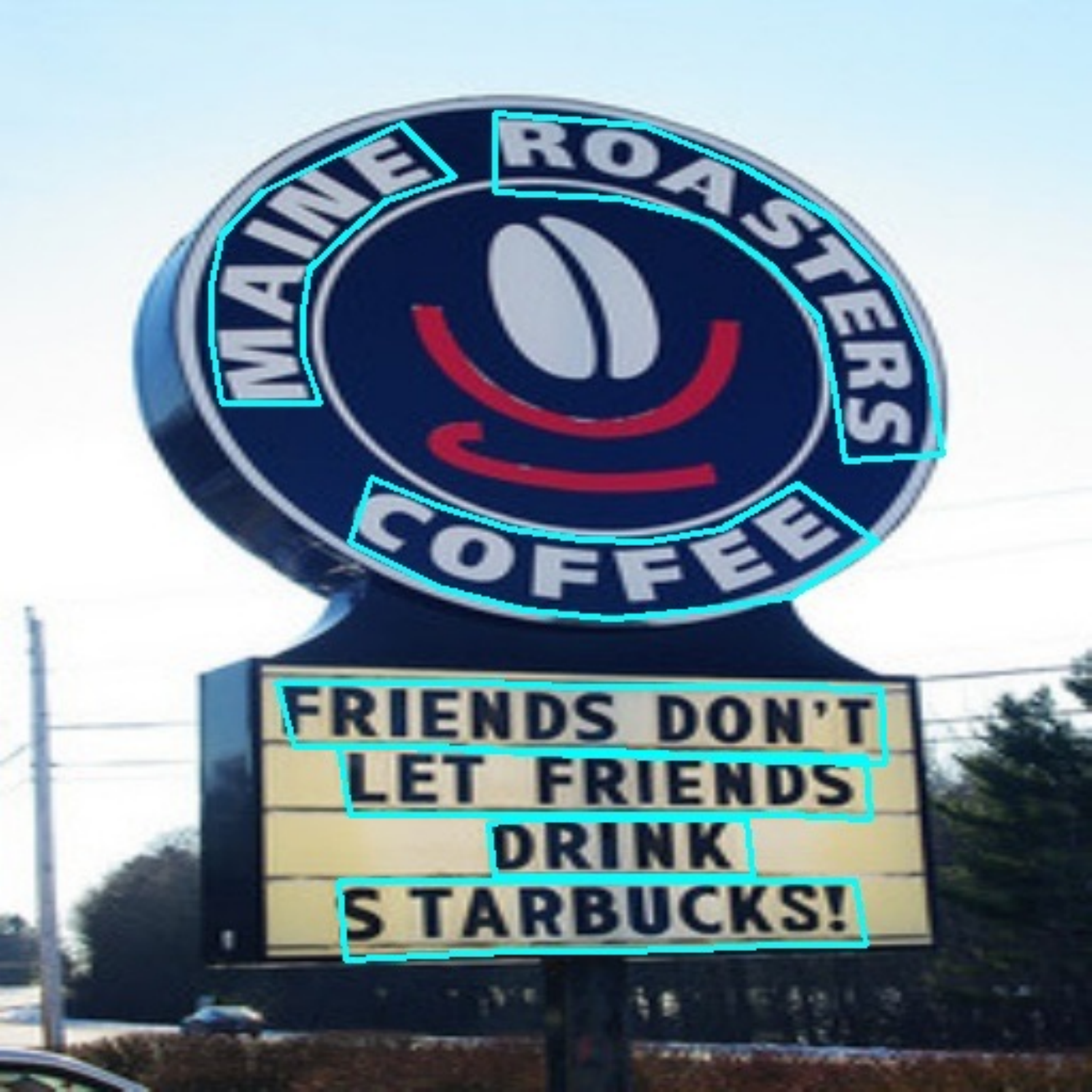}\\
			\includegraphics[width=2.2cm,height=1.7cm]{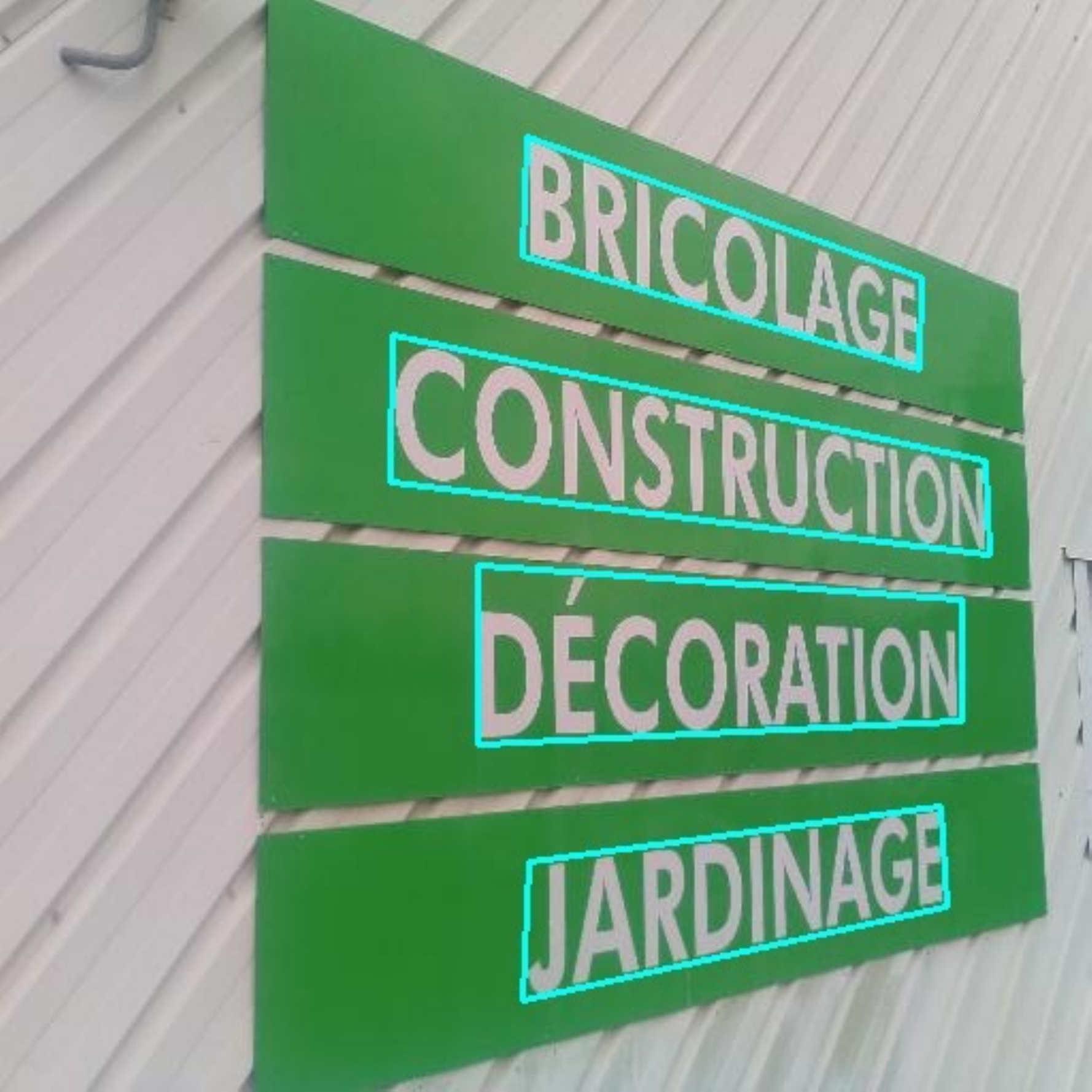}\\
		\end{minipage}%
	}
	\subfigure[GT]{
		\begin{minipage}[t]{0.17\linewidth}
			\centering
			\includegraphics[width=2.2cm,height=1.7cm]{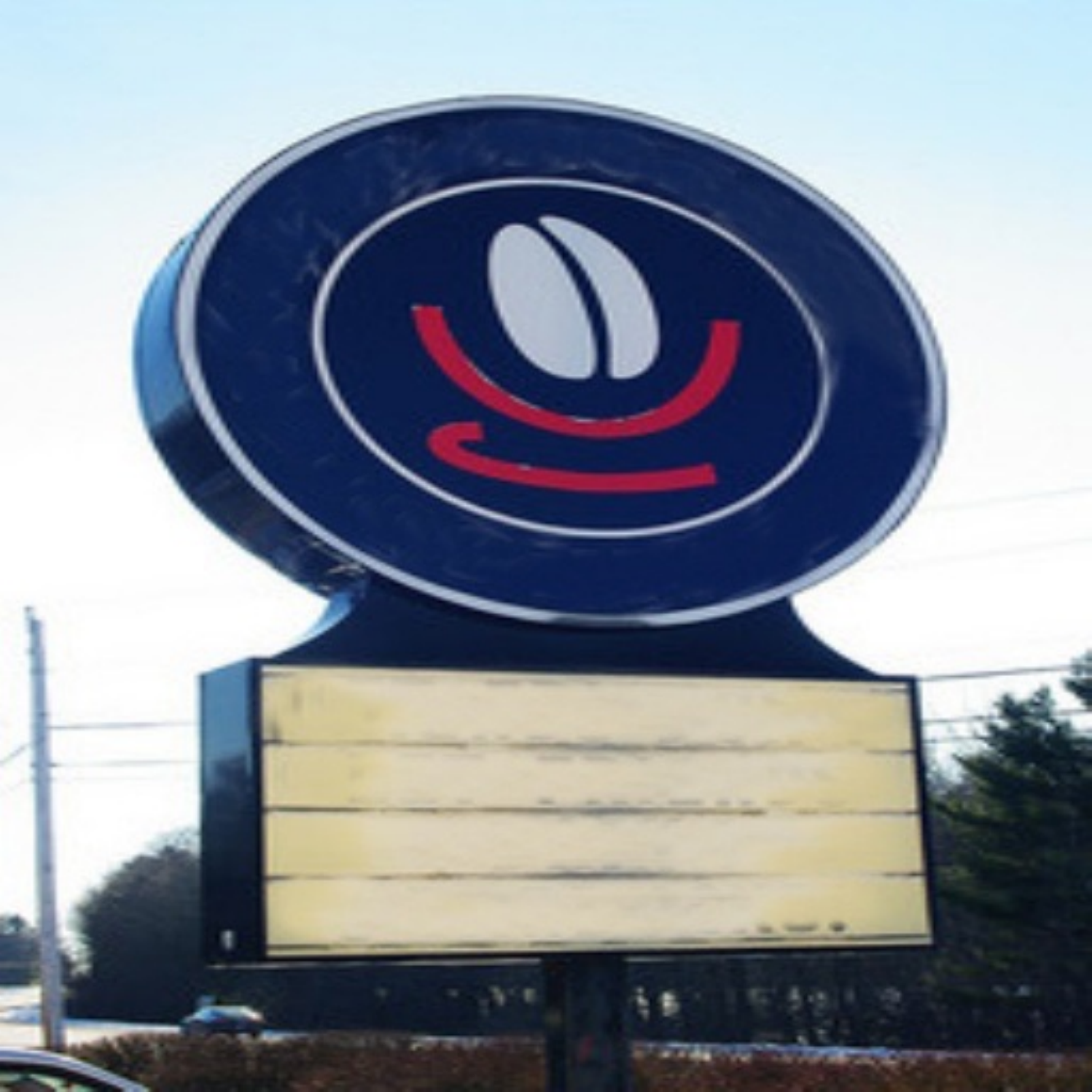}\\
			\includegraphics[width=2.2cm,height=1.7cm]{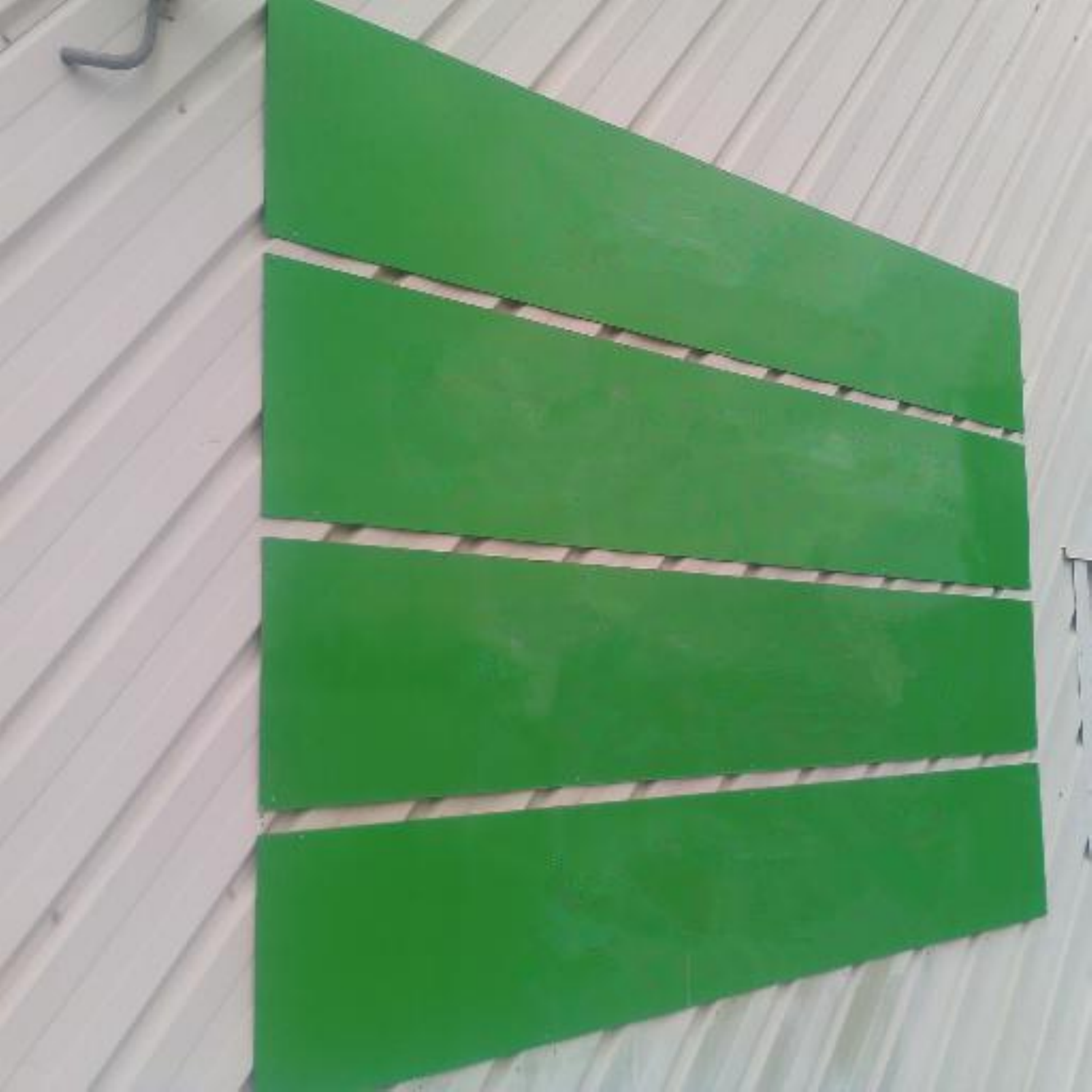}\\
		\end{minipage}%
	}
	\subfigure[CTDSG\cite{Guo_2021_ICCV}]{
		\begin{minipage}[t]{0.17\linewidth}
			\centering
			\includegraphics[width=2.2cm,height=1.7cm]{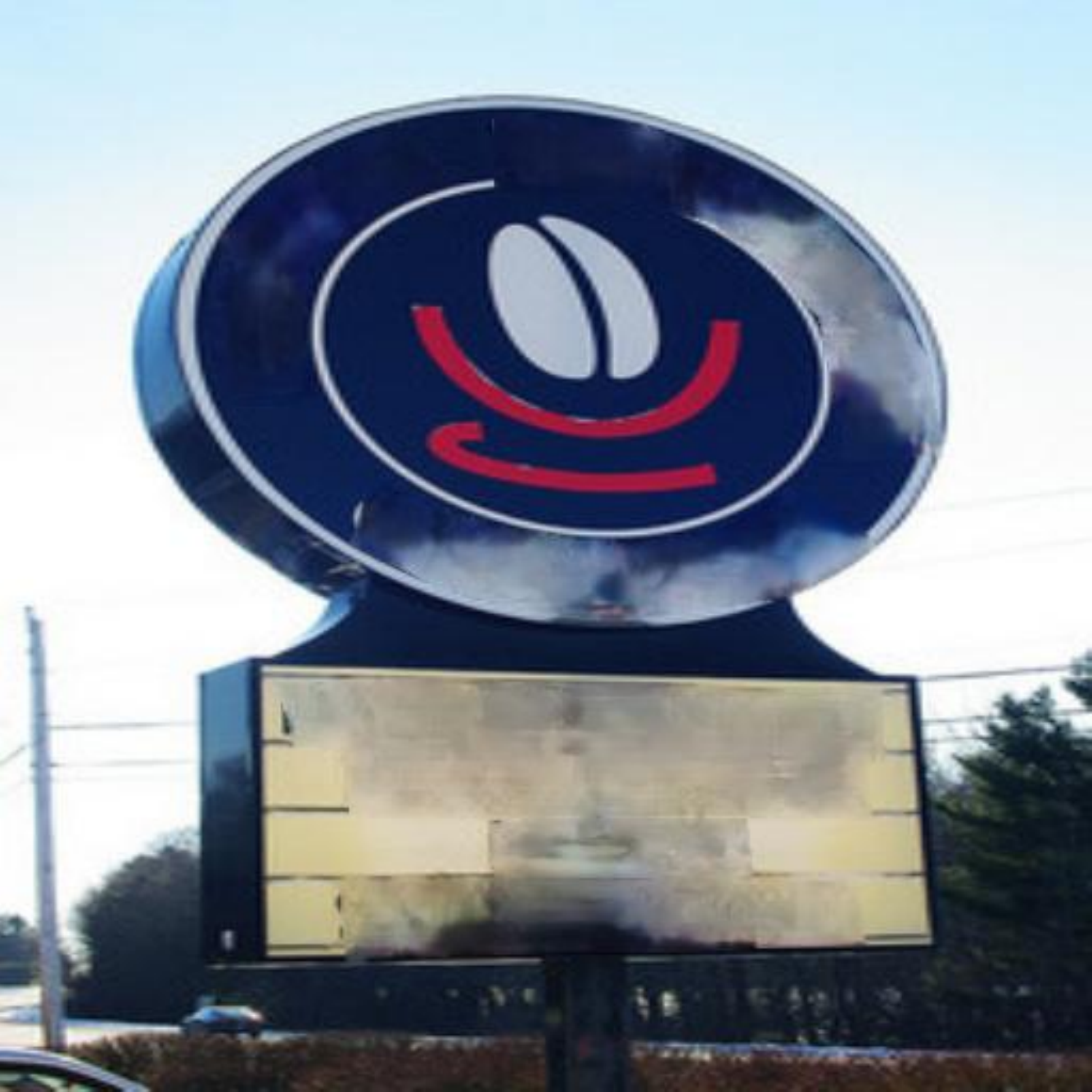}\\
			\includegraphics[width=2.2cm,height=1.7cm]{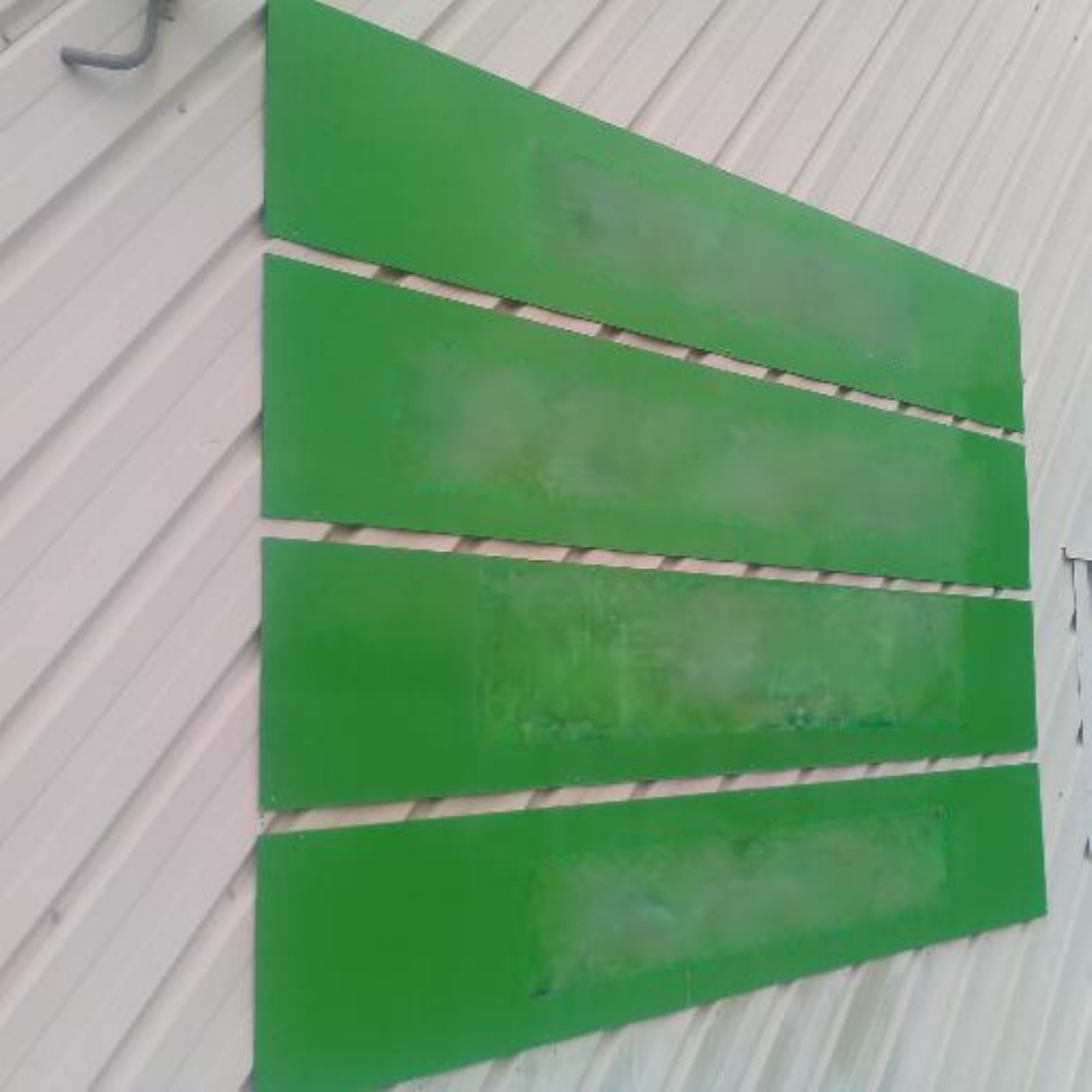}\\
		\end{minipage}%
	}
	\subfigure[SPL\cite{SPL}]{
		\begin{minipage}[t]{0.17\linewidth}
			\centering
			\includegraphics[width=2.2cm,height=1.7cm]{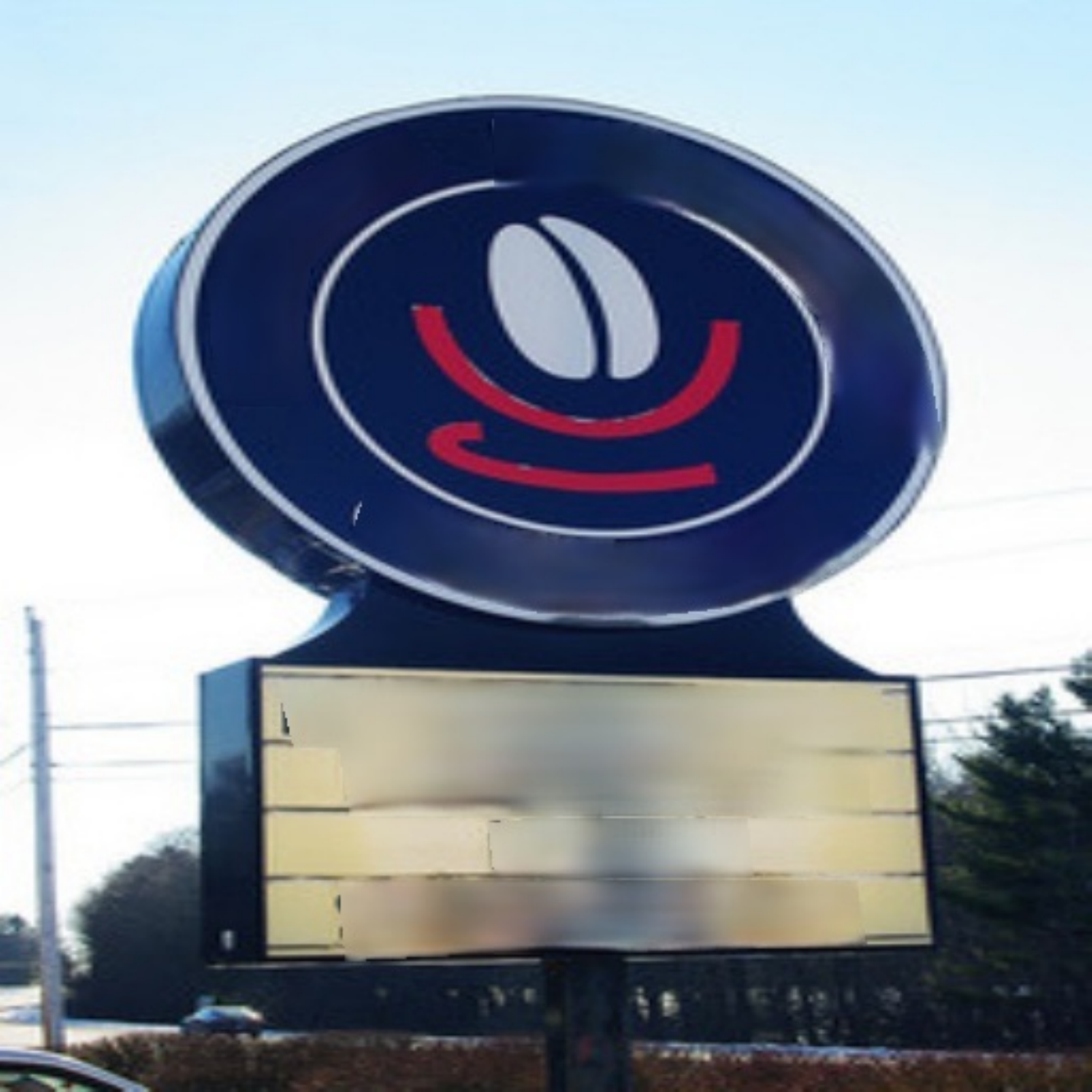}\\
			\includegraphics[width=2.2cm,height=1.7cm]{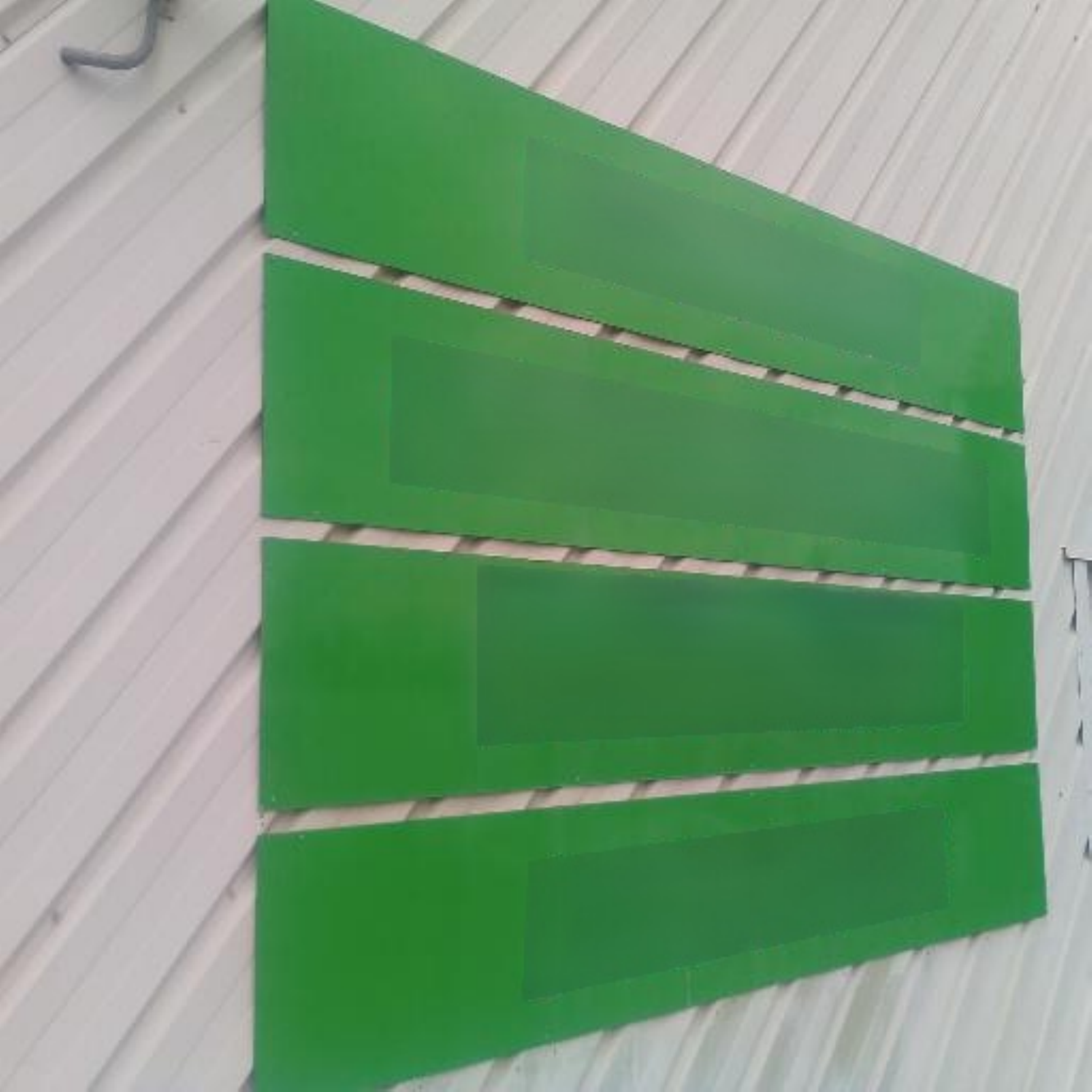}\\
		\end{minipage}%
	}
	\subfigure[Ours]{
		\begin{minipage}[t]{0.17\linewidth}
			\centering
			\includegraphics[width=2.2cm,height=1.7cm]{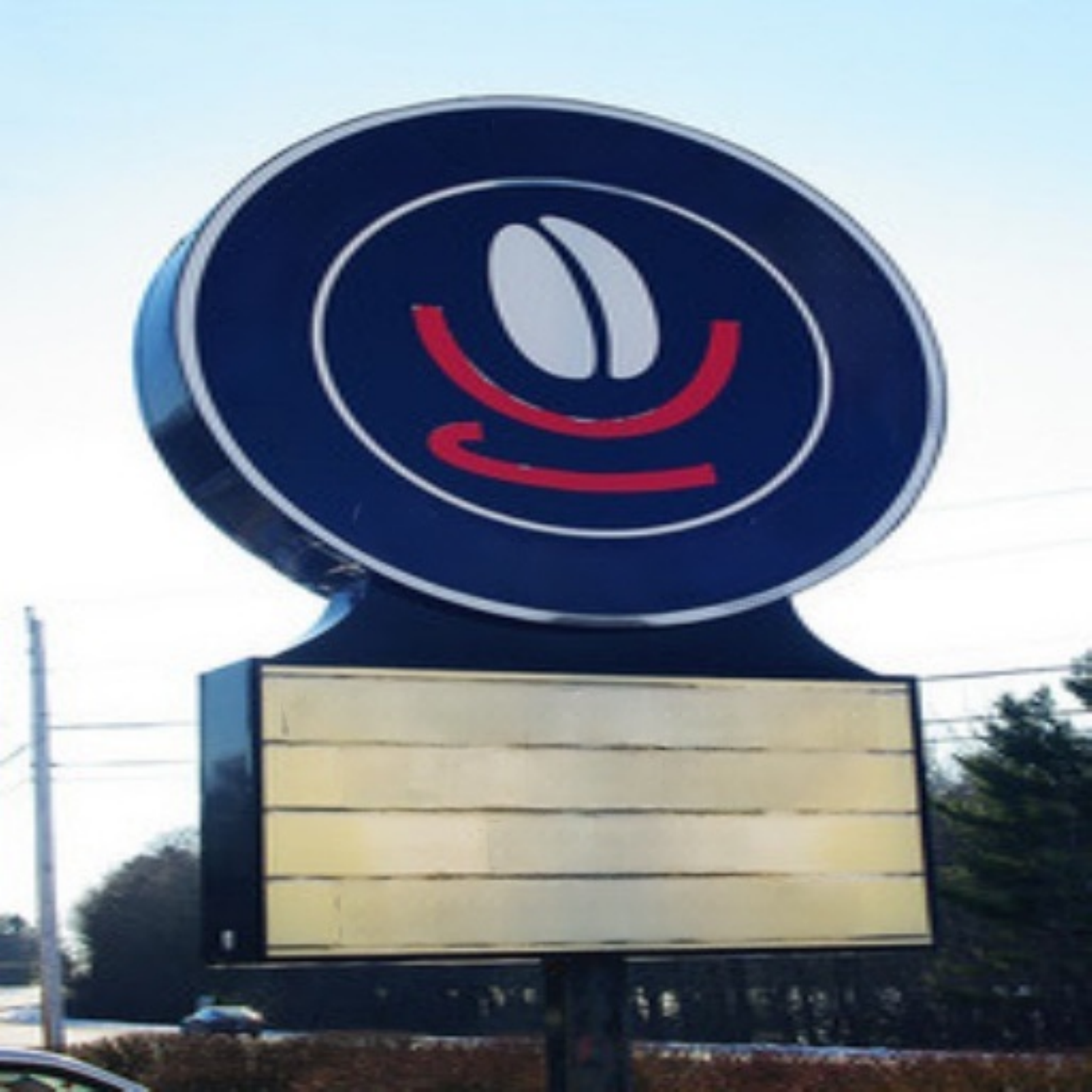}\\
			\includegraphics[width=2.2cm,height=1.7cm]{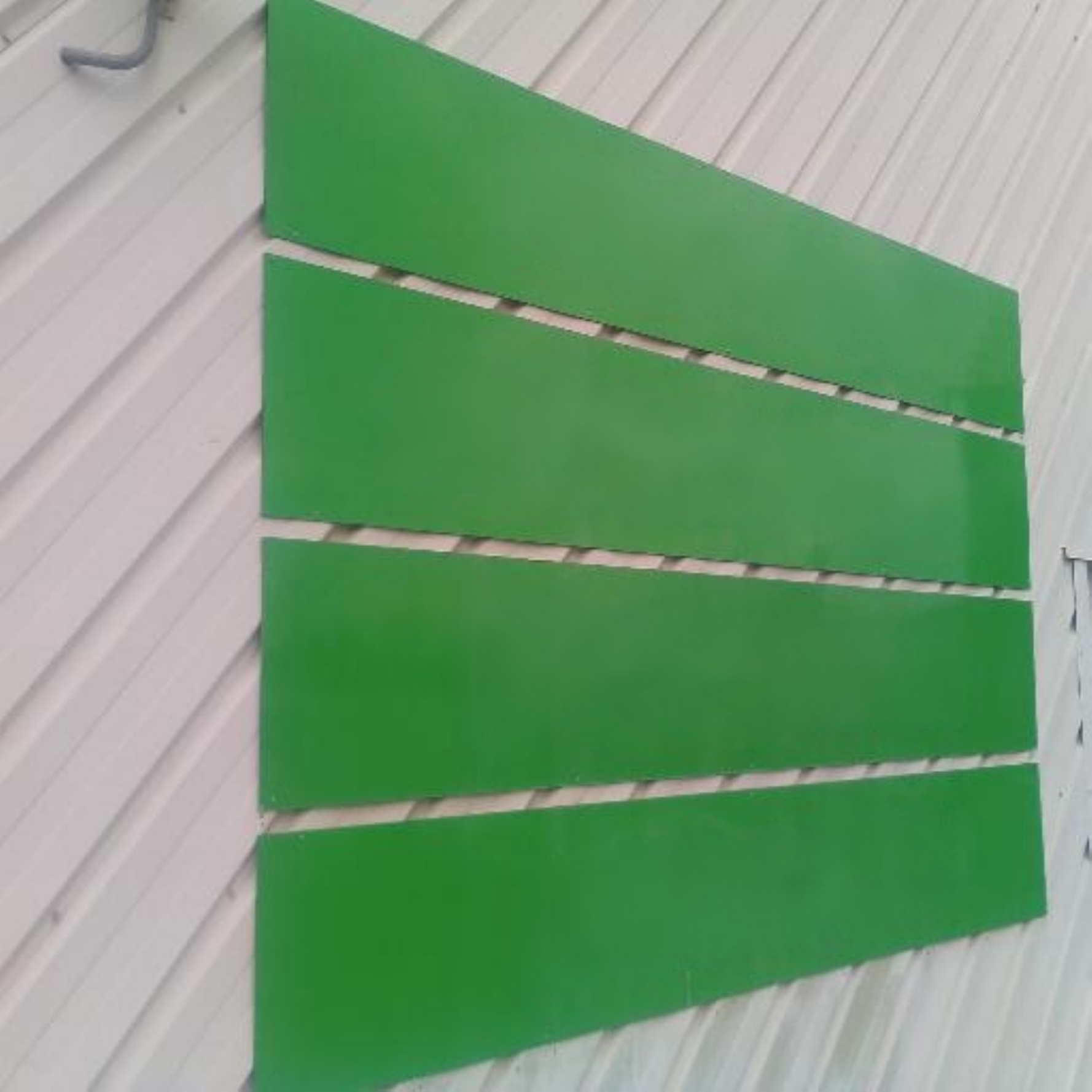}\\
		\end{minipage}%
	}
	\centering
	\caption{Qualitative results on SCUT-EnsText for comparing our model with state-of-the-art image inpainting methods. Zoom in for best view.} \label{fig:sota_inpaint}
\end{figure}

\begin{figure}[]
	\subfigbottomskip=2pt
	\subfigcapskip=2pt
	\setlength{\abovecaptionskip}{-0.0cm}
	\setlength{\belowcaptionskip}{-0.0cm}
	\centering
	\subfigure[Input]{
		\begin{minipage}[t]{0.23\linewidth}
			\centering
			\includegraphics[width=2.3cm,height=2.0cm]{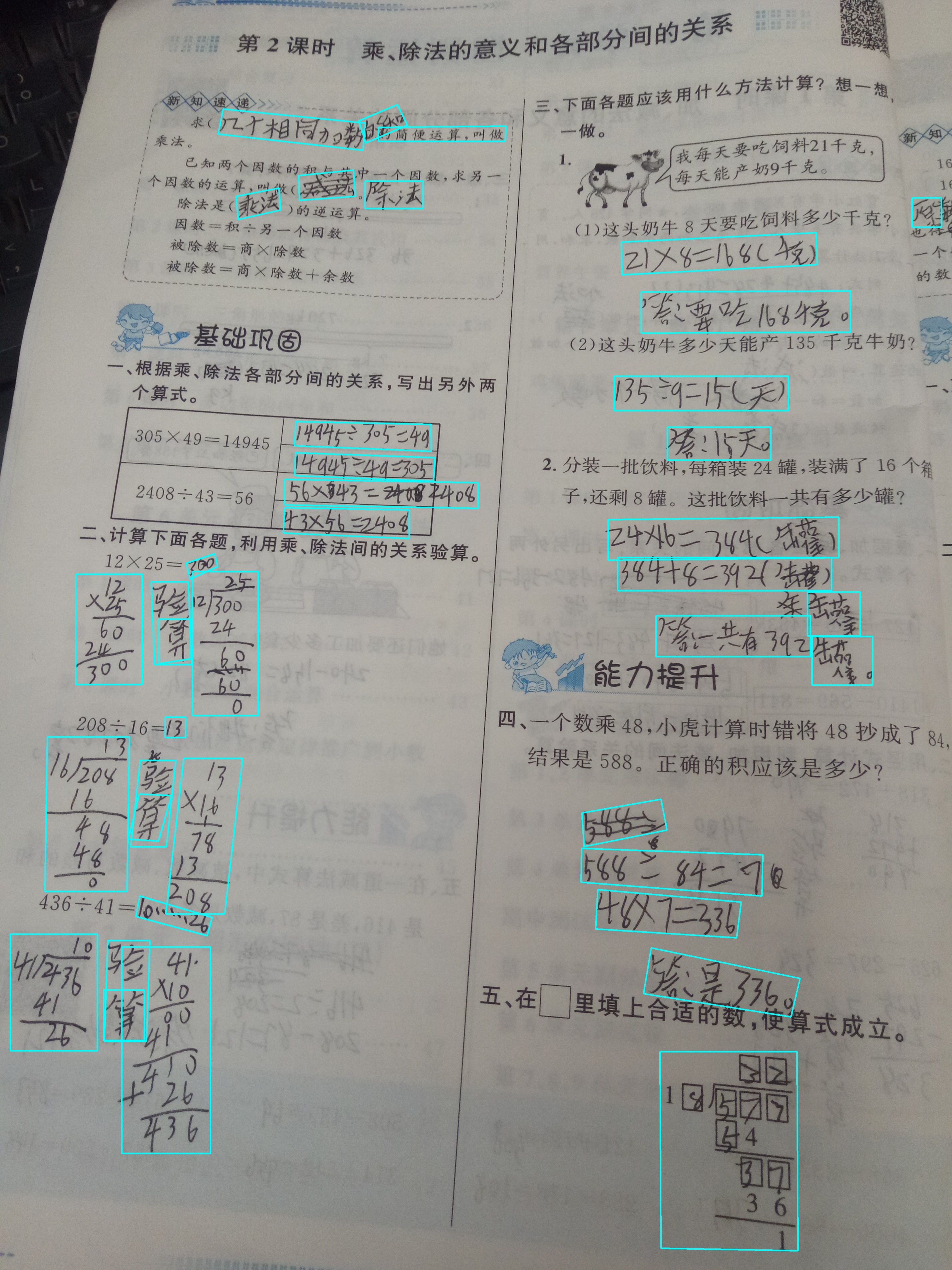}\\
			\includegraphics[width=2.3cm,height=2.0cm]{./figure/other_task/hehe/input/315.pdf}\\
		\end{minipage}%
	}
	\subfigure[SPL\cite{SPL}]{
		\begin{minipage}[t]{0.23\linewidth}
			\centering
			\includegraphics[width=2.3cm,height=2.0cm]{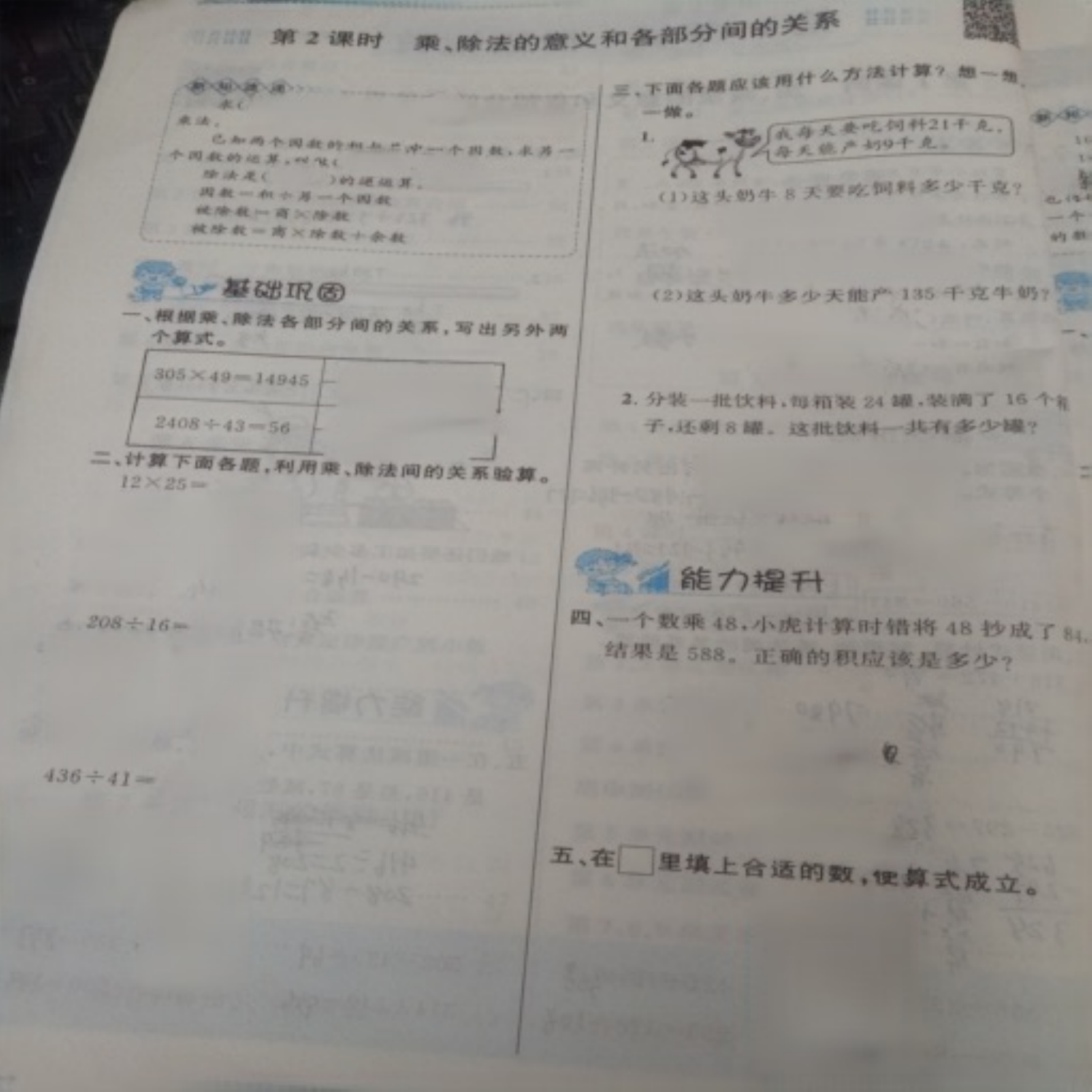}\\
			\includegraphics[width=2.3cm,height=2.0cm]{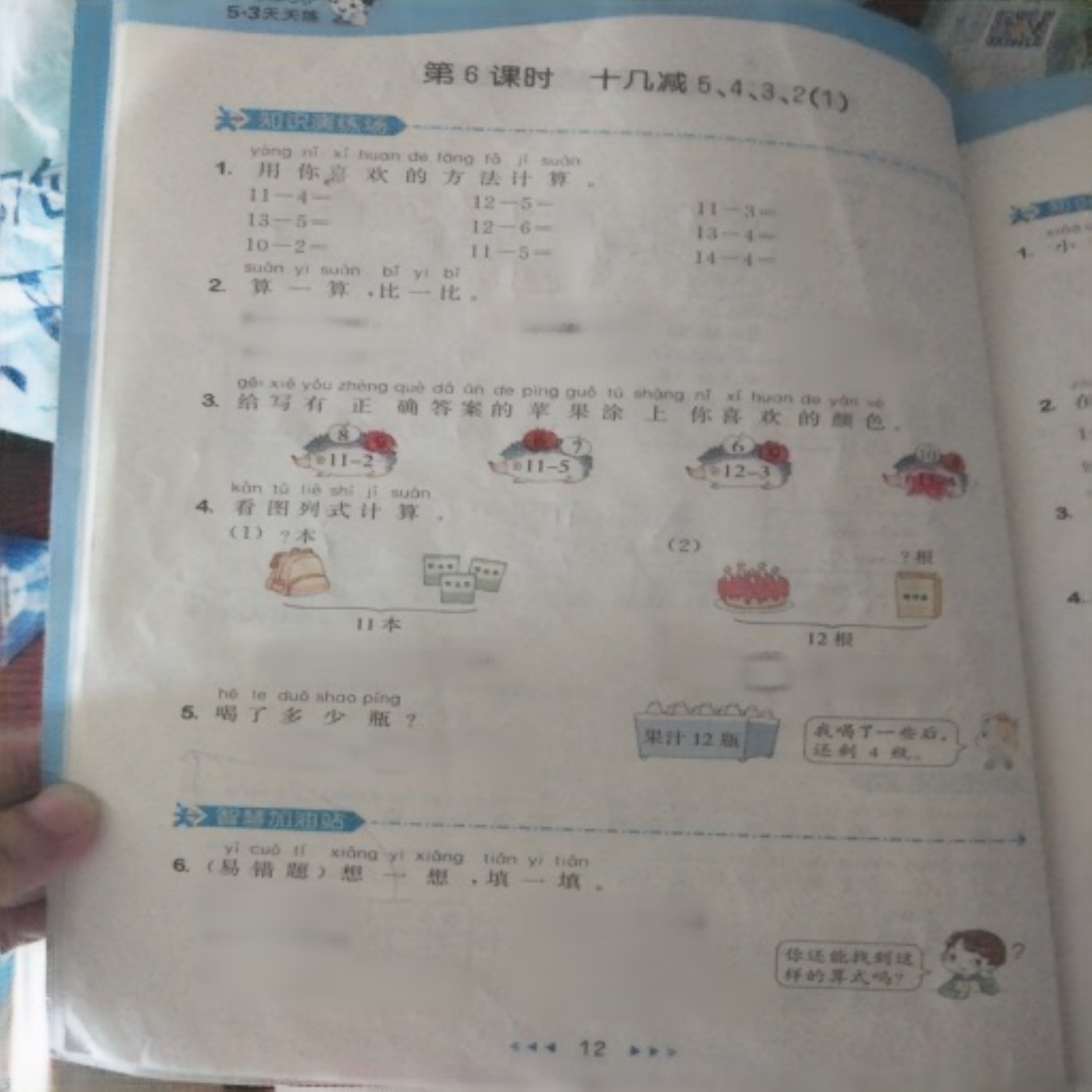}\\
		\end{minipage}%
	}
	\subfigure[EraseNet\cite{liu2020erasenet}]{
	\begin{minipage}[t]{0.23\linewidth}
		\centering
		\includegraphics[width=2.3cm,height=2.0cm]{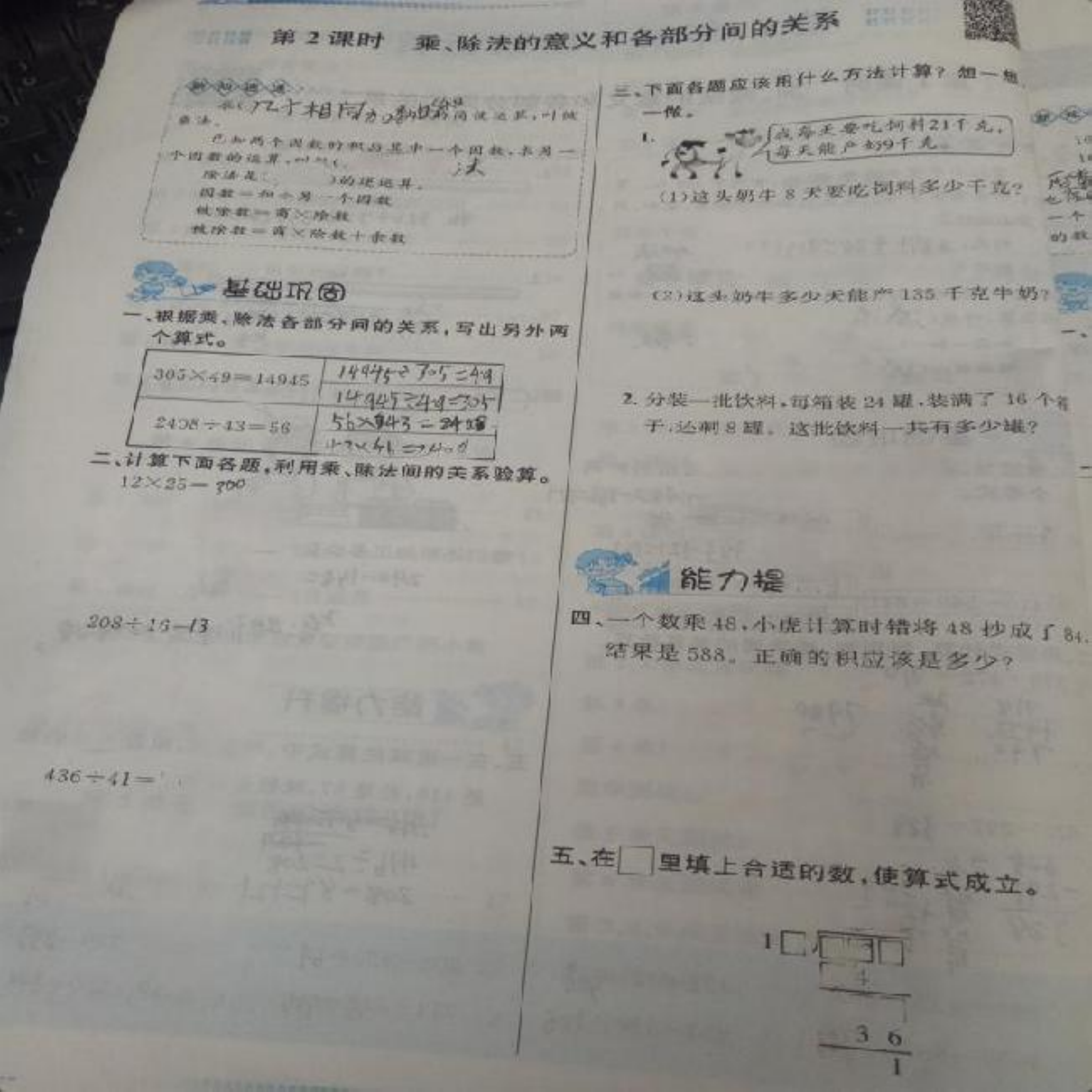}\\
		\includegraphics[width=2.3cm,height=2.0cm]{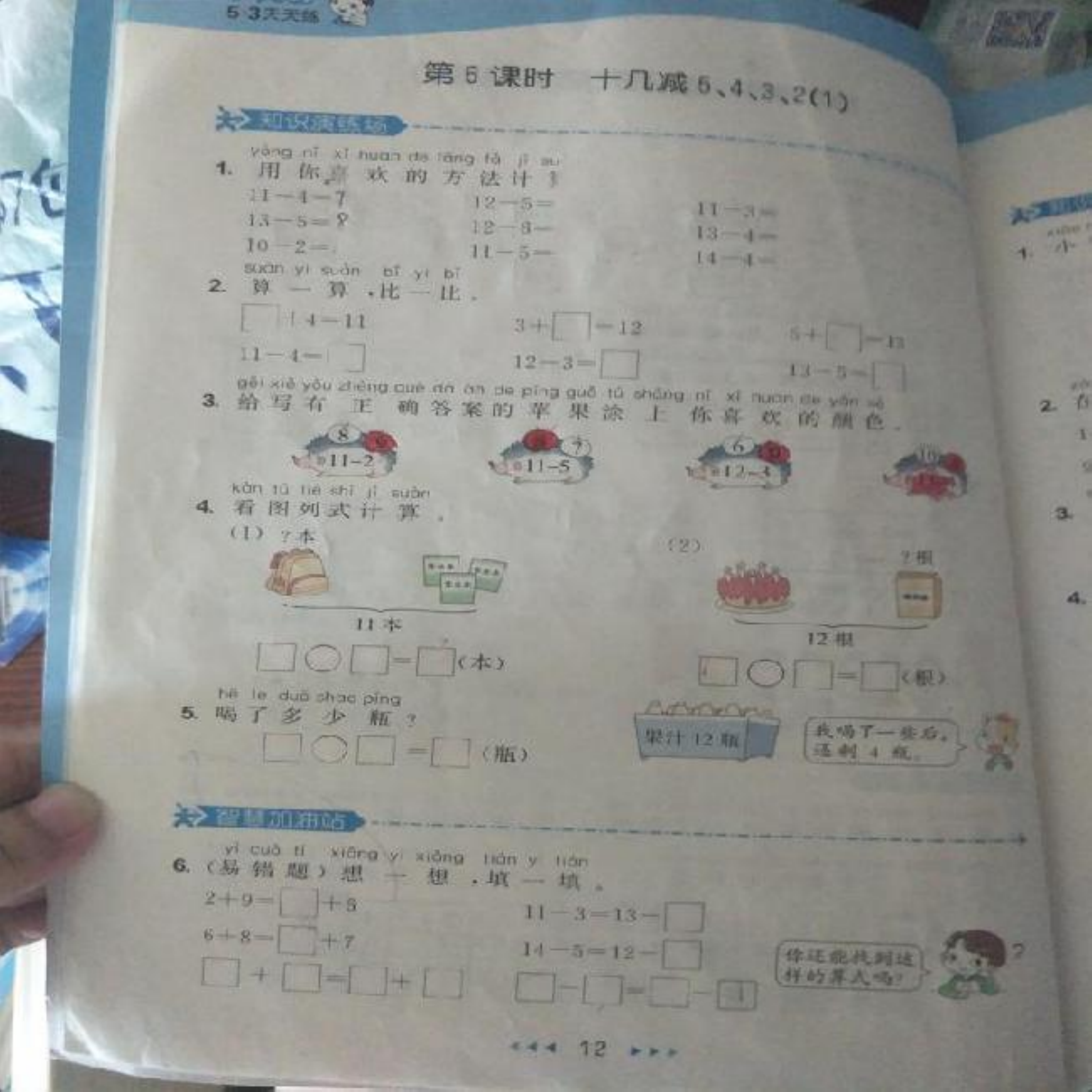}\\
	\end{minipage}%
}
	\subfigure[Ours]{
		\begin{minipage}[t]{0.23\linewidth}
			\centering
			\includegraphics[width=2.3cm,height=2.0cm]{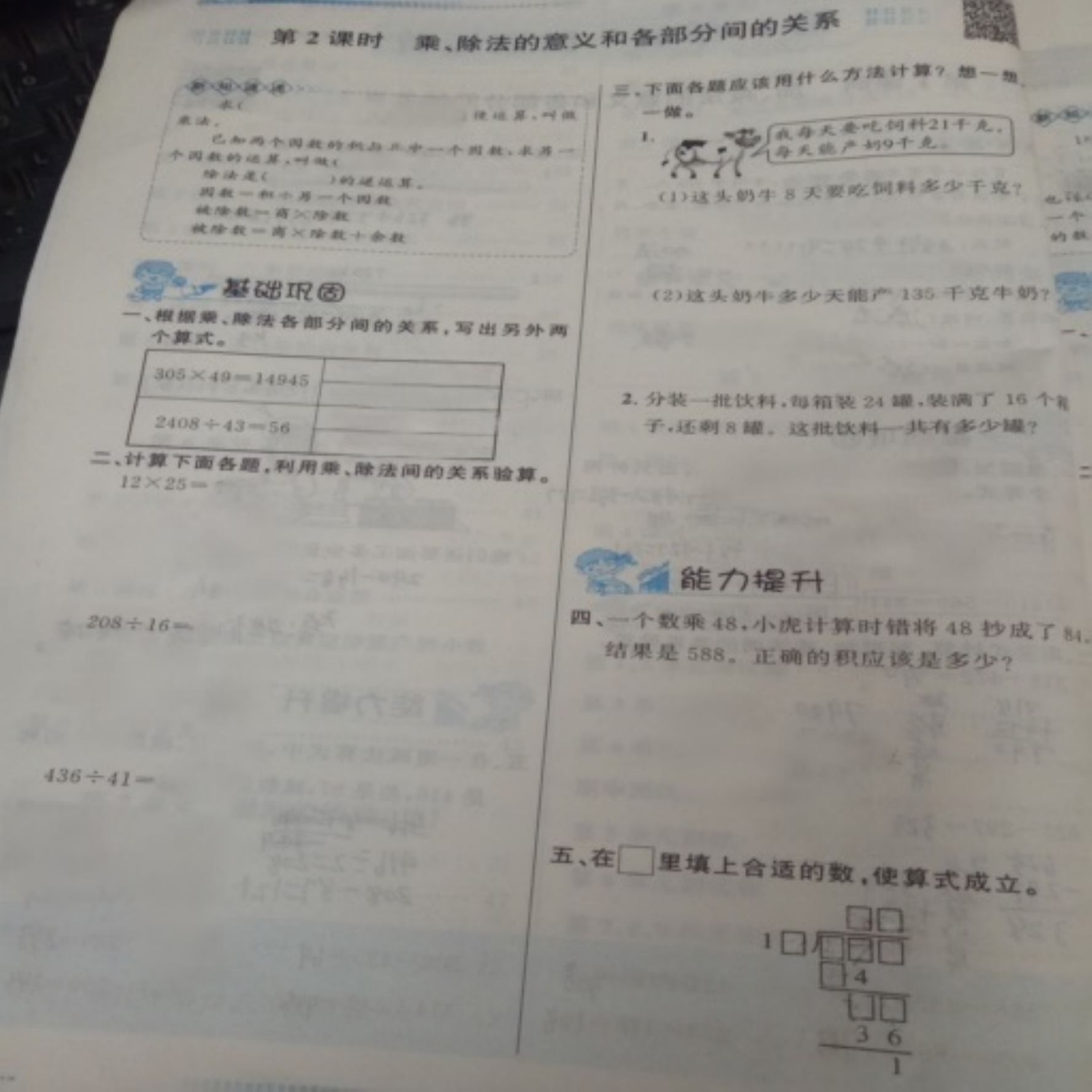}\\
			\includegraphics[width=2.3cm,height=2.0cm]{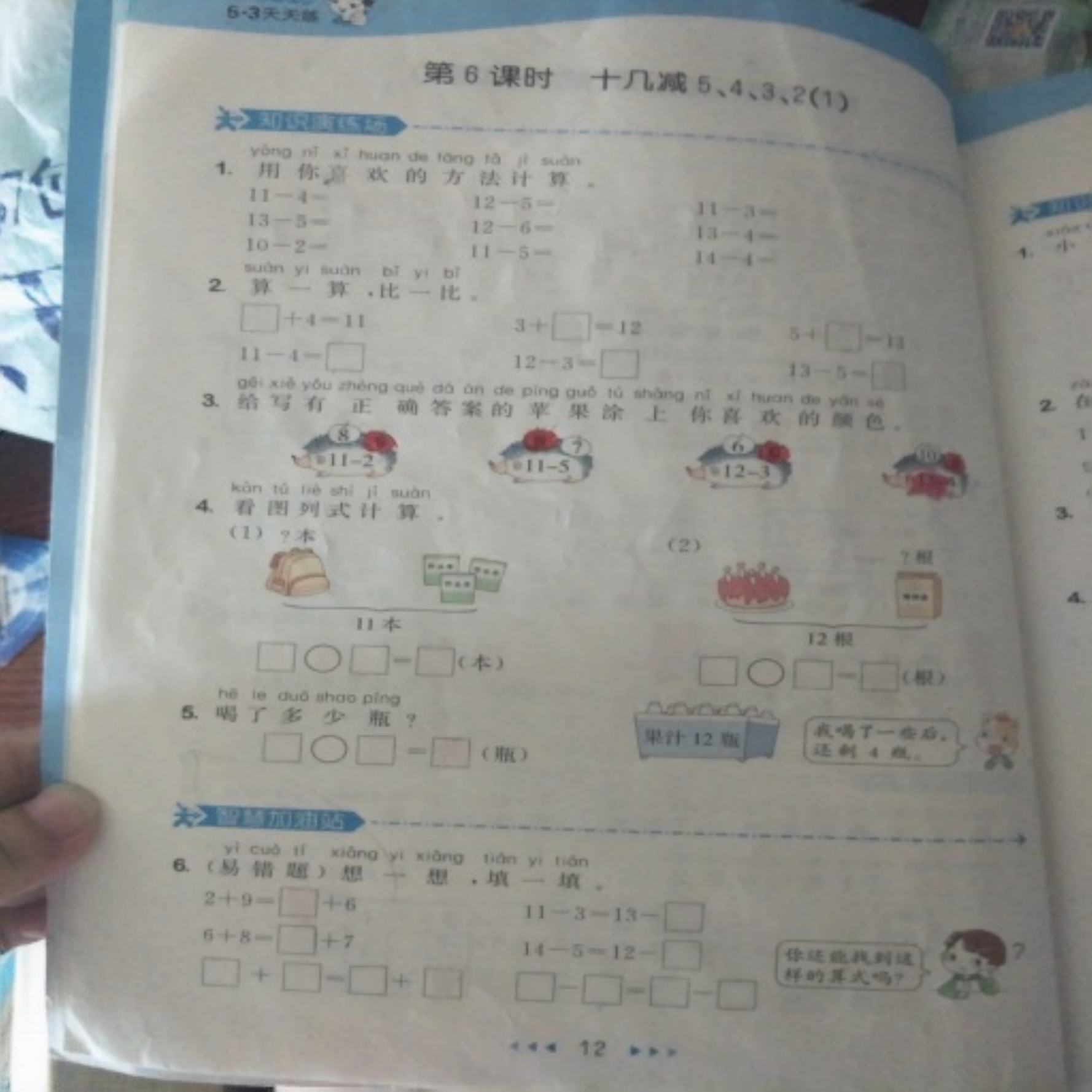}\\
		\end{minipage}%
	}
	\centering
	\caption{Qualitative results on examination papers. Zoom in for best view.} \label{fig:hehe}
\end{figure}

\subsection{Comparison with State-of-the-art Image Inpainting Methods}

We conduct experiments to compare CTRNet with existing SOTA image inpainting methods, CTSDG \cite{Guo_2021_ICCV} and SPL \cite{SPL} on SCUT-EnsText. The quantitative and qualitative results are given in Table \ref{table:sota_inpaint} and Fig. \ref{fig:sota_inpaint}, respectively. Our model outperforms these two methods in all metrics with a remarkable margin, meanwhile can restore the text region background with more reasonable and realistic textures. 
The reason is that while we simply apply image inpainting methods on scene text removal, the text regions will be directly abandoned by masking according to the bounding boxes (blue boxes in Fig. \ref{fig:sota_inpaint} (a)), causing that the model can not effectively deduce the background information.

\subsection{Application on Handwritten Text Removal}

In this section, we apply CTRNet to help restore document images to verify its generalizability. We collect 1,000 in-house examination paper images and manually annotate them by erasing the handwriting with PhotoShop. Then we train CTRNet and EraseNet~\cite{liu2020erasenet} on the data and evaluate them with other paper images. Besides, we also train SPL \cite{SPL} for comparison to further illustrate the difference between text removal and image inpainting. The visualization results are shown in Fig. \ref{fig:hehe}. Our method can retain more printed words than SPL and EraseNet, which is more suitable for document restoration task. 
More results are given in the supplement materials.


\section{Conclusion}
In this paper, we propose 
a new text removal model called CTRNet. CTRNet introduces both low-level and high-level contextual guidance, which can effectively promote the performance on texture restoration for complex backgrounds.
We further use smooth structure images and discriminative context features to represent the low-level and high-level context, respectively.
Besides, the learned contextual guidance is incorporated into the image features and modeled in a local-global manner to effectively capture both sufficient context information and long-term correlation among all of the pixels.   
The experiments conducted on three benchmark datasets have demonstrated the effectiveness of the proposed
CTRNet, outperforming previous state-of-the-art methods significantly.

\par\vfill\par

\clearpage
%
%

\end{document}